\documentclass{article}

\PassOptionsToPackage{numbers, compress}{natbib}

\usepackage[final]{neurips_2022}

\usepackage{listings}

\usepackage[utf8]{inputenc} 
\usepackage[T1]{fontenc}    
\usepackage{hyperref}       
\usepackage{url}            
\usepackage{booktabs}       
\usepackage{amsfonts}       
\usepackage{nicefrac}       
\usepackage{microtype}
\usepackage{graphicx}
\usepackage{booktabs} 
\usepackage{makecell}
\usepackage{graphicx}
\usepackage{caption}
\usepackage{courier}
\usepackage{wrapfig}
\usepackage{bm}
\usepackage{multirow}
\usepackage{color,xcolor}
\usepackage{subcaption}
\usepackage{amssymb}
\usepackage{amsmath}
\usepackage{gensymb}
\usepackage{amsfonts}       
\usepackage{enumitem}
\usepackage{multirow}
\usepackage{algorithm}
\usepackage[noend]{algpseudocode}
\usepackage{amsmath,nccmath}
\usepackage{amsthm}
\renewcommand\footnotemark{}
\usepackage{hyperref}
\definecolor{dkgreen}{rgb}{0,0.6,0}
\definecolor{gray}{rgb}{0.5,0.5,0.5}
\definecolor{mauve}{rgb}{0.58,0,0.82}

\newcommand{\revision}[1]{{\color{black}#1}}
    
\title{Wild-Time: A Benchmark of in-the-Wild Distribution Shift over Time}
\author{Huaxiu Yao$^{1*}$\thanks{$^*$Huaxiu Yao and Caroline Choi contributed equally.}, Caroline Choi$^{1*}$, Bochuan Cao$^{2}$, Yoonho Lee$^{1}$, Pang Wei Koh$^{1}$, Chelsea Finn$^{1}$\\ $^{1}$Stanford University, $^{2}$Pennsylvania State University\\ \href{mailto:wildtime@googlegroups.com}{wildtime@googlegroups.com}}
\begin{document}

\maketitle
\begin{abstract}
Distribution shift occurs when the test distribution differs from the training distribution, and it can considerably degrade performance of machine learning models deployed in the real world. \textit{Temporal shifts} -- distribution shifts arising from the passage of time -- often occur gradually and have the additional structure of timestamp metadata. By leveraging timestamp metadata, models can potentially learn from trends in past distribution shifts and extrapolate into the future. While recent works have studied distribution shifts, temporal shifts remain underexplored. To address this gap, we curate Wild-Time, a benchmark of 5 datasets that reflect temporal distribution shifts arising in a variety of real-world applications, including patient prognosis and news classification. On these datasets, we systematically benchmark 13 prior approaches, including methods in domain generalization, continual learning, self-supervised learning, and ensemble learning. We use two evaluation strategies: evaluation with a fixed time split (Eval-Fix) and evaluation with a data stream (Eval-Stream). Eval-Fix, our primary evaluation strategy, aims to provide a simple evaluation protocol, while Eval-Stream is more realistic for certain real-world applications. Under both evaluation strategies, we observe an average performance drop of 20\% from in-distribution to out-of-distribution data. Existing methods are unable to close this gap. Code is available at \href{https://wild-time.github.io/}{https://wild-time.github.io/}.

\end{abstract}
\section{Introduction}
\label{sec:intro}
\begin{wrapfigure}{r}{0.5\textwidth}
\vspace{-1em}
\centering\includegraphics[width=0.5\textwidth]{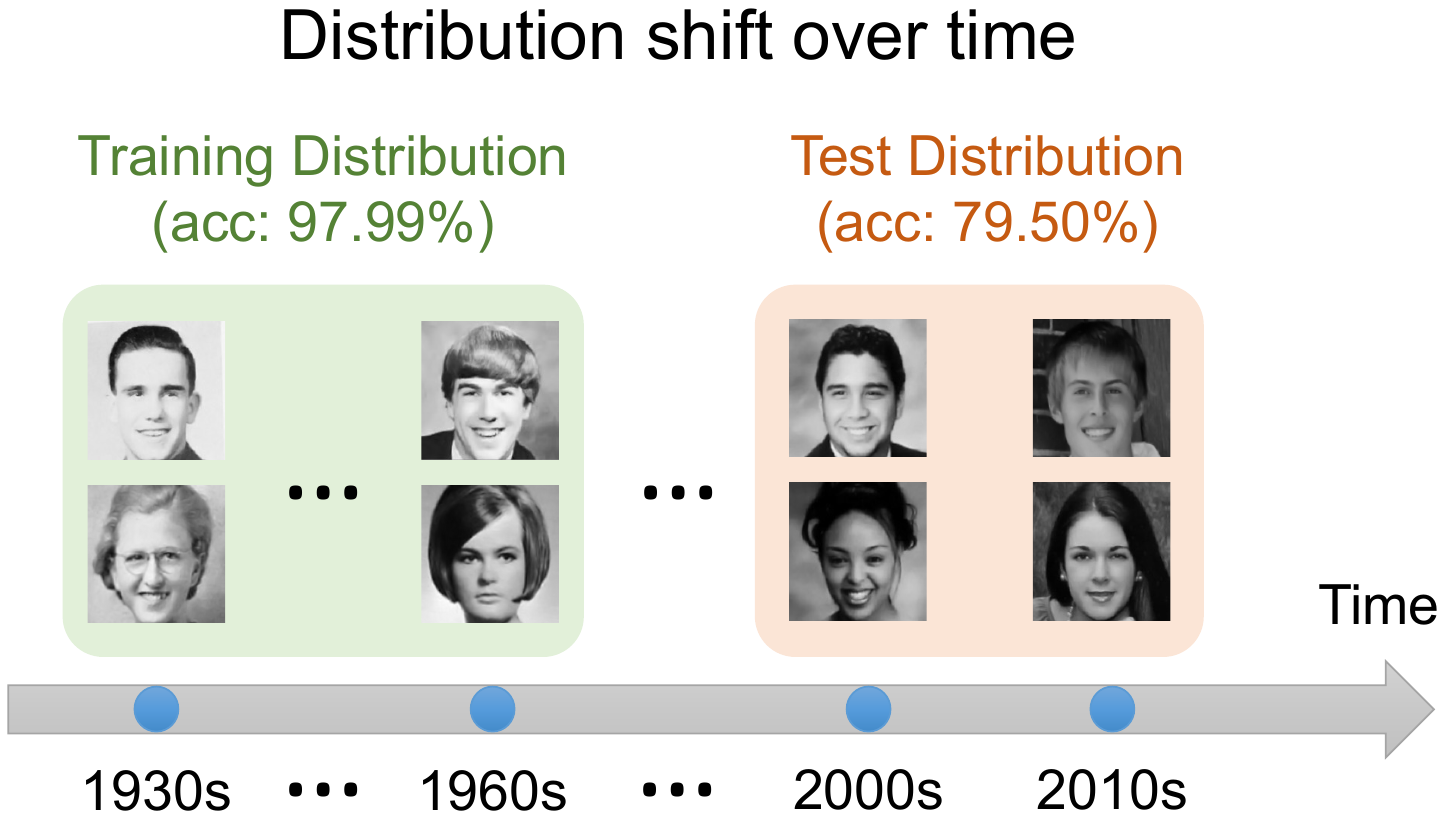}
\caption{An illustration of temporal distribution shift on Yearbook. In Wild-Time, we split the train and test sets by timestamp and observe performance drops between train and test distributions.}
\vspace{-0.2cm}
\label{fig:temporal_distribution_shift}
\end{wrapfigure}

Distribution shift occurs when the test distribution differs from the training distribution. \emph{Temporal shifts} -- distribution shifts that arise from the passage of time -- are a common type of distribution shift.
Due to non-stationarity, production (i.e. test) data shifts over time \cite{huyen2022designing}.
This degrades the performance of machine learning systems deployed in the real world.
For example, \citet{lazaridou2021mind} found that neural language models perform worse when predicting future utterances from beyond their training period, and that their performance worsens with time. As another example, flu incidence prediction from Internet search queries performed remarkably well in 2008~\cite{ginsberg2009detecting}. However, using the same model in 2013 incorrectly predicted double the incidence~\cite{butler2013google}. Finally, in Figure \ref{fig:temporal_distribution_shift}, the style of yearbook portraits of American high schoolers \cite{ginosar2015century} change over the decades. As a result, models trained on earlier years and evaluated on future years suffer substantial drops in performance. 

Though temporal shifts are ubiquitous in real-world scenarios, they remain understudied. Prior benchmarks for out-of-distribution robustness in the wild focus on domain shifts and subpopulation shifts \cite{koh2021wilds,malinin2021shifts,ye2022ood,sagawa2022extending}. 
Many popular benchmarks that feature a stream of data, such as those used in continual learning \cite{adel2019continual,chaudhry2018riemannian,kirkpatrick2017overcoming,schwarz2018progress,zenke2017continual,chaudhry2018efficient,lopez2017gradient,rebuffi2017icarl,shin2017continual}, contain a manually delineated set of tasks and artificial sequential variations, which are not representative of natural temporal shifts. These include small-image sequences with disparate label splits (e.g., Split TinyImageNet~\cite{le2015tiny}, Split CIFAR~\cite{krizhevsky2009learning}), different kinds of image transformations to MNIST digits (e.g., Rainbow MNIST \cite{finn2019online}), or different visual recognition targets \cite{li2019learn} (cf. Section \ref{sec6}.) Recent works have investigated natural temporal distribution shifts in different domains such as drug discovery \cite{huang2021therapeutics}, visual recognition \cite{cai2021online}, and sepsis prediction \cite{guo2022evaluation} and created datasets in each of these domains. However, there does not exist a systematic study of real-world temporal distribution shifts and a benchmark spanning various domains. Here, we curate a collection of these datasets and create a benchmark that allows researchers to easily evaluate their methods across multiple domains.

\begin{figure*}[t]
\centering
  \includegraphics[width=0.9\textwidth]{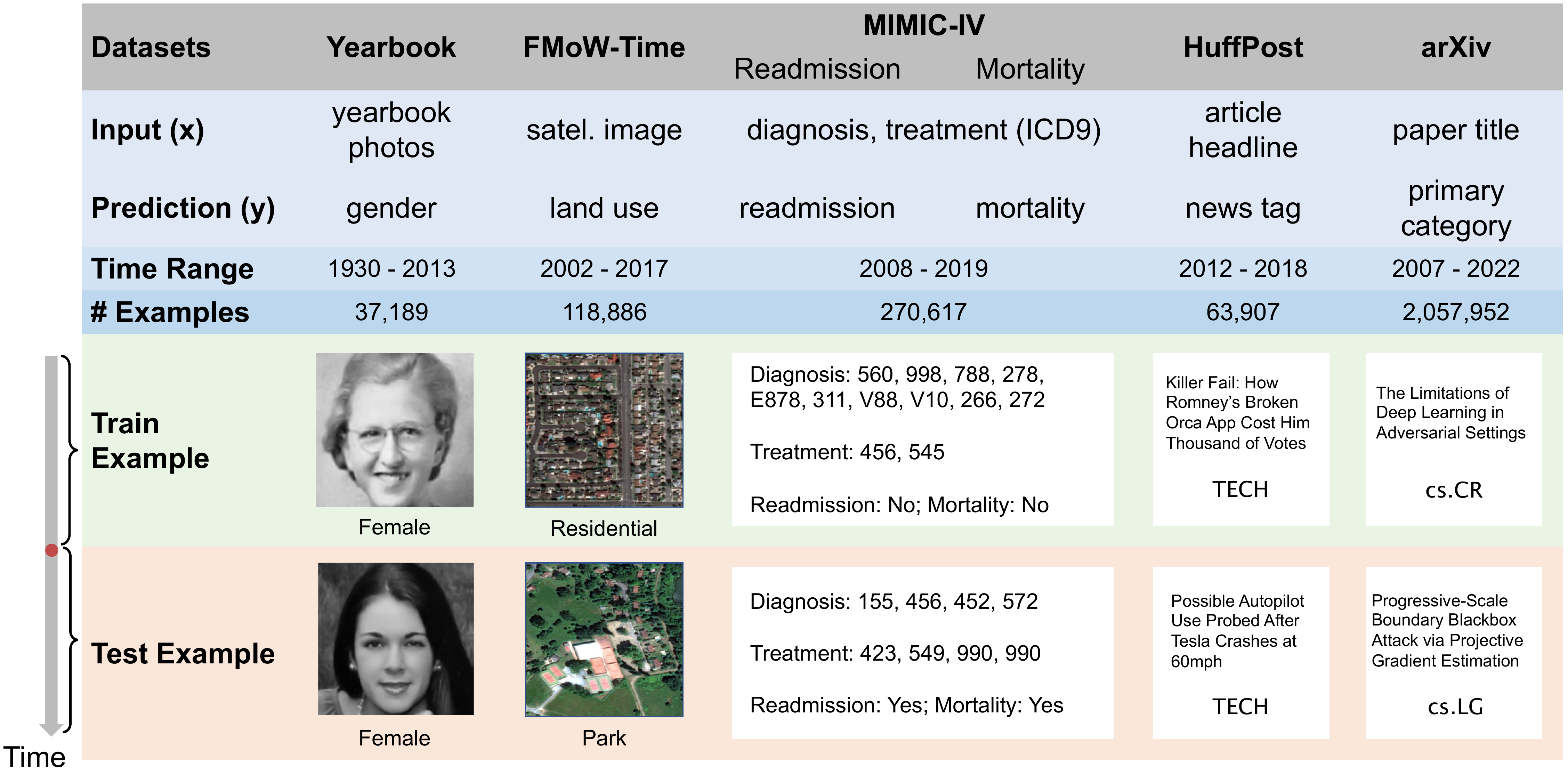}
  \caption{The Wild-Time benchmark includes a collection of 5 datasets with 6 tasks, from \cite{ginosar2015century}, \cite{mimiciv}, \cite{huffpostdatasetbook}, \cite{clement2019arxiv}. For each task, we train models on the past and evaluate it in the future. We list the input, prediction, time range and the number of examples for each task.}\label{fig:benchmark_list}
  \vspace{-2em}
\end{figure*}
This paper presents \textbf{Wild-Time} (``in-the-Wild distribution shifts over Time"), a benchmark of in-the-wild gradual temporal distribution shifts together with two comprehensive evaluation protocols. In Wild-Time, we investigate real-world temporal distribution shifts across a diverse set of tasks (Figure~\ref{fig:benchmark_list}), including portraits classification~\cite{ginosar2015century}, ICU patient readmission prediction~\cite{mimiciv}, ICU patient mortality prediction~\cite{mimiciv}, news tag classification~\cite{huffpostdatasetbook}, and article category classification~\cite{clement2019arxiv}. The distribution shifts in these applications happen naturally due to the passage of time, which the datasets reflect through changing fashion and social norms~\cite{ginosar2015century}, atmospheric conditions \cite{malinin2021shifts}, and current events \cite{huffpostdataset, huffpostdatasetbook}. We propose two evaluation strategies for Wild-Time: evaluation with a fixed time split (Eval-Fix) and with a data stream (Eval-Stream).

\revision{On these datasets, we evaluate several representative approaches in continual learning, invariant learning, self-supervised learning, and ensemble learning.} We extend invariant learning methods to the temporal distribution shift setting. While prior invariant learning approaches are trained on clearly delineated sets of distributions, we consider a stream of unsegmented observations, where domain labels are not provided. To extend domain invariant methods to the temporal distribution shift setting, we construct domains to be different windows of time. More specifically, all temporal windows of a certain window size are treated as a domain, allowing us to directly apply invariant learning approaches over the constructed domains.

The main conclusions of our benchmark analysis are that invariant learning, self-supervised learning, and continual learning approaches do not show substantial improvements compared to standard ERM training. To make the Wild-Time datasets accessible for future research, we released a Python package that automates data loading and baseline training at \href{https://wild-time.github.io/}{https://wild-time.github.io/}. We hope that Wild-Time will accelerate the development of temporally robust models.

\section{Problem and Evaluation Settings}
\label{sec:pre}
We define the temporal robustness setting. Following \cite{koh2021wilds}, we view the entire data distribution as a mixture of $T$ timestamps $\mathcal{T}=\{1,\ldots,T\}$. Each timestamp $t$ is associated with a data distribution $P_t$ over ($x$, $y$), where $x$ and $y$ represent input features and labels, respectively, and all examples are sampled from the data distribution $P_t$. To formulate the temporal distribution shift setting, we define the training distribution as $P^{tr}=\sum_{t=1}^T \lambda^{tr}_t P_t$, and the test distribution as $P^{ts}=\sum_{t=1}^T \lambda^{ts}_t P_t$. Note that, here, timestamp differs from the notion of ``domain" used in other works on distribution shift \cite{ganin2016domain,li2018domain,arjovsky2019invariant,ahuja2021invariance,koh2021wilds,yao2022improving}. In the temporal shift setting, we do not require distribution shift between consecutive timestamps, i.e., we can have $P_t=P_{t-1}$. Based on the problem setting, we will detail the criteria to select datasets and the evaluation strategies in Wild-Time.

\subsection{Criteria for Dataset Selection}
\label{sec:criteria}
In Wild-Time, we select datasets using three criteria:
\begin{itemize}[leftmargin=*]
    \item \textbf{Naturally Occurring Temporal Shifts.} We select real-world datasets that consist of data collected over time and contain timestamp metadata. We select datasets for which it is natural to train on the past and test into the future, and we include datasets from a diverse collection of domains, including vision, healthcare, and language modeling.
    
    \item \textbf{Temporal Distribution Shifts with Performance Drops.} We require that there is substantial performance degradation between the training and test splits, i.e., we observe large drops in performance between the in-distribution and out-of-distribution times.
    
    \item \textbf{Gradual Temporal Distribution Shifts.} Sudden shifts are well-represented by existing benchmarks on domain shift and subpopulation shift. Models can more effectively extrapolate temporal correlations when the distribution shifts occur gradually over time, as opposed to sudden shifts. Thus, in this paper, we focus on gradual temporal distribution shifts, where we require gradual performance drops between consecutive periods of time. 
\end{itemize}

\subsection{Evaluation Strategies}

Before presenting the datasets, we first discuss two evaluation strategies in Wild-Time.

\textbf{Evaluation with a fixed time split (Eval-Fix).} Eval-Fix evaluates models on a single, fixed train-test time split and offers a simple and quick evaluation protocol. Eval-Fix is the primary evaluation strategy in Wild-Time. Concretely, we denote the split timestamp as $t_s$. The train and test sets are $\mathcal{T}^{tr}=\{t \le t_{s}| \forall t\}$, $\mathcal{T}^{ts}=\{t > t_{s}| \forall t\}$, respectively. Eval-Fix evaluates performance using two metrics, average and worst-time performance ($\mathrm{Avg}$) and worst-time performance ($\mathrm{Worst}$). Specifically, let $R_t$ denote the performance at each timestamp $t$. We define the average performance ($ \mathrm{Avg}$) and worst-time performance ($\mathrm{Worst}$) as
\begin{equation}
\small
    \mathrm{Avg}=\frac{1}{|\mathcal{T}^{ts}|}\sum_{t\in \mathcal{T}^{ts}}R_{t},\; \mathrm{Worst}=\min_{t\in \mathcal{T}^{ts}}R_{t}.
\end{equation}
\revision{Here, average performance measures the overall out-of-distribution performance. Worst-time performance evaluates the model's robustness over time.}

\textbf{Evaluation with data stream (Eval-Stream).} Eval-Stream evaluates models at each timestamp, evaluating average and worst-time performance on the next $K$ timestamps. Eval-Stream mimics standard machine learning development pipelines, where models are updated frequently and evaluated on timestamps in the near future.

Specifically, we construct a performance matrix $\mathcal{R}\in \mathbb{R}^{T\times K}$, where each element $R_{i,j}$ is the test accuracy of the model trained on timestamp $t_i$ and evaluated on timestamp $t_{j}$. Following~\cite{lopez2017gradient}, we define the average performance ($\mathrm{Avg_{stream}}$) and worst-time performance ($\mathrm{Worst_{stream}}$) as
\begin{equation}
\small
    \mathrm{Avg_{stream}}=\frac{1}{|\mathcal{T}|K}\sum_{t\in \mathcal{T}}\sum_{j=t+1}^{t+K}R_{i,j},\; \mathrm{Worst_{stream}}=\frac{1}{|\mathcal{T}|}\sum_{t\in \mathcal{T}}\min_{j\in\{t+1\ldots t+K\}}R_{i,j}
\end{equation}
Here, $K$ is a hyperparameter. Compared with typical continual learning metrics in \citet{lopez2017gradient}, we evaluate performance on the next few timestamps, rather than just the subsequent timestamp, to assess the model's robustness across time.

\section{Datasets}
In this section, we briefly discuss the datasets and tasks included in Wild-Time, which reflect natural gradual temporal distribution shifts. We provide more detailed descriptions of all datasets in Appendix~\ref{sec:app_data}. Additionally, in Appendix~\ref{sec:app_additional_data}, we discuss some datasets that violate our criteria of dataset selection discussed in Section~\ref{sec:criteria}, e.g., datasets with sudden temporal distribution shifts.

\textbf{Yearbook (Appendix~\ref{app:data:yearbook}).}  
    Social norms, fashion styles, and population demographics change over time. This is captured in the Yearbook dataset, which consists of 37,921 frontal-facing American high school yearbook photos \cite{ginosar2015century}. We exclude portraits from $1905-1929$ due to the limited number of examples in these years, resulting in 33,431 examples from $1930-2013$. Each photo is a $32\times 32\times 1$ grey-scale image associated with a binary label $y$, which represents the student's gender. In Eval-Fix, the training set consists of data from before $1970$, and the test set comprises data after $1970$, which corresponds to $40$ and $30$ years, respectively.

\textbf{FMoW-Time (Appendix~\ref{app:data:fmow}).} Machine learning models can be used to analyze satellite imagery and aid humanitarian and policy efforts by monitoring croplands \cite{jia2019recurrent} and predicting crop yield \cite{sharifi2021yield} and poverty levels \cite{jean2016combining}. Due to human activity, satellite imagery changes over time, requiring models that are robust to temporal distribution shifts.

We study this problem on the Functional Map of the World (FMoW) dataset \cite{christie2018functional}, adapted from the WILDS benchmark \cite{koh2021wilds} and is named as FMoW-Time. Given a satellite image, the task is to predict the type of land usage. The FMoW-Time dataset~\cite{koh2021wilds} consists of 141,696 examples from $2002 - 2017$. Each input $x$ is a $224 \times 224$ RGB satellite image, and the corresponding label $y$ is one of $62$ land use categories. We use the train/val/test splits in WILDS to construct FMoW-Time dataset in Wild-Time. The train/val/test data splits from WILDS contain images from disjoint location coordinates, and all splits
contain data from all 5 geographic regions. In Eval-Fix, the training set includes data from $2002 - 2015$, and the test set includes data from $2016 - 2017$.

\textbf{MIMIC-IV (Appendix~\ref{app:data:mimic}).} Many machine learning healthcare applications have emerged in the last decade, such as predicting disease risk \cite{ma2017dipole}, medication changes \cite{yang2021change}, patient subtyping \cite{baytas2017patient}, in-hospital mortality \cite{guo2022evaluation}, and length of hospital stay \cite{ettema2010prediction}. \revision{However, changes in healthcare over time, such as the emergence of new treatments and changes in patient demographics, are an obstacle in deploying machine learning-based clinical decision support systems \cite{guo2022evaluation}.}

We study this problem on MIMIC-IV, one of the largest public healthcare datasets that comprises abundant medical records of over 40,000 patients. In MIMIC-IV, we treat each admission as one record, resulting in 216,487 healthcare records from $2008 - 2019$. To protect patient privacy, the reported admission year is in a three year long date range. Hence, our timestamps are groups of three years: $2008-2010$, $2011-2013$, $2014-2016$, $2017-2019$. We consider two classification tasks:
\begin{itemize}[leftmargin=*]
    \item \textbf{MIMIC-Readmission} aims to predict the risk of being readmitted to the hospital within 15 days.
    \item \textbf{MIMIC-Mortality} aims to predict in-hospital mortality for each patient.
\end{itemize}
For each record, we concatenate the corresponding ICD9 codes~\cite{world2004international} of diagnosis and treatment. A binary indicator is used to indicate the codes are come from diagnosis or treatment. We use the concatenated one as the input feature. The label is a binary value that indicates whether the patient is readmitted or passed away for MIMIC-Readmission and MIMIC-Mortality, respectively. For the Eval-Fix setting, the train set consists of patient data fom $2008 - 2013$, while the test set consists of data from $2014 - 2020$.

\textbf{Huffpost (Appendix~\ref{app:data:huffpost}).} In many language models which deal with information correlated with time, temporal distribution shifts cause performance degradation in downstream tasks such as Twitter hashtag classification~\cite{jin2021lifelong} or question-answering systems~\cite{lin2022continual}. Performance drops across time reflect changes in the style or content of current events.

We study this temporal shift on the Huffpost dataset \cite{huffpostdatasetbook}. The task is to identify tags of news articles from their headlines. Each input feature $x$ is a news headline, and the output $y$ is the news category. We only keep categories that appear in all years from $2012 - 2022$, resulting 11 categories in total. We choose year $2015$ as the split timestamp in Eval-Fix.

\textbf{arXiv (Appendix~\ref{app:data:arxiv}).} Due to the evolution of research fields, the style of arXiv pre-prints also changes over time, reflected by the change in article categories. For example, ``neural network attack" was originally a popular keyword in the security community, but gradually became more prevalent in the machine learning community. We study this temporal shift in the arXiv dataset~\cite{clement2019use}, where the task is to predict the primary category of arXiv pre-prints given the paper title as input. The entire dataset includes 172 pre-print categories from $2007 - 2022$.
\section{Baselines for Temporal Distribution Shifts} \label{sec:4}
\vspace{-0.3em}
Many algorithms have been proposed to improve a model's robustness to distribution shifts or improve a model's performance on a stream of data.
For our evaluation, we choose several representative methods from five main categories: classical supervised learning (empirical risk minimization), continual learning, invariant learning, self-supervised learning, and ensemble learning. These methods have been successful on domain generalization and continual learning benchmarks. We extend the selected invariant learning approaches to the temporal distribution shift setting. See Appendix~\ref{sec:app_algorithm} for a detailed discussion of these algorithms and how we apply them to our tasks.


\textbf{Classical Supervised Learning (Appendix~\ref{sec:app_method_classical}).} We evaluate the performance of empirical risk minimization (ERM) on all tasks. In Eval-Fix, we directly train a machine learning model with ERM. In Eval-Stream, we apply ERM to every timestamp and report the performance.


\textbf{Continual Learning (Appendix~\ref{sec:app_method_continual}).} Continual learning, also known as lifelong learning or incremental learning, aims to effectively learn from non-stationary distributions via a stream of data \cite{adel2019continual,chaudhry2018riemannian,kirkpatrick2017overcoming,schwarz2018progress,zenke2017continual,chaudhry2018efficient,lopez2017gradient,rebuffi2017icarl,shin2017continual}. The goal is to accumulate and reuse knowledge in future learning without forgetting information needed for previous tasks, a phenomenon known as catastrophic forgetting \cite{kirkpatrick2017overcoming}, which may enable such models to robustly extrapolate into the future in the temporal shift setting. We evaluate four representative algorithms, including fine-tuning, regularization-based (EWC, SI) and memory-based (A-GEM) methods. These methods have been successful on several continual learning benchmarks, such as permuted MNIST \cite{goodfellow2013empirical} and Split TinyImagenet \cite{le2015tiny}.

\textbf{Temporally Invariant Learning (Appendix~\ref{sec:app_method_invariant}).} Invariant learning methods learn representations or predictors that are invariant across different domains. Common approaches include aligning feature representation over different domains \cite{ganin2016domain,long2015learning,sun2016deep,tzeng2014deep,xu2020adversarial,yue2019domain,zhou2020deep}, learning invariant predictors via selective augmentation~\cite{yao2022improving} or by strengthening the correlations between representations and labels~\cite{ahuja2021invariance,arjovsky2019invariant,khezeli2021invariance,krueger2021out}, and optimizing worst-group performance~\cite{sagawa2019distributionally,zhang2020coping,zhou2021examining}. In Wild-Time, we select four representative invariant learning methods: CORAL \cite{sun2016deep}, IRM \cite{arjovsky2019invariant}, LISA \cite{yao2022improving}, and GroupDRO \cite{sagawa2019distributionally}. We adapt these methods to a temporal setting and train them incrementally at each timestamp $t$. In the following, we discuss how we adapt these approaches to the temporal shift setting. 

As mentioned in Section \ref{sec:pre}, in the temporal shift setting, the ``timestamp" information is not the same as the notion of a ``domain," as the distribution shift may not occur between consecutive timestamps. The data streams in our benchmark are unsegmented and do not include domain boundaries. This setting poses new challenges to the above invariant learning approaches, which rely on domain labels. 

\begin{wrapfigure}{r}{0.42\textwidth}
\vspace{-1.5em}
\centering\includegraphics[width=0.4\textwidth]{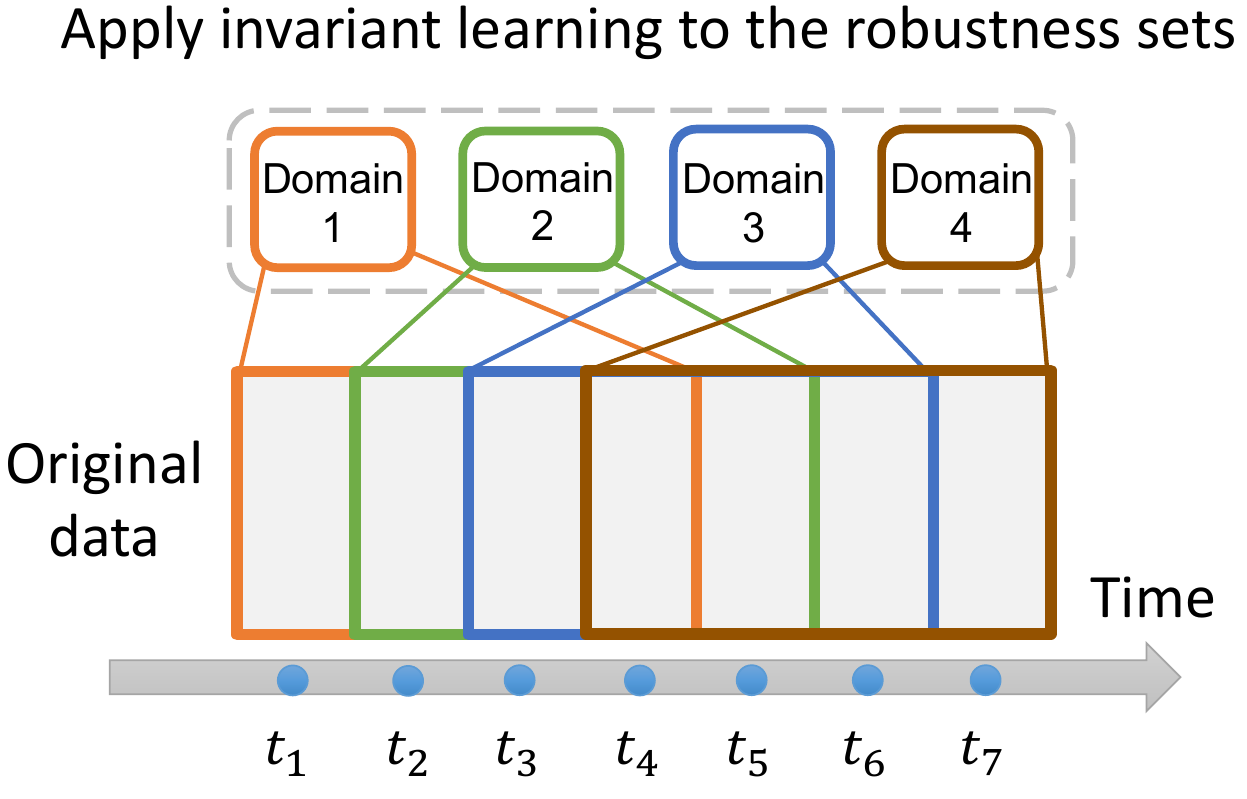}
\vspace{-0.5em}
\caption{Construction of a temporal robustness set. Here, we have $T = 7$ and $L = 3$. By applying sliding window-based segmentation, we obtain $4$ substreams of data. We apply invariant learning approaches to this robustness set.}
\vspace{-2em}
\label{fig:temporal_adaptation}
\end{wrapfigure}
To address this challenge, we adapt the above invariant learning approaches to the temporal distribution shift setting. We leverage timestamp metadata to create a temporal robustness set consisting of substreams of data, where each substream is treated as one domain. Specifically, as shown in Figure~\ref{fig:temporal_adaptation}, we define a sliding window $\mathcal{G}$ with length $L$. For a data stream with $T$ timestamps, we apply the sliding window $\mathcal{G}$ to obtain $T-L+1$ substreams. We treat each substream as a ``domain" and apply the above invariant algorithms on the robustness set. We name the adapted CORAL, GroupDRO and IRM as CORAL-T, GroupDRO-T, IRM-T, respectively. Note that we do not adapt LISA since the intra-label LISA performs well without domain information, which is also mentioned in the original paper. 

See Appendix~\ref{sec:app_temporal_invariant_method} for details on the temporal adaptation of these invariant learning algorithms and additional ablations. In particular, we compare vanilla invariant learning algorithms with their temporally adapted versions, assess the effect of the length of the time window $L$, and compare performance using overlapping versus non-overlapping time windows.

\textbf{Self-Supervised Learning (Appendix~\ref{sec:app_method_ssl}).}  Self-supervised learning has been shown to improve out-of-distribution robustness ~\cite{ji2021power,shen2022connect}. Here, we introduce two representative self-supervised learning methods, SimCLR \cite{chen2020simple} and SwaV \cite{caron2020unsupervised}, and evaluate their performance on the Wild-Time image classification datasets, Yearbook and FMoW-Time.

\textbf{Ensemble Learning (Appendix~\ref{sec:app_method_bayesian}).} Typically, using ensemble learning improves the performance of machine learning models. We introduce Stochastic Weighted Averaging (SWA) as the representative ensemble learning approach, which is an approximate Bayesian method which averages multiple parameter values along the trajectory of stochastic gradient descent~\cite{izmailov2018averaging}.

\section{Experiments}
\vspace{-0.5em}
We benchmark the performance of all methods in Section \ref{sec:4} on each dataset in Wild-Time. Each baseline is evaluated using both the Eval-Fix and Eval-Stream settings. See Appendix~\ref{sec:app_eval_stream_results} for all evaluation results and experimental details under the Eval-Stream setting.

\subsection{Experimental Setup}
\textbf{Data Split.} For both Eval-Fix and Eval-Stream, the training and test sets are subsets of the entire dataset such that the training timestamps are earlier than the test timestamps. We measure temporal out-of-distribution (OOD) robustness as performance on the test set. To compare out-of-distribution with in-distribution (ID) performance, we measure the average per-timestep performance on a held-out set of 10\% training examples (20\% for MIMIC-Mortality and MIMIC-Readmission) from each training ID timestamp. More details of our data split protocols are described in Appendix~\ref{sec:app_eval_fix_split}.

\textbf{Evaluation Metrics.} We measure accuracy in most classification tasks, including Yearbook, FMoW-Time, MIMIC-Readmission, HuffPost, and arXiv. For the MIMIC-Mortality task, we use ROC-AUC due to label imbalance.

\textbf{Hyperparameter Settings.}
For each dataset, we use the same backbone for all baselines.The choice of backbones are based on the original paper (e.g., DenseNet101~\cite{huang2017densely} for FMoW-Time~\cite{koh2021wilds}, 
or the commonly used ones (e.g., DistilBERT~\cite{sanh2019distilbert} for arXiv and Huffpost). For each method, we tune hyperparameters using cross-validation \revision{with grid search}. In Eval-Fix, we hold out 10\% of the data of each training timestamp (20\% for MIMIC-Readmission, and MIMIC-Mortality) to construct the validation set for hyperparameter tuning. Here, we use examples from the remaining 90\% of the data to train the model and evaluate the performance on the corresponding validation set. We repeat this process three times via cross-validation with different held-out 10\% of the data. After selecting all hyperparameters, we use the entire training set to train the model. See Appendix~\ref{sec:app_set_split} for a detailed hyperparameter search setting.

\subsection{Performance Drops from Temporal Distribution Shifts}
The Wild-Time datasets should exhibit observable performance drops between training and test times. In this section, we demonstrate this for every Wild-Time dataset. Table~\ref{tab:gap} shows the ID and OOD performance of ERM on each Wild-Time dataset. (See Table~\ref{tab:results_offline} in the Appendix for comprehensive results on all remaining baselines.)

We observe that OOD performance is substantially lower than ID performance. We conduct further experiments, in which we train ERM on both ID and OOD examples, in Appendix~\ref{sec:split_compare}. The substantial drop in ID versus OOD performance indicates that the performance drop is caused by distribution shift rather than the difficulty of training timestamps.

\begin{table*}[t]
\small
\caption{The in-distribution versus out-of-distribution test performance evaluated on Wild-Time under the Eval-Fix setting. For each dataset, higher value means better performance.}
\vspace{-1em}
\label{tab:gap}
\begin{center}
\begin{tabular}{l|c|cc}
\toprule
Dataset (Metric) & In-distribution & Out-of-distribution\\\midrule
Yearbook (Acc) & 97.99 (1.40) & 79.50 (6.23)  \\
\revision{FMoW-Time (Acc)}  & 58.07 (0.15) & 54.07 (0.25) \\
MIMIC-Readmission (Acc) & 73.00 (2.94) & 61.33 (3.45) \\
MIMIC-Mortality (AUC) & 90.89 (0.59) & 72.89 (8.96) \\
Huffpost (Acc)  & 79.40 (0.05) & 70.42 (1.15) \\
arXiv (Acc)  & 53.78 (0.16) & 45.94 (0.97) \\
\bottomrule
\end{tabular}
\vspace{-1.5em}
\end{center}
\end{table*}

\subsection{Baseline Comparison} 

\begin{figure}[t]
\vspace{-1em}
\centering
\begin{subfigure}[c]{0.31\textwidth}
		\centering
\includegraphics[width=\textwidth]{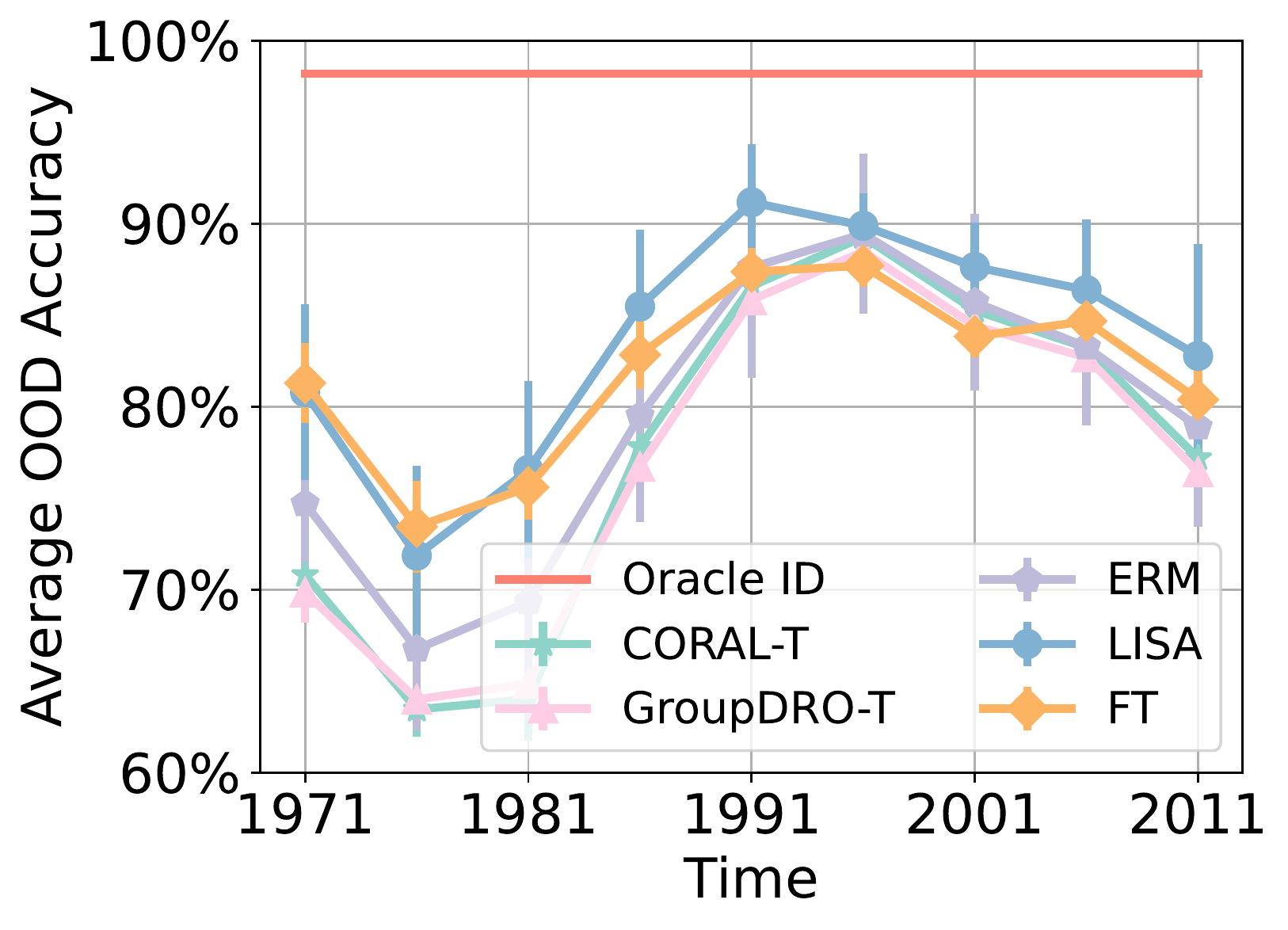}
    \caption{\label{fig:yearbook}: Yearbook}
\end{subfigure}
\begin{subfigure}[c]{0.31\textwidth}
		\centering
\includegraphics[width=\textwidth]{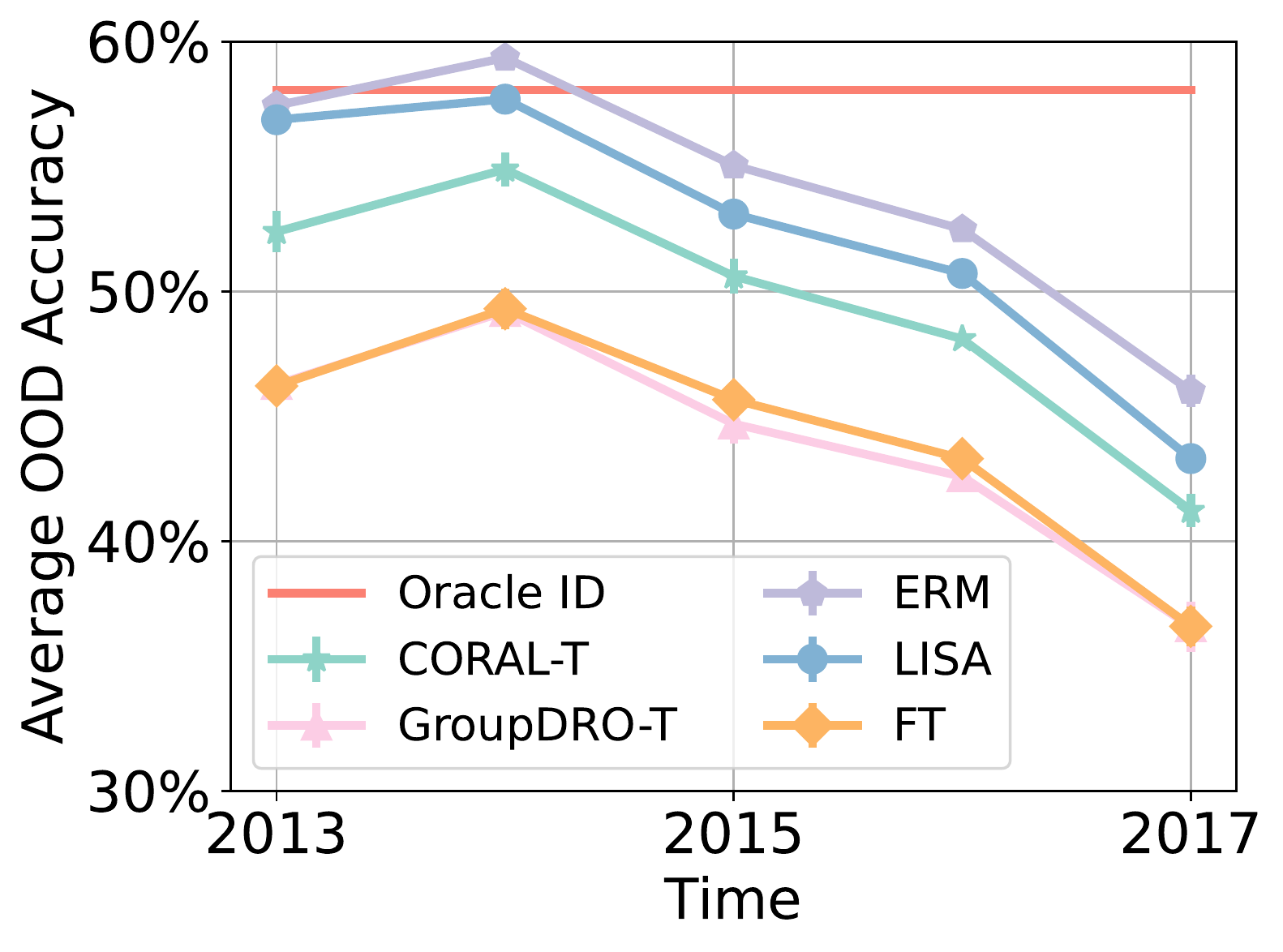}
    \caption{\label{fig:FMoW}: FMoW-Time}
\end{subfigure}
\begin{subfigure}[c]{0.31\textwidth}
		\centering
\includegraphics[width=\textwidth]{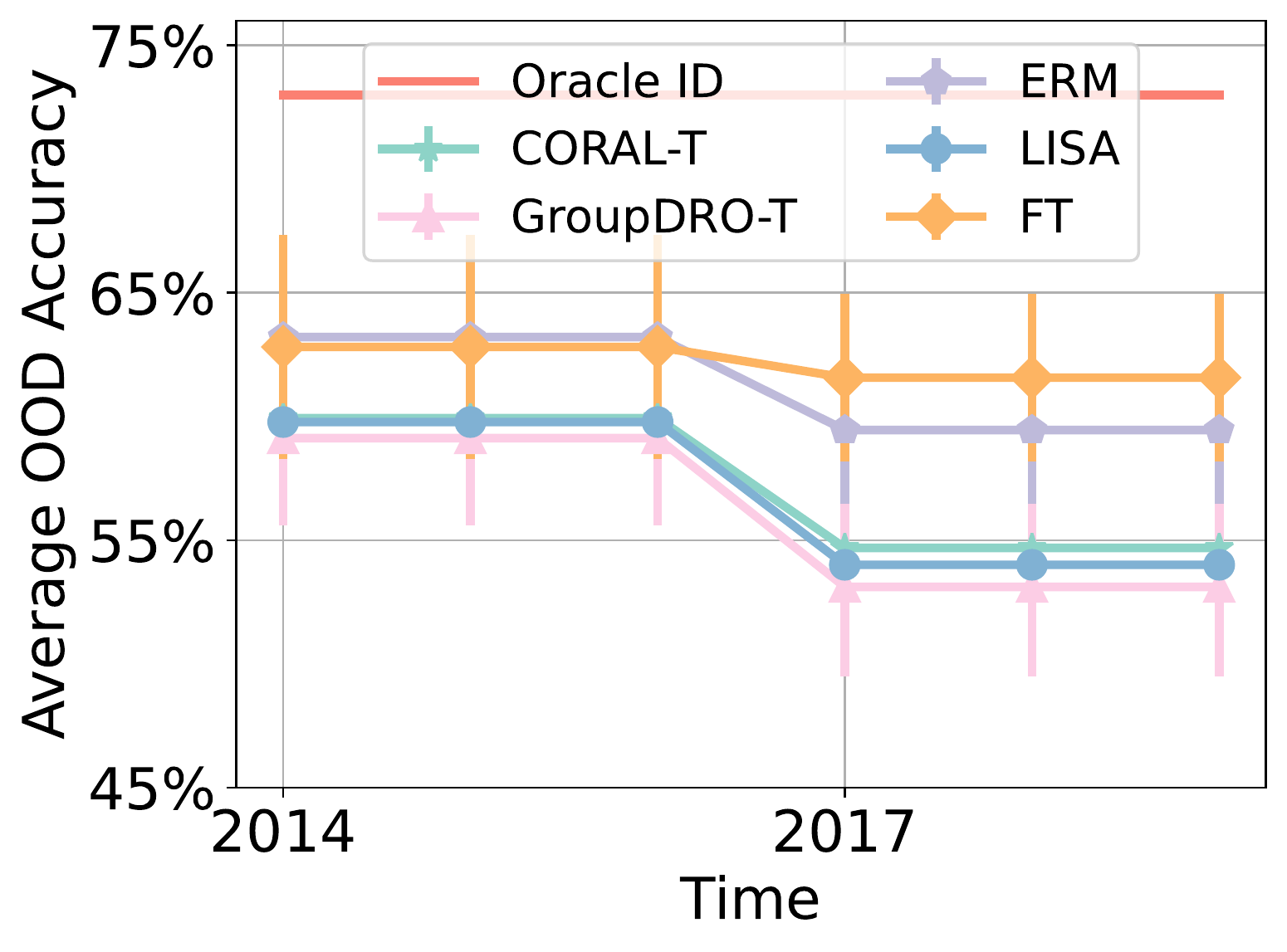}
    \caption{\label{fig:MIMIC-RD}: MIMIC-Readmission}
\end{subfigure}
\begin{subfigure}[c]{0.31\textwidth}
		\centering
\includegraphics[width=\textwidth]{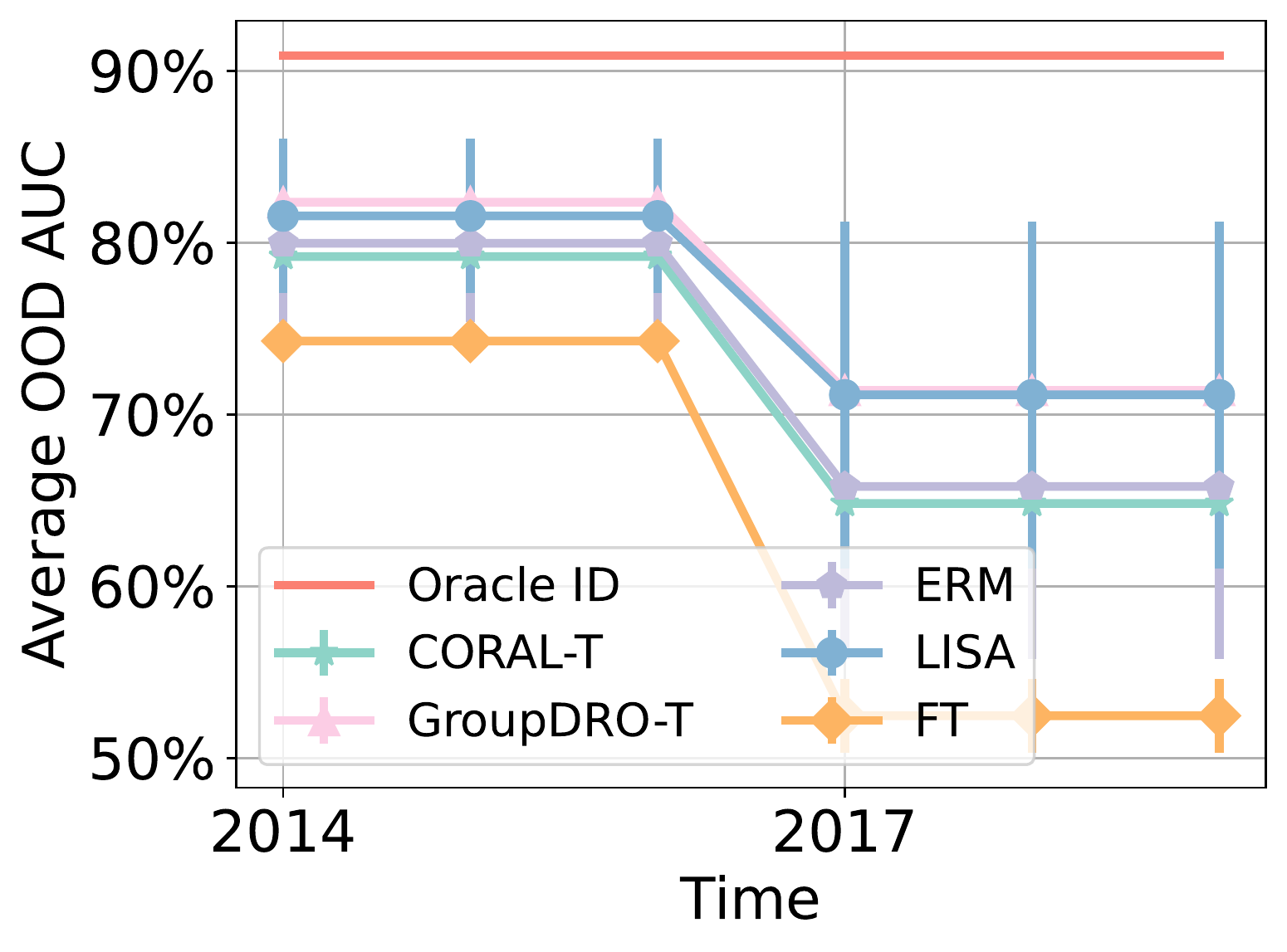}
    \caption{\label{fig:MIMIC-Mort}: MIMIC-Mortality}
\end{subfigure}
\begin{subfigure}[c]{0.31\textwidth}
		\centering
\includegraphics[width=\textwidth]{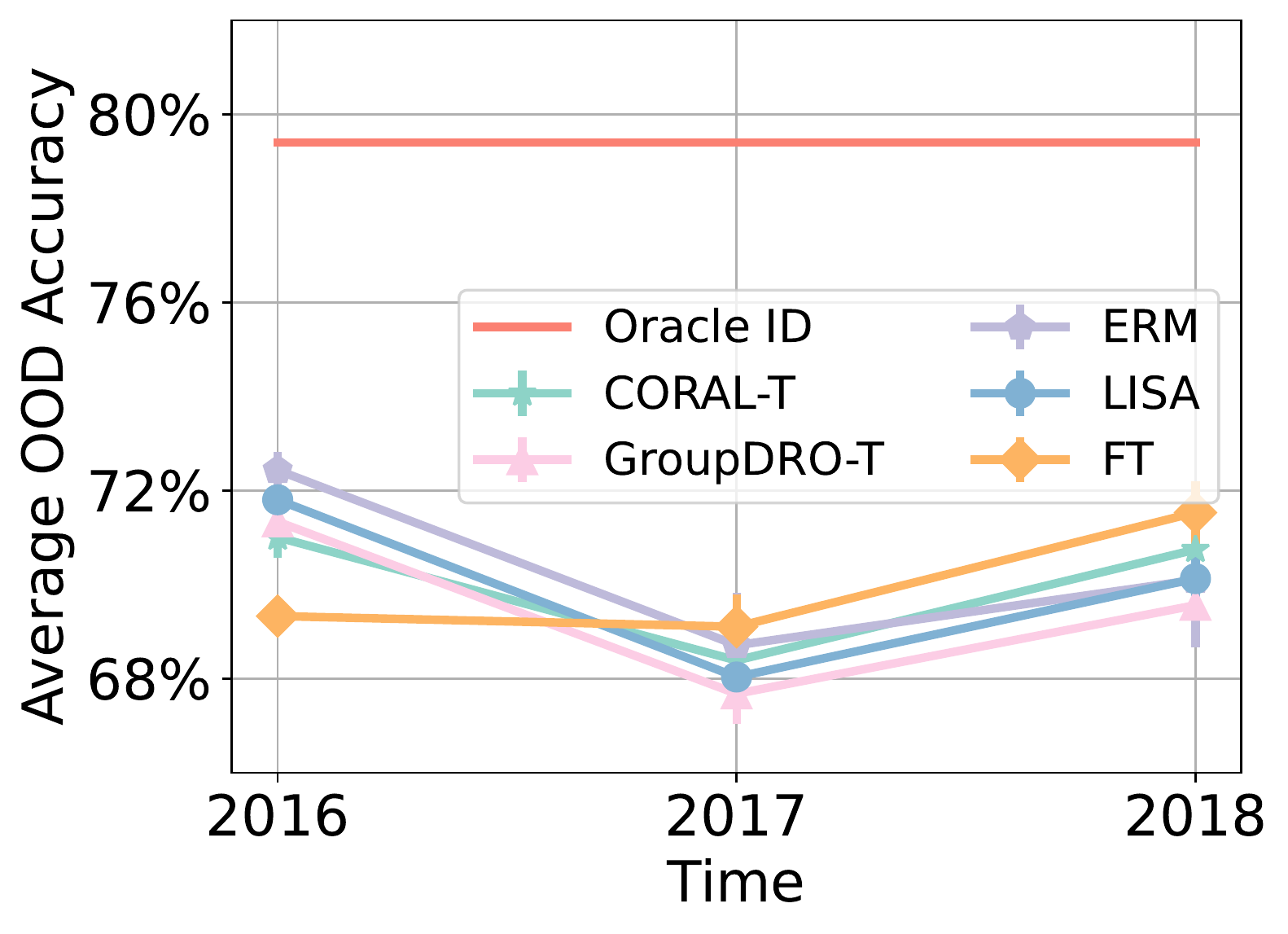}
    \caption{\label{fig:huffpost}: HuffPost}
\end{subfigure}
\begin{subfigure}[c]{0.31\textwidth}
		\centering
\includegraphics[width=\textwidth]{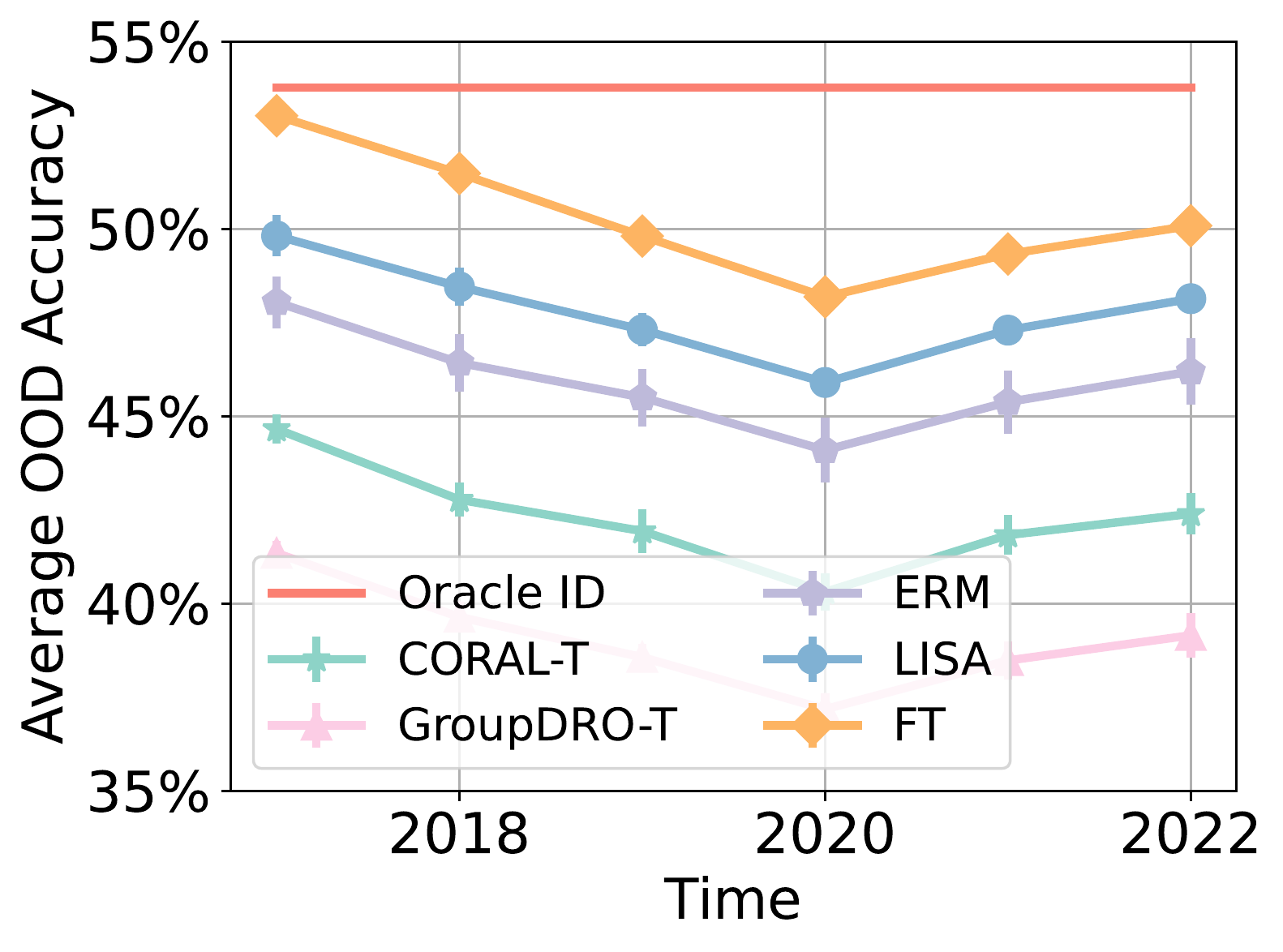}
    \caption{\label{fig:arxiv}: arXiv}
\end{subfigure}
\caption{Out-of-distribution performance per test timestamp. We select five representative baselines -- ERM, FT (Fine-tuning), CORAL-T, GroupDRO-T, LISA, and show the corresponding performance. Oracle ID represent the best ID performance over all compared baselines. Note that for MIMIC-Readmission and MIMIC-Mortality, our OOD timestamps are the three-year blocks $2014-2016$, $2017-2019$. Hence in Figures 4(c) and 4(d), the performance over this three-year block is the same.}
\vspace{-1em}
\label{fig:ood_per_step}
\end{figure}

\begin{table*}[t]
\small
\caption{The out-of-distribution test performance of each method evaluated on Wild-Time under the Eval-Fix setting. Different groups of rows correspond to different categories of methods. Full table with standard deviation are computed over three random seeds and reported in Table~\ref{tab:results_offline} of Appendix. We bold the best OOD performance for each dataset.}
\vspace{-1em}
\label{tab:results_offline_main}
\begin{center}
\begin{tabular}{l|cc|cc|cc}
\toprule
\multirow{3}{*}{} & \multicolumn{2}{c|}{Yearbook} & \multicolumn{2}{c|}{FMoW-Time} & \multicolumn{2}{c}{MIMIC-Readmission} \\
& \multicolumn{2}{c|}{(Accuracy (\%) $\uparrow$)} & \multicolumn{2}{c|}{(Accuracy (\%) $\uparrow$)} & \multicolumn{2}{c}{(Accuracy (\%) $\uparrow$)}\\
& Avg. &  Worst & Avg. &  Worst & Avg. &  Worst\\\midrule
Fine-tuning & 81.98 & \textbf{69.62} & 44.25 & 37.14 & 62.19 & 59.57 \\
EWC  & 80.07 & 66.61 & 44.02 & 36.42 & \textbf{66.40} & \textbf{64.69} \\
SI & 78.70 & 65.18 & 44.25 & 37.14 & 62.60 & 61.13 \\
A-GEM & 81.04 & 67.07 & 44.10 & 36.02 & 63.95 & 62.66 \\\midrule
ERM & 79.50 & 63.09 & \textbf{54.07} & \textbf{46.00} & 61.33 & 59.46 \\
GroupDRO-T & 77.06 & 60.96 & 43.87 & 36.60 & 56.12 & 53.12 \\
mixup & 76.72 & 58.70 & 53.67 & 44.57 & 58.82 & 57.30 \\
LISA & 83.65 & 68.53 & 52.33 & 43.30 & 56.90 & 54.01 \\
CORAL-T & 77.53 & 59.34 & 49.43 & 41.23 & 57.31 & 54.69 \\
IRM-T & 80.46 & 64.42 & 45.00 & 37.67 & 56.53 & 52.67 \\\midrule
\revision{SimCLR} & \revision{78.59} & \revision{60.15} & 44.76 & 37.00 & \revision{n/a} & \revision{n/a} \\
\revision{SwaV} & \revision{78.38} & \revision{60.73} & 44.92 & 37.17 & \revision{n/a} & \revision{n/a} \\\midrule
\revision{SWA} & \revision{\textbf{84.25}} & \revision{67.90} & \textbf{54.06} & \textbf{46.01} & 59.10 & 56.54 \\\midrule\midrule
\multirow{3}{*}{} & \multicolumn{2}{c|}{MIMIC-Mortality} & \multicolumn{2}{c|}{HuffPost} & \multicolumn{2}{c}{arXiv} \\
& \multicolumn{2}{c|}{(AUC (\%) $\uparrow$)} & \multicolumn{2}{c|}{(Accuracy (\%) $\uparrow$)} & \multicolumn{2}{c}{(Accuracy (\%) $\uparrow$)} \\
& Avg. &  Worst & Avg. & Worst & Avg. &  Worst \\\midrule
Fine-tuning  & 63.37 & 52.45 & 69.59 & 68.91 & 50.31 & 48.19 \\
EWC & 62.07 & 50.41 & 69.42 & 68.61 & \textbf{50.40} & \textbf{48.18} \\
SI  & 61.76 & 50.19 & 70.46 & 69.05 & 50.21 & 48.07 \\
A-GEM  & 61.78 & 50.40 & 70.22 & 69.15 & 50.30 & 48.14 \\\midrule
ERM & 72.89 & 65.80 & 70.42 & 68.71 & 45.94 & 44.09 \\
GroupDRO-T & 76.88 & \textbf{71.40} & 69.53 & 67.68 & 39.06 & 37.18 \\
mixup & 73.69 & 66.83 & \textbf{71.18} & 68.89 & 45.12 & 43.23 \\
LISA  & 76.34 & 71.14 & 69.99 & 68.04 & 47.82 & 45.91 \\
CORAL-T  & \textbf{77.98} & 64.81 & 70.05 & 68.39 & 42.32 & 40.31 \\
IRM-T & 76.16 & 70.64 & 70.21 & 68.71 & 35.75 & 33.91 \\\midrule
\revision{SWA} & 69.53 & 60.83 & \revision{70.98} & \revision{\textbf{69.52}} & \revision{44.36} & \revision{42.54} \\\bottomrule
\end{tabular}
\vspace{-2em}
\end{center}
\end{table*}
Table~\ref{tab:results_offline_main} shows the performance of all methods on Wild-Time under the Eval-Fix setting. Due to space constraints, we report results on the Eval-Stream setting in Appendix~\ref{sec:app_eval_stream_results}. We report the average and standard deviation of each method's performance across three different random seeds. For each task, we visualize the OOD performance on every test timestamp in Figure~\ref{fig:ood_per_step}, and show best ID performance over all approaches as an upper bound on temporally robust performance. The following high-level observations summarize our findings:
\begin{itemize}[leftmargin=*]
    \item In FMoW-Time, MIMIC-Readmission, and MIMIC-Mortality, model performance degrades with time (Figures~\ref{fig:FMoW},~\ref{fig:MIMIC-RD},~\ref{fig:MIMIC-Mort}), as models exhibit higher OOD accuracy on timestamps closer to that of the training data. Such gradual temporal shifts correspond to our motivation in dataset selection, as discussed in Section~\ref{sec:criteria}. In Yearbook (Figure~\ref{fig:yearbook}), performance fluctuates significantly, with models achieving higher OOD accuracy at later timestamps (e.g., $1991-1996$) compared to earlier timestamps (e.g., $1981-1986$). In HuffPost and arXiv, models achieve the best performance on the earliest test timestamps. Nevertheless, there is a significant gap between the OOD performance and best ID performance for all datasets and methods. Furthermore, this performance gap changes in a continual manner over time, indicating that the nature of the distribution shift is correlated with the provided timestamps.
    
    \item Most invariant learning approaches (CORAL-T, GroupDRO-T, IRM-T, LISA, mixup) did not show clear improvements over ERM. \revision{In some cases, invariant learning approaches performed worse than ERM, corroborating the findings in other natural distribution shift benchmarks, such as WILDS~\cite{koh2021wilds}.}
    
    \item \textcolor{black}{Incremental training approaches (Fine-tuning, EWC, SI, A-GEM) improve OOD performance on the arXiv and MIMIC-Readmission datasets, and worst OOD performance on the HuffPost dataset. This is expected, since these datasets exhibit more gradual temporal shifts, and incremental training tends to bias the trained model towards the last few timestamps. 
    In all tasks other than Yearbook, incremental training methods perform worse than invariant learning approaches, indicating the power of invariance in learning temporally robust models.}
    
    \item \revision{Neither self-supervised learning nor ensemble learning approaches show consistent benefits over ERM. In summary, ERM has been shown to be a strong baseline in Wild-Time, even when we reduce the number of training examples, as discussed in Appendix~\ref{sec:app_reduce_training_examples}.}
    
    \item Most results from the Eval-Stream setting (Appendix~\ref{sec:app_eval_stream_results}) concur with the above findings from the Eval-Fix setting. In particular, invariant learning approaches outperform continual learning approaches in more scenarios except FMoW-Time, Huffpost and arXiv, though we do not restrict the buffer size for invariant learning approaches. We hope that Wild-Time will be used to investigate more memory-efficient invariant learning approaches.
\end{itemize}

\section{Comparison with Existing Benchmarks}\label{sec6}
Wild-Time offers a unified framework to facilitate the development of models robust to in-the-wild temporal distribution shifts. We discuss how Wild-Time is related to existing distribution shift and continual learning benchmarks.

\textbf{Relation to Distribution Shift Benchmarks.} Distribution shift has been widely studied in the machine learning community. Early works presented small-scale benchmarks to study distribution shifts in sentiment analysis \cite{fang2014domain} and object detection~\cite{saenko2010adapting}. Subsequent distribution shift benchmarks focused on larger-scale, real-world data. The first line of such benchmarks induce distribution shifts by applying different transformations to object recognition datasets. These benchmarks include: (1) ImageNet-A~\cite{hendrycks2021natural},  ImageNet-C~\cite{hendrycks2019benchmarking}, and CIFAR-10.1~\cite{recht2018cifar}, which add noise or adversarial examples to the original Imagenet ~\cite{russakovsky2015imagenet} and CIFAR~\cite{krizhevsky2009learning} datasets, respectively; (2) Colored MNIST~\cite{arjovsky2019invariant}, which changes the color of digits from the original MNIST. More recent works created domain generalization benchmarks by collecting sets of images with different styles or backgrounds, \revision{such as PACS~\cite{li2017deeper}, DomainNet~\cite{peng2019moment}, VLCS~\cite{fang2013unbiased}, OfficeHome~\cite{venkateswara2017deep}, ImageNet-R~\cite{hendrycks2021many}, BREEDS~\cite{santurkar2020breeds}, Waterbirds~\cite{sagawa2019distributionally}, NICO~\cite{he2021towards}, and MetaShift~\cite{liang2022metashift}.} While these datasets are useful testbeds for verifying the efficacy of new algorithms, they do not reflect natural distribution shifts that arise in real-world applications.

Recently, a few works have constructed datasets and benchmarks for real-world distribution shifts. WILDS benchmark consists of ten datasets spanning a wide range of real-world applications, such medical image recognition, sentiment classification, land-use classification with satellite image, and code autocompletion, with a focus on domain shifts and subpopulation shifts~\cite{koh2021wilds}. WILDS 2.0 extends WILDS and introduces unlabeled data to help boost model robustness to distribution shifts~\cite{sagawa2022extending}. SHIFTS~\cite{malinin2021shifts} is composed of three datasets, concerning weather prediction, machine translation, and self-driving vehicle motion prediction. Unlike these works that focus on general distribution shifts, we target temporal distribution shifts arising in real-world applications. A few recent works have started investigating model robustness over time, in real-world applications such as healthcare-related prediction \cite{guo2022evaluation}, drug discovery~\cite{huang2021therapeutics}, image-based geo-localization ~\cite{cai2021online}, machine reading comprehension~\cite{lin2022continual}, and tweet hashtag prediction \cite{jin2021lifelong}. Unlike prior datasets that target specific applications, Wild-Time presents a comprehensive benchmark comprised of 7 datasets from diverse domains and offers systematic evaluation protocols.

\textbf{Relation to Continual Learning Benchmarks.} Continual learning methods are often benchmarked on image classification datasets. Some popular benchmarks such as RainbowMNIST \cite{finn2019online} and permuted MNIST \cite{kaushik2021understanding} apply various image transformations to a small-scale image dataset to obtain a sequence of tasks. Others such as Split CIFAR100~\cite{krizhevsky2009learning}, Split TinyImagenet~\cite{le2015tiny}, F-CelebA~\cite{ke2020continual}, and Stanford Cars~\cite{KrauseStarkDengFei-Fei_3DRR2013} split a large image dataset into multiple non-overlapping class sets, where each is regarded as one task. A third collection of related benchmarks treats each object recognition dataset as a different task. For example, Visual Domain Decathlon~\cite{li2019learn} consists of 10 datasets from various domains, such as Aircraft~\cite{maji13fine-grained}, SVHN~\cite{netzer2011reading}, Omniglot~\cite{lake2015human}, VGG-Flowers~\cite{nilsback2008automated}, CLEAR~\cite{lin2021clear}. In the natural language processing (NLP) domain, continual learning benchmarks such as ASC~\cite{ke2021adapting} and DSC~\cite{ke2020continual} have been used to evaluate the performance of large-scale pretrained models over time. Unlike these prior benchmarks, Wild-Time presents a collection of datasets that reflect natural temporal distribution shifts arising in real-world applications as well as an evaluation strategy (Eval-Stream) to assess incremental learning approaches.

\vspace{-0.5em}
\section{Conclusion and Discussion}
\vspace{-0.5em}
In this paper, we present the Wild-Time benchmark, and examine in-the-wild distribution shifts over time. We leverage timestamp metadata, which is largely ignored by existing robustness techniques and benchmarks. Wild-Time includes 6 tasks from 5 datasets, which span a range of applications (facial recognition, news, healthcare) and tasks (classification, regresssion).  \revision{On each of these datasets, we systematically benchmark 13 approaches, including continual learning, invariant learning, self-supervised learning, and ensemble learning approaches.} Our experiments show a large gap between ID and OOD performance on all tasks due to temporal distribution shift. We conclude that no existing invariant learning, continual learning, self-supervised, or ensemble learning approach is consistently more robust to temporal distribution shifts than ERM. We hope that Wild-Time facilitates further research in developing temporally robust methods that can be safely deployed in the wild.

\vspace{-0.5em}
\section*{Acknowledgement}
\vspace{-0.5em}
We thank Zhenbang Wu, Shiori Sagawa, Kexin Huang, Ananya Kumar, Scott Lanyon Fleming, and members of the IRIS lab for the many insightful discussions and helpful feedback. This research was funded in part by Apple, Intel, Juniper Networks, and JPMorgan Chase \& Co. Any views or opinions expressed herein are
solely those of the authors listed, and may differ from the views and opinions expressed by JPMorgan Chase
\& Co. or its affiliates. This material is not a product of the Research Department of J.P. Morgan Securities
LLC. This material should not be construed as an individual recommendation for any particular client and is
not intended as a recommendation of particular securities, financial instruments or strategies for a particular
client. This material does not constitute a solicitation or offer in any jurisdiction. CF is a CIFAR fellow.
\bibliographystyle{plainnat}
\bibliography{ref}
\appendix

\section{Dataset Description}
\label{sec:app_data}

\subsection{Yearbook} \label{app:data:yearbook}

\subsubsection{Setup}
\textbf{Problem Setting.} The task is classifying the gender of an American high schooler from a yearbook photo. The input $x$ is a $32 \times 32$ grayscale image, and the label $y$ is male or female. 

\textbf{Data.} Yearbook is based on the Portraits dataset \cite{ginosar2015century} (MIT license), which collected and processed 37,921 frontal-facing yearbook portraits from $1905 - 2013$ from 128 American high schools in 27 states. The Portraits dataset reflects changing fashion styles and social norms over the decades.

The original Portraits dataset did not evaluate models under a distribution shift setting. We use a subset of the Portraits dataset, consisting of data from $1930 - 2013$. Our fixed time split (Eval-Fix) uses the first 41 years ($1930 - 1970$) for ID, and the remaining 43 years for OOD ($1971 - 2013$). For streaming evaluation (Eval-Stream), we treat each year as a single timestamp. \revision{For each timestamp, we randomly allocate 10\% of the data to training, and the remaining 90\% for validation.} For OOD testing, all samples in each year are used. We provide the number of examples allocated to ID Train, ID Test, and OOD Test for each timestamp in Table \ref{tab:yearbook_count}.

The original Portraits dataset is provided as a set of hierarchical directories, organized by year, with PNG images of size $96 \times 96$ pixels. To reduce download times and I/O usage, we downsample the images from \cite{ginosar2015century} to $32 \times 32$ pixels. We exclude the first 25 years ($1905 - 1929$) due to few samples in these years. 

\begin{table}[h]
\small
\centering
\caption{
\label{tab:yearbook_count}
Data subset sizes for the Yearbook task.}
\begin{tabular}{l|rrr}
\toprule
Years & ID Train & ID Test & OOD Test \\
\midrule
1930 - 1934 & 1,051 & 120 & 1,171 \\
1935 - 1939 & 1,361 & 154 & 1,515 \\
1940 - 1944 & 2,047 & 230 & 2,277 \\
1945 - 1949 & 1,979 & 222 & 2,201 \\
1950 - 1954 & 1,604 & 181 & 1,785 \\
1955 - 1959 & 1,820 & 205 & 2,025 \\
1960 - 1964 & 1,482 & 167 & 1,649 \\
1965 - 1969 & 2,812 & 315 & 3,127 \\
\midrule
1970 - 1974 & 2,326 & 260 & 2,586 \\
1975 - 1979 & 2,329 & 261 & 2,590 \\
1980 - 1984 & 2,654 & 298 & 2,952 \\
1985 - 1989 & 2,239 & 251 & 2,490 \\
1990 - 1994 & 2,207 & 249 & 2,456 \\
1995 - 1999 & 2,564 & 287 & 2,851 \\
2000 - 2004 & 2,447 & 274 & 2,721 \\
2005 - 2009 & 1,407 & 159 & 1,566 \\
2010 - 2013 & 1,102 & 125 & 1,227\\
\midrule
Fixed-time split & 14,901 & 1,677 & 21,439 \\
\bottomrule
\end{tabular}
\end{table}

\textbf{Evaluation Metrics.} We evaluate models by their average and worst-time OOD accuracies. The former measures the model's ability to generalize across time, while the latter additionally measures model robustness to trends in time-specific visual patterns.

Eval-Stream evaluates performance across the next 10 years to test on visual trend changes over the decade without resulting in an unreasonably long evaluation time, due to the large number of timestamps in this dataset.

\subsubsection{Broader Context}
Facial recognition has been widely adopted in recent years. Employed by governments and private companies, facial recognition models are used in smartphones, robotics, advanced human-computer interaction systems. However, human appearance shifts over time due to changing social norms (e.g., the practice of smiling to the camera) and fashion trends (e.g., hair styles, popularity of eyewear). To remain reliable and effective, facial recognition models must be robust to changes in human appearance over time.

While Yearbook is not a facial recognition task, the Yearbook dataset can be used to train facial image analysis models that are robust to changes in appearance over time. 


\subsection{FMoW-Time} \label{app:data:fmow}

\subsubsection{Setup}
\textbf{Problem Setting.}
The task in the FMoW-Time dataset is to classify the functional purpose of a region inside a satellite image.
The input $x$ is an $224 \times 224$ RGB satellite image, and the label $y$ is one of $62$ categories of building or land use.
The data was collected from $16$ different years. 
Our fixed time split (Eval-Fix) allocates the first $12$ years for training and the last $4$ years for testing.
For streaming evaluation (Eval-Stream), we treat each year as a single timestamp.

\textbf{Data.}
FMoW-Time is based on the Functional Map of the World dataset (license: \href{https://github.com/fMoW/dataset/blob/master/LICENSE}{https://github.com/fMoW/dataset/blob/master/LICENSE})~\citep{christie2018functional}, a dataset of satellite images taken from $2002 - 2018$, from over $200$ countries.
Each satellite image is labeled according to the functional purpose of the buildings or land depicted in the image.

We adapt the version of the FMoW dataset from the WILDS benchmark \cite{koh2021wilds}, which consists of 141,696 RGB satellite images resized to $224 \times 224$ pixels. The train/val/test data splits in FMoW-WILDS contain images from disjoint location coordinates, and all splits contain data from all 5 geographic regions.
For FMoW-Time, we partition each year's data into train/validation as follows. 
For $2002 - 2013$, we use the FMoW-WILDS Training (ID) split for training, and the Validation (ID) and Test (ID) splits for validation. 
For $2013 - 2015$, we use data in the Validation (OOD) split and allocate 90\% of the data from each year to train, and the remaining 10\% to validation. 
For $2015 - 2017$, we allocate 90\% of the data from each year in the Test (OOD) split to train, and the remaining 10\% to validation. 
Our fixed time split (Eval-Fix) uses the first $11$ years ($2002 - 2013$) for training, and the remaining $5$ years ($2013 - 2018$) for testing.
For streaming evaluation (Eval-Stream), we treat each year as a single timestamp. We provide the number of examples allocated to ID Train, ID Test, and OOD Test at each timestamp in Table \ref{tab:fmow_n}.

\begin{table}
\centering
\small
\caption{
\label{tab:fmow_n}
Data subset sizes for the FMoW-Time task.}
\begin{tabular}{l|rrr}
\toprule
Year & ID Train & ID Test & OOD Test \\
\midrule
2002 & 1,455 & 448 & 1,903 \\
2003 & 1,985 & 570 & 2,555 \\
2004 & 1,545 & 450 & 1,995 \\
2005 & 2,207 & 629 & 2,836 \\
2006 & 2,765 & 796 & 3,561 \\
2007 & 1,338 & 349 & 1,687 \\
2008 & 1,975 & 584 & 2,559 \\
2009 & 6,454 & 1,920 & 8,374 \\
2010 & 16,498 & 4,915 & 21,413 \\
2011 & 19,237 & 5,711 & 24,948 \\
2012 & 21,404 & 6,438 & 27,842 \\
\midrule
2013 & 3,465 & 385 & 3,850 \\
2014 & 5,572 & 620 & 6,192 \\
2015 & 8,885 & 988 & 9,873 \\
2016 & 14,363 & 1,596 & 15,959 \\
2017 & 5,534 & 615 & 6,149 \\
\midrule
Eval-Fix split & 76,863 & 22,810 & 42,023 \\
\bottomrule
\end{tabular}
\vspace{-1em}
\end{table}

\textbf{Evaluation Metrics.}
We evaluate models with the top-1 accuracy, both in terms of the average across all OOD timestamps and the accuracy on the worst timestamp.
The former measures the model's ability to reliably generalize across time and the latter more specifically tests the robustness at the most severe shifts.
For Eval-Stream, we evaluate performance across the next $6$ years.

\subsubsection{Broader Context}
ML models for satellite imagery can automate applications such as deforestation tracking, population density prediction, crop yield prediction~\citep{hansen2013high,tiecke2017mapping,wang2020weakly}.
Visual features in satellite data change over time due to both human and environmental activity, requiring a model that makes predictions for recent images using labeled data from the past.
Through such applications, policy and humanitarian efforts would greatly benefit from temporally robust models which can reliably monitor global-scale satellite imagery even when conditions change over time.


\subsection{MIMIC-IV}\label{app:data:mimic}

\subsubsection{Setup}
\textbf{Problem Setting.} The MIMIC-IV dataset contains two tasks: MIMIC-Readmission and MIMIC-Mortality. For both of these tasks, the input $x$ is the concatenated ICD9 codes of diagnosis and treatment for a single patient.

\begin{itemize}[leftmargin=*]
    \item \textbf{MIMIC-Readmission}: the task is predicting hospital readmission for a patient. The label $y$ is whether the patient was readmitted to the hospital within 15 days.
    \item \textbf{MIMIC-Mortality}: the task is predicting in-hospital mortality for each patient. The label $y$ is whether the patient passed away during their hospital stay.
\end{itemize}

\textbf{Data.} The MIMIC-IV database \cite{mimiciv} contains deidentified EHRs of 382,278 patients admitted to the emergency department or intensive care unit (ICU) at the Beth Israel Deaconess Medical Center (BIDMC) from $2008 - 2019$. To protect patient privacy, the reported admission year is in a three year long date range. Hence, our timestamps are groups of three
years: $2008 -- 2010$, $2011 -- 2013$, $2014 -- 2016$, $2017 -- 2019$. We considered ICU patient data sourced from the clinical information system MetaVision at the BIDMC, released in the MIMIC-IV v1.0 dataset, which contains 53,150 patient records. MIMIC-IV requires PhysioNet credentialing for use of human subject data. 

We use a subset of the original MIMIC-IV dataset, where we regard each admission as one entry. For each admission, we collect the ICD9 codes of diagnosis and treatment. For each record, we concatenate the corresponding ICD9 codes~\cite{world2004international} of diagnosis and treatment. We use the concatenated diagnosis and treatment ICD9 codes as the input feature. Our fixed time split (Eval-Fix) uses the first 6 years ($2008 - 2013$) for training, and the remaining 6 years for testing ($2014 - 2019$). For streaming evaluation (Eval-Stream), we treat each three-year block as a single timestamp. We allocate 20\% of the data at each timestamp for test, and the rest for training. For OOD testing, all samples in each three-year block are used. We provide the number of examples allocated to ID Train, ID Test, and OOD Test for each timestamp in Table \ref{tab:mimic_n}.

\begin{table}
\centering
\small
\caption{
\label{tab:mimic_n}
Data subset sizes for the two MIMIC-IV tasks, MIMIC-Mortality and MIMIC-Readmission.}
\begin{tabular}{l|rrr}
\toprule
3-Year Block & ID Train & ID Test & OOD Test \\
\midrule
2008 - 2010 & 60,851 & 15,215 & 76,066 \\
2011 - 2013 & 55,714 & 13,930 & 69,644 \\
\midrule
2014 - 2016 & 53,932 & 13,485 & 67,417 \\
2017 - 2019 & 45,990 & 11,500 & 57,490 \\
\midrule
Eval-Fix split & 116,565 & 29,145 & 124,907 \\
\bottomrule
\end{tabular}
\end{table}

\textbf{Evaluation Metrics.} For MIMIC-Readmission, we evaluate models by their average and worst-time OOD accuracies. For MIMIC-Mortality, we evaluate models by their average and worst-time ROC-AUC due to label imbalance. The average metric measures the model's ability to generalize across time, while the worst-time metric additionally measures model robustness to temporal distribution shifts in patient data. Eval-Stream evaluates performance across the next 3 years, which represents 25\% of all timestamps in the entire dataset.

\subsubsection{Broader Context} 
Many applications of machine learning to clinical healthcare have emerged in the last decade, such as predicting disease risk \cite{ma2017dipole}, medication changes \cite{yang2021change}, patient subtyping \cite{baytas2017patient}, in-hospital mortality \cite{guo2022evaluation}, and length of hospital stay \cite{ettema2010prediction}. However, a key obstacle in deploying machine learning-based clinical decision support systems is distribution shift associated with changes in healthcare over time \cite{guo2022evaluation}. Existing domain generalization and unsupervised domain adaptation algorithms have been shown to produce less robust models compared to ERM in a variety of tasks (e.g., mortality, length of stay, sepsis, and invasive ventilation prediction) on the MIMIC-IV dataset  \cite{guo2022evaluation}, underscoring the need for better approaches.

The MIMIC-IV Mortality and Readmission tasks evaluate model robustness to temporal shifts in clinical medicine.

\subsection{Huffpost} \label{app:data:huffpost}
\subsubsection{Setup}
\textbf{Problem Setting.} The task is classifying the news category of an article from the headline. The input $x$ is a news headline, and the label $y$ is one of 11 news categories.

\textbf{Data.} Huffpost is based on the Kaggle News Category Dataset \cite{huffpostdatasetbook} (license: CC0: Public Domain),  which contains approximately 200,000 news headlines and their corresponding news categories from the Huffington Post from $2012 - 2018$. The Kaggle News Category Dataset contains 41 different news categories.

We use a subset of the Huffpost dataset, consisting of 7 years from $2012 - 2018$ and samples from 11 news categories (Black Voices, Business, Comedy, Crime, Entertainment, Impact, Queer Voices, Science, Sports, Tech, Travel). We partition the data by year. 

Our fixed time split (Eval-Fix) uses $2016$ as the time split, allocating $2012 - 2015$ (4 years) for ID and $2016 - 2018$ (3 years) for OOD. For streaming evaluation (Eval-Stream), we treat each year as a single timestamp. We allocate 10\% of the data at each timestamp for test, and the rest for training. For OOD testing, all samples are used. Table \ref{tab:huffpost_n} lists the number of examples allocated to ID Train, ID Test, and OOD Test for each timestamp.

The News Category Dataset is provided as a CSV file. We exclude news categories which do not appear in all years $2012 - 2018$ to obtain the 11 news categories in Huffpost. We shuffle samples in each year, and randomly select 10\% of the samples in each year as ID test and allocate the remaining 90\% for training. For OOD testing, all samples in each year are used.

\begin{table}
\small
\centering
\caption{
\label{tab:huffpost_n}
Data subset sizes for the Huffpost task.}
\begin{tabular}{l|rrr}
\toprule
Year & ID Train & ID Test & OOD Test \\
\midrule
2012 & 6,701 & 744 & 7,446 \\
2013 & 7,492 & 832 & 8,325 \\
2014 & 9,539 & 1,059 & 10,599 \\
2015 & 11,826 & 1,313 & 13,140 \\
\midrule
2016 & 10,548 & 1,172 & 11,721 \\
2017 & 7,907 & 878 & 8,786 \\
2018 & 3,501 & 388 & 3,890 \\
\midrule
Eval-Fix split & 35,558 & 3,948 & 24,397 \\
\bottomrule
\end{tabular}
\vspace{-1.5em}
\end{table}

\textbf{Evaluation Metrics.} We evaluate models by their average and worst-time OOD accuracies. The former measures the model's ability to generalize across time, while the latter additionally measures model robustness to trends in time-specific visual patterns.

Eval-Stream evaluates performance across the next 3 years, which represents 42.9\% of the timestamps in the entire dataset.

\subsubsection{Broader Context}
Many language models which deal with information correlated with time exhibit performance degradation in downstream tasks such as Twitter hashtag classification~\cite{jin2021lifelong} or question answering systems~\cite{lin2022continual}. These performance drops along the temporal dimension reflect changes in the style or content of news that change over time. For instance, American politics in 2022 is more polarized than it was in 2012, according to a study by the Pew Research Center \cite{desilver2022polarization}. Models must be robust to such changes in factual knowledge.


\subsection{arXiv} \label{app:data:arxiv}

\subsubsection{Setup}
\textbf{Problem Setting.} The task is classifying the primary classification category of a research paper from the title. The input $x$ is the paper title, and the label $y$ is one of 172 paper categories.

\textbf{Data.} arXiv is based on the Kaggle arXiv Dataset~\cite{clement2019arxiv} (license: CC0: Public Domain), which provides metadata of arXiv preprints from $2007 - 2023$. These include: arXiv id, submitter, authors, title, comments, journal-ref, doi, abstract, categories, and versions.  

We use a subset of the Kaggle arXiv dataset for arXiv, which consists of paper titles and theirr corresponding primary categories. Our fixed time split (Eval-Fix) uses $2016$ as the time split, allocating data from $2007 - 2016$ (10 years) for ID, and data from $2017 - 2022$ (6 years) for OOD. For streaming evaluation (Eval-Stream), we treat each year as a single timestamp. We allocate 10\% of the data at each timestamp for test, and the rest for training. For OOD testing, all samples are used. Table \ref{tab:arxiv_n} lists the number of examples allocated to ID Train, ID Test, and OOD Test for each timestamp.

\begin{table}
\small
\centering
\caption{
\label{tab:arxiv_n}
Data subset sizes for the arXiv task.}
\begin{tabular}{l|rrr}
\toprule
Year & ID Train & ID Test & OOD Test \\
\midrule
2007 & 131,550 & 14,616 & 146,167 \\
2008 & 62,460 & 6,939 & 69,400 \\
2009 & 206,244 & 22,916 & 229,161 \\
2010 & 50,665 & 5,629 & 56,295 \\
2011 & 55,741 & 6,193 & 61,935 \\
2012 & 51,678 & 5,741 & 57,420 \\
2013 & 64,951 & 7,216 & 72,168 \\
2014 & 79,498 & 8,833 & 88,332 \\
2015 & 193,979 & 21,553 & 215,533 \\
2016 & 120,682 & 13,409 & 134,092 \\
\midrule
2017 & 111,024 & 12,336 & 123,361 \\
2018 & 123,891 & 13,765 & 137,657 \\
2019 & 142,767 & 15,862 & 158,630 \\
2020 & 166,014 & 18,445 & 184,460 \\
2021 & 201,241 & 22,360 & 223,602 \\
2022 & 89,765 & 9,973 & 99,739 \\
\midrule
Eval-Fix split & 1,017,448 & 113,045 & 927,449 \\
\bottomrule
\end{tabular}
\vspace{-1.5em}
\end{table}

The arXiv Dataset metadata is provided as a JSON file. We store only the primary category and the preprint title, and sort the data by update date, partitioning by year. We shuffle samples in each year, and randomly select 10\% of the samples in each year as ID test and allocate the remaining 90\% for training. For OOD testing, all samples in each year are used.

\textbf{Evaluation Metrics.} We evaluate models by their average and worst-time OOD accuracies. The former measures the model's ability to generalize across time, while the latter additionally measures model robustness to trends in time-specific visual patterns.

Eval-Stream evaluates performance across the next 6 years, which represents 37.5\% of all timestamps in the dataset.

\subsubsection{Broader Context}
Similar to changes in news and current events reflected in the Huffpost dataset, the content of arXiv preprints also change over time as research fields evolve. For example, ``neural network attack" was originally a popular keyword in the security community, but it gradually became more prevalent in the machine learning community. As a result, primary categories of arXiv preprints shift over time.

\section{Algorithm Description}
\label{sec:app_algorithm}
Before introducing all algorithms, we recall that each example is $(x, y, t)$, where $x$, $y$, $t$ represent input feature, label, and timestamp, respectively.

\subsection{Classical Supervised Learning}
\label{sec:app_method_classical}
\begin{itemize}[leftmargin=*]
    \item \textbf{Empirical Risk Minimization (ERM).}
We first consider Empirical Risk Minimization (ERM). This algorithm ignores the time information ($t$) and minimizes the average training loss
\begin{equation}
    \theta^{*} = \arg\min_{\theta} \ell(x,y;f_{\theta})
\end{equation}
over the entire training dataset.
\end{itemize}

\subsection{Continual Learning}
\label{sec:app_method_continual}
\begin{itemize}[leftmargin=*]
\item \textbf{Fine-tuning.}
In fine-tuning, we use the newly observed labeled examples to continuously fine-tune the learned model without any explicit regularizer between consecutive timestamps.
\item \textbf{Elastic Weight Consolidation (EWC).}
Inspired by synaptic consolidation, EWC slows down the learning process for new tasks based on their relevance to previous tasks. Specifically, when adapting to a new task, EWC's loss function keeps the post-adaptation network parameters close to the parameters learned on previous tasks.
\item \textbf{Synaptic Intelligence (SI).} Motivated by synaptic dynamics, SI enables deep neural network to learn sequence of tasks by using synaptic state to track the parameter values and maintain online estimation of the importance of past learned experience.
\item \textbf{Averaged Gradient Episodic Memory (A-GEM).} Gradient Episodic Memory (GEM) leverages an episodic memory to store a selected set of examples from previous tasks in a continual learning setting. When adapting to a new task, the algorithm aims to make the updated model simultaneously perform well on examples in the new task and examples from the episode memory. A-GEM provides an efficient training strategy for Gradient Episodic Memory that significantly improves its computation and memory efficiency. Specifically, instead of making the updated model perform better on each individual previous tasks in the memory, A-GEM aims to produce a model that shows high average performance across the tasks in the episode memory.

\end{itemize}

\subsection{Temporal Invariant Learning}
\label{sec:app_method_invariant}
\begin{itemize}[leftmargin=*]
    \item \textbf{CORAL.}
CORAL penalizes the differences in the mean and covariance of the feature distributions of each domain. For CORAL, we adapted our implementation from the public repositories for DomainBed and WILDS~\cite{koh2021wilds}. CORAL is applicable to all datasets used in Wild-Time.
\item \textbf{IRM.}
Invariant risk minimization aims to learn an invariant predictor that performs well across all domains. The vanilla IRM objective can be reformulated as a bi-level optimization, which is challenging to solve. Following the original paper~\cite{arjovsky2019invariant}, we adopt IRM-v1 in this paper, an efficient approximation to the original IRM objective for learning invariant predictors.
\revision{\item \textbf{Mixup} is an interpolation-based approach, which generates new training examples by applying the same interpolation strategies on the input features and their corresponding labels \cite{zhang2017mixup}. The original training samples are replaced by the newly generated samples for training.}
\item \textbf{LISA.}
Motivated by mixup~\cite{zhang2017mixup}, LISA selectively interpolates examples to cancel out domain information. LISA has two variants --- intra-label LISA and intra-domain LISA. Intra-label LISA interpolates examples with the same label but from different domains. Intra-domain LISA interpolates examples with the same domain but different labels. Furthermore, as mentioned in~\cite{yao2022improving}, intra-LISA performs better in domain shifts without considering domain information. We follow the implementation of~\citet{yao2022improving} and only apply intra-label in Wild-Time. 

\item \textbf{GroupDRO.}
GroupDRO uses distributionally robust optimization to optimize the worst-domain loss during the training stage. We follow the implementation of~\citet{sagawa2019distributionally} and apply group adjustments, strong penalty and early stopping in GroupDRO.
\end{itemize}

\revision{
\subsection{Self-Supervised Learning}
\label{sec:app_method_ssl}
\begin{itemize}[leftmargin=*]
    \item \textbf{SimCLR} \cite{chen2020simple} is a simple contrastive learning approach for visual recognition. It uses normalized temperature-scaled cross entropy as the loss function and introduces a nonlinear transformation between the learned representation and the contrastive loss. We follow the implementation of Chen et al. \cite{chen2020simple}.
    \item \textbf{SwAV} \cite{caron2020unsupervised} simultaneously clusters the data and encourages the consistency of 
    cluster assignments generated by different kinds of data augmentations. We follow the implementation of Caron et al. \cite{caron2020unsupervised}.
\end{itemize}}

\revision{
\subsection{Bayesian Learning}
\label{sec:app_method_bayesian}
\begin{itemize}[leftmargin=*]
\item \textbf{SWA.} Stochastic Weight Averaging \cite{izmailov2018averaging} averages multiple parameter values along the trajectory of SGD with almost no computational overhead. This method has been shown to lead to better in-distribution generalization due to its ability to find a better approximation to the posterior distribution over parameters. This property is reflected through the flatness of the learned optima. We follow the official implementation of SWA with the same learning rate as ERM and use default values for other hyperparameters.
\end{itemize}}

\section{Experimental Details}

All reported results are averaged over 3 random seeds. Experiments are conducted on a GPU-cluster with 6 GPU nodes. All classification tasks (i.e., Yearbook, FMoW-Time, MIMIC Mortality, MIMIC Readmission, Precipitation, HuffPost, arXiv) were trained with cross-entropy loss. 
\revision{In our experiments, we tune hyperparameters of all baselines by applying cross-validation with grid search.}

For all methods, we use minibatch stochastic optimizers to train models, sampling uniformly from the ID set (in the Eval-Fix setting) or from each timestamp (in the Eval-Stream setting).

We report the number of train iterations used to train baselines for each dataset, under both the Eval-Fix and Eval-Stream settings. A single train iteration corresponds to one update via loss backpropagation. Under the Eval-Fix setting, the number of train iterations is the number of updates to the model on the ID train set. Under the Eval-Stream setting, in which models are trained incrementally, the number of train iterations corresponds to the number of updates to the model at each timestamp.

\subsection{Eval-Fix Split Determination}
\label{sec:app_eval_fix_split}
To determine the time splits for the Eval-Fix setting of each dataset, we considered all ID/OOD splits ranging from 40\%-60\% ID/OOD to 80\%-20\% ID/OOD. For each of these time splits, we ran ERM on the ID and OOD sets, and selected the split with the largest discrepancy between average ID accuracy and average OOD accuracy.

\revision{\subsection{Detailed Set Split Strategy}
\label{sec:app_set_split}
Suppose we have $T$ timestamps. At each timestamp, we randomly sample 90\% of the examples for training, and allocate the remaining 10\% validation examples for ID evaluation. We detail the difference between the Eval-Fix and Eval-Stream setting as follows:

\textbf{Eval-Fix Setting. }In Eval-Fix, as shown in Figure~\ref{fig:split_fix}, we have a split timestamp $t_s$. The ID timestamps are $t<t_s$, and the OOD timestamps are $t \geq t_s$. The training set consists of all training examples from the ID timestamps $t < t_s$. The ID validation set consists of all validation examples from the ID timestamps $t < t_s$.  All examples in all test timestamps $t \geq t_s$ are used as the OOD test set. 

\begin{figure}[h]
\centering
\includegraphics[width=0.8\textwidth]{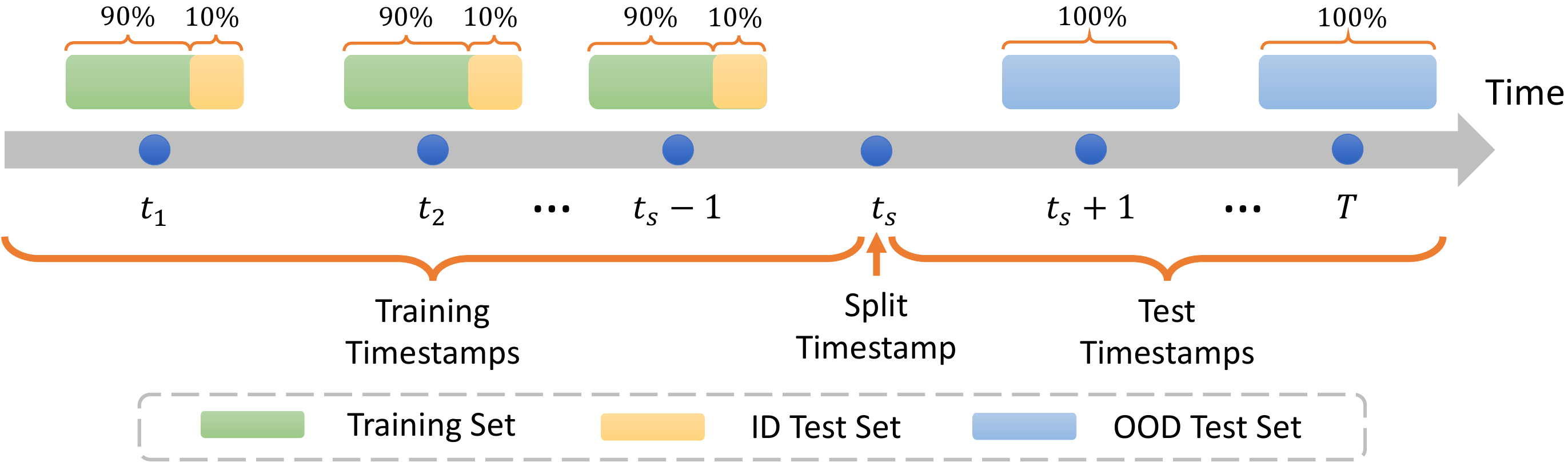}
\caption{\revision{Data split under Eval-Fix setting.}}
\label{fig:split_fix}
\end{figure}

\textbf{Eval-Stream Setting.} In Eval-Stream, at each evaluation timestamp, we evaluate across the next $K$ timestamps. Specifically, at each timestamp $t \in [1, …, T]$, we evaluate our model across the timestamps \{$t + 1, \dots, t + K$\}, which is illustrated in Figure~\ref{fig:split_stream}.

\begin{figure}[h]
\centering
\includegraphics[width=0.8\textwidth]{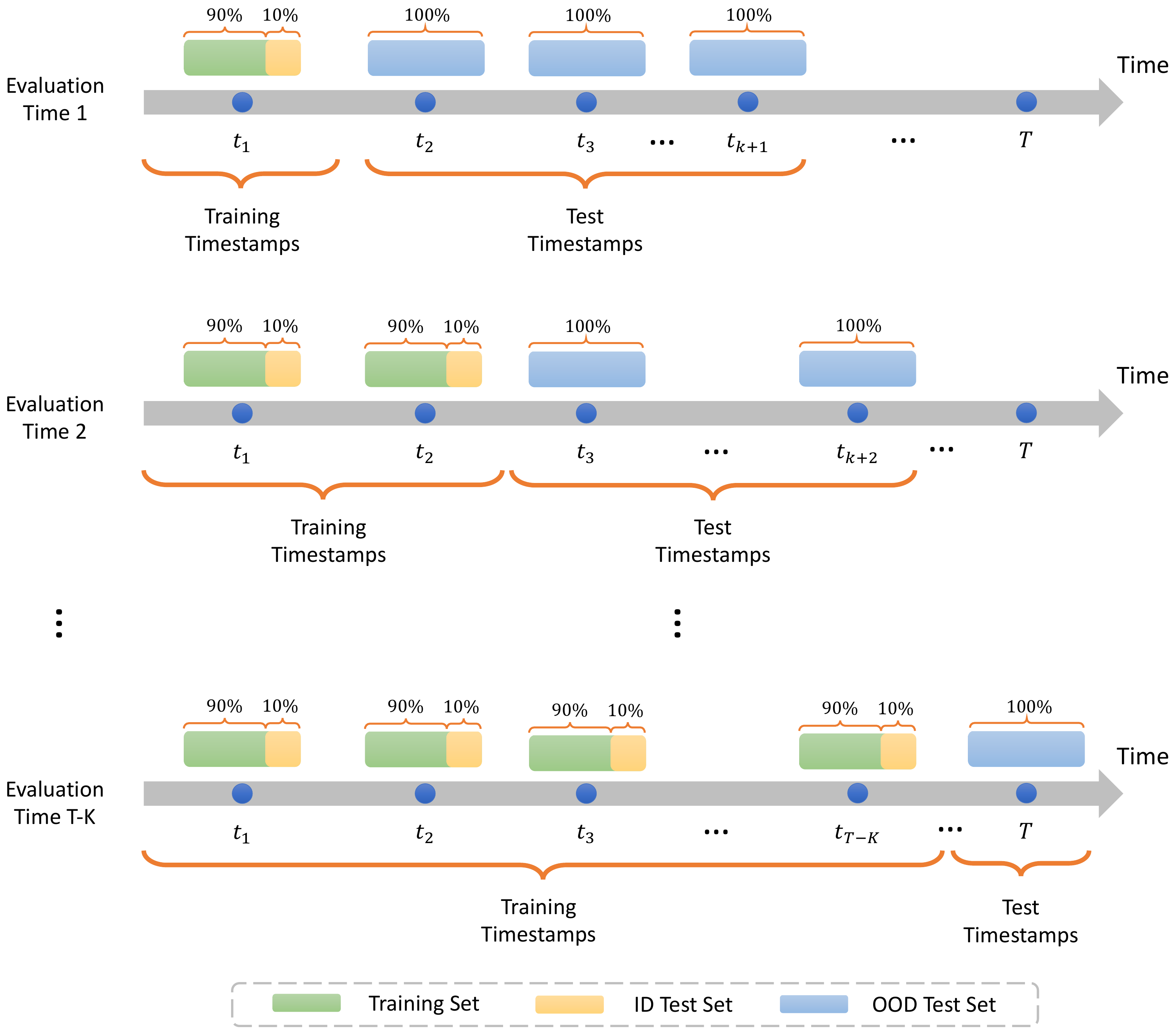}
\caption{\revision{Data split under Eval-Stream setting.}}
\label{fig:split_stream}
\end{figure}

Hence, Eval-Fix can be viewed as a single timestamp evaluation within Eval-Stream, where we evaluate only at $t_s$ and set $n = T - t_s$.}

\subsection{Hyperparameter Settings and Model Architectures}

\subsubsection{General Settings}
\textbf{Yearbook.} We use a 4-layer convolutional network. Each convolutional layer has kernel size $3 \times 3$, stride of $1 \times 1$, padding of size 1, 32 output channels, a spatial batch norm layer, ReLU activation, and a 2D max pool layer with kernel size $2 \times 2$.

We use the Adam optimizer with a fixed learning rate of $10^{-3}$ and train with a batch size of 32. Baselines were trained for 3000 iterations under the Eval-Fix setting and for 100 iterations under the Eval-Stream setting.

\textbf{FMoW-Time.} Following \cite{koh2021wilds} and \cite{christie2018functional}, we use a DenseNet-121 model \cite{huang2017densely} pretrained on ImageNet with no $L_2$ regularization.

We use the Adam optimizer with an initial learning rate of $10^{-4}$ that decays by 0.96 per epoch and a batch size of 64. Baselines were trained for 3000 iterations for the Eval-Fix setting and for 500 iterations for the Eval-Stream setting.

\textbf{MIMIC-IV.} We use a Transformer, consisting of an encoder and an decoder. Here, we collect the vocabulary based on the ICD9 codes.

We use the Adam optimizer with a learning rate of $5\times 10^{-4}$ and a batch size of 128. Baselines were trained for 3000 iterations under the Eval-Fix setting and for 500 iterations under the Eval-Stream setting.


\textbf{Huffpost.} We use a network backbone comprising of a pretrained DistilBERT base model (uncased) from \cite{sanh2019distilbert} and a fully-connected, classification layer.

We use the AdamW optimizer with a learning rate of $2\times10^{-5}$, weight decay of $10^{-2}$, and train with a batch size of 32. Baselines were trained for 6000 iterations under the Eval-Fix setting and for 1000 iterations under the Eval-Stream setting.

\textbf{arXiv.} We use the same network backbone, optimizer, learning rate, weight decay, and number of train iterations as those used for the Huffpost dataset. We train all baselines with a batch size of 64.

\subsubsection{Algorithm-Specific Hyperparameters}

\paragraph{Temporally Invariant Methods}

\begin{table}[t]
\centering
\caption{Hyperparameters for CORAL, GroupDRO, and IRM baselines on all WildT datasets.}
\label{table:group-baselines-hyperparameters}
\small
\begin{tabular}{l|ccccc}
\toprule
Dataset & CORAL Penalty & IRM Penalty & lr & \# Substreams & Substream Size \\
\midrule
Yearbook & 0.9 & 1.0 & 1e-1 & 10 & 5 \\
FMoW-Time & 0.9 & 1.0 & 1e-4 & 3 & 3 \\
MIMIC-IV-Mort & 1.0 & 1.0 & 5e-4 & 4 & 3 \\
MIMIC-IV-Readmit & 1.0 & 1.0 & 5e-4 & 3 & 3 \\
HuffPost & 0.9 & 1.0 & 2e-5 & 3 & 2 \\
arXiv & 0.9 & 1.0 & 2e-5 & 4 & 4 \\
\bottomrule
\end{tabular}
\end{table}

For GroupDRO, CORAL, and IRM, we follow WILDS \cite{koh2021wilds} and use minibatch stochastic optimizers to train models, sampling uniformly from each substream (i.e., the domain in our temporal robustness setting), regardless of the number of training examples in the substream.

\begin{itemize}[leftmargin=*]
    \item \textbf{GroupDRO.} We adapted the implementation of GroupDRO from~\citet{sagawa2019distributionally} and~\citet{koh2021wilds}. Each example in the minibatch is sampled independently with uniform probabilities across substreams.

    We list the hyperparameters used for GroupDRO on all WildT datasets in Table \ref{table:group-baselines-hyperparameters}; namely, the number of substreams (e.g., number of groups) and substream size (e.g., group size).
    \item \textbf{CORAL.} We adapted the implementations of DeepCORAL from~\citet{gulrajani2020search} and~\citet{koh2021wilds}, and compute CORAL penalties between features from all pairs of substreams, which we treat as groups/domains.

    We list the hyperparameters used for CORAL on all WildT datasets in Table \ref{table:group-baselines-hyperparameters}, which include the CORAL penalty $\lambda_c$, number of substreams (e.g., number of groups), and substream size (e.g., group size). CORAL was trained with a penalty of $\lambda_c = 0.1$ on the MIMIC-Mortality task, and $\lambda_c = 1.0$ on the MIMIC-Readmission task. For all remaining datasets, we used a default penalty of $\lambda_c = 0.9$.
    \item \textbf{IRM.} We adapted the implementations of IRM from~\citet{arjovsky2019invariant} and~\citet{koh2021wilds}.

    We list the hyperparameters used for IRM on all WildT datasets in Table \ref{table:group-baselines-hyperparameters}, which include the IRM penalty $\lambda_i$, number of substreams (e.g., number of groups), and substream size (e.g., group size). IRM was trained with a penalty of $\lambda_i = 1.0$ on all datasets.

    \item \textbf{LISA.} We adapted the implementation of LISA from \citet{yao2022improving} and implemented intra-label LISA, where training samples with the same label are interpolated. For the Yearbook and FMoW-Time datasets, the input image tensors were interpolated. For the arXiv, Huffpost, and MIMIC-IV datasets, the learned feature representations were interpolated.
    
    All LISA experiments were conducted with $\alpha = 2.0$, where the interpolation ratio $\lambda \in [0,1]$ is drawn from a $\text{Beta}(\alpha, \alpha)$ distribution.
    
    \item \revision{\textbf{Mixup.} For mixup, we use the same hyperparameters as ERM.}

\end{itemize}

\paragraph{Continual Learning Methods}
\begin{itemize}[leftmargin=*]
    \item \textbf{A-GEM.} We adapted the implementation of A-GEM from~\citet{chaudhry2018efficient} and ``Mammoth - An Extendible (General) Continual Learning Framework for Pytorch" \cite{boschini2022class, buzzega2020dark}. All A-GEM experiments were conducted with a default buffer size of $1000$.
    \item \textbf{EWC.} We adapted the implementation of EWC from~\citet{kirkpatrick2017overcoming},~\citet{vandeven2018generative}, and~\citet{vandeven2019three}.
    
    For the EWC loss regularization strength, we use a default value of $0.5$ for the Yearbook, FMoW-Time, Huffpost, arXiv, and MIMIC-IV-Readmit datasets. For MIMIC-IV-Mortality, we use $1.0$.

    \item \textbf{SI.} We adapted the implementation of SI from~\citet{zenke2017continual},~\citet{vandeven2018generative},~\citet{vandeven2019three}.

    For the SI loss regularization strength $\lambda_s$, we use a default value of $0.1$ for all datasets.
\end{itemize}

\revision{
\paragraph{Self-Supervised Methods}
\begin{itemize}[leftmargin=*]
    \item \textbf{SimCLR.} We implement SimCLR using the Lightly framework ~\cite{susmelj2020lightly}. We apply SimCLR to learn representations, and then fine-tune the model with the same (labeled) training data.
    
    For both Yearbook and FMoW-Time, we use the set of image transforms from Chen et al. \cite{chen2020simple}. Specifically, we sequentially apply the following three random augmentations: random cropping followed by resize back to the original size, color distortions, and Gaussian blur. We list all hyperparameters in Table \ref{table:simclr-hyperparameters}.

    \item \textbf{SwaV.} We implement SwaV using the Lightly framework ~\cite{susmelj2020lightly}. We apply SwaV to learn representations, and then fine-tune the model with the same training data. We follow the multi-crop augmentation strategy proposed by Caron et al. \cite{caron2020unsupervised}. We use 2 views and list all hyperparameters in Table \ref{table:swav-hyperparameters}.
\end{itemize}}

\begin{table}
\begin{minipage}{0.46\linewidth}
\centering
\small
\caption{\revision{Hyperparameters for SimCLR on Yearbook and FMoW-Time.}}
\label{table:simclr-hyperparameters}

\resizebox{\columnwidth}{!}{\setlength{\tabcolsep}{1.5mm}{
\begin{tabular}{l|cc}
\toprule
Dataset & Yearbook & FMoW-Time \\
\midrule
Prob. Color Jitter & 0.8 & 0.8 \\
Color Jitter Strength & 0.5 & 0.5 \\
Min. Crop Scale & 0.08 & 0.08 \\
Prob. Grayscale & 0.2 & 0.2 \\
Kernel Size & 0.1 $\times$ 32 & 0.1 $\times$ 224 \\
Prob. Vertical Flip & 0.5 & 0 \\
Prob. Horizontal Flip & 0.5 & 0.5 \\
Prob. Rotation (+90) & 0.0 & 0.5 \\ \midrule
Embedding Dim. & 128 & 128 \\ \midrule 
No. SSL Iters. & 2700 & 1500 \\
No. Finetune Iters. & 300 & 1500 \\
\bottomrule
\end{tabular}}}
\end{minipage}\hfill%
\begin{minipage}{0.46\linewidth}
\centering
\small
\caption{\revision{Hyperparameters for SwaV on Yearbook and FMoW-Time.}}
\label{table:swav-hyperparameters}
\resizebox{\columnwidth}{!}{\setlength{\tabcolsep}{1.5mm}{
\begin{tabular}{l|cc}
\toprule
Dataset & Yearbook & FMoW-Time \\
\midrule
No. Views & 2 & 2 \\
Crop Sizes & 224, 96 & 224, 96 \\
No. Crops & 2, 6 & 2, 6 \\
Min. Crop Scale & 0.08, 0.05 & 0.08, 0.05 \\
Max. Crop Scale & 1.0, 0.14 & 1.0, 0.14 \\
Prob. Horizontal Flip & 0.5 & 0.5 \\
Prob. Color Jitter & 0.8 & 0.8 \\
Color Jitter Strength & 0.8 & 0.8 \\
Prob. Grayscale & 0.2 & 0.2 \\ \midrule
Embedding Dim. & 128 & 128 \\ 
No. Prototypes & 32 & 1024 \\ \midrule 
No. SSL Iters. & 2700 & 1500 \\
No. Finetune Iters. & 300 & 1500 \\
\bottomrule
\end{tabular}}}
\end{minipage}
\end{table}

\revision{
\paragraph{Bayesian Methods}
\begin{itemize}[leftmargin=*]
    \item \textbf{SWA.} We follow the official implementation of SWA \cite{izmailov2018averaging, athiwaratkun2018there}. We use the same learning rate as ERM and use default values for other hyperparameters.
\end{itemize}
}

\begin{table}
\centering
\small
\caption{Hyperparameters for EWC and SI baselines on all WildT datasets.}
\label{table:ewc-si-baselines-hyperparameters}

\begin{tabular}{l|cc}
\toprule
Dataset & EWC $\lambda_e$ & SI $\lambda_s$ \\
\midrule
Yearbook & 0.5 & 0.1 \\
FMoW-Time & 0.5 & 0.1 \\
MIMIC-IV-Mort & 1.0 & 0.1 \\
MIMIC-IV-Readmit & 0.5 & 0.1 \\
HuffPost & 0.5 & 0.1 \\
arXiv & 0.5 & 0.1 \\
\bottomrule
\end{tabular}
\end{table}

\section{Results Under Eval-Stream Setting}
\label{sec:app_eval_stream_results}
\begin{figure}[t]
\centering
\begin{subfigure}[c]{0.48\textwidth}
		\centering
\includegraphics[width=\textwidth]{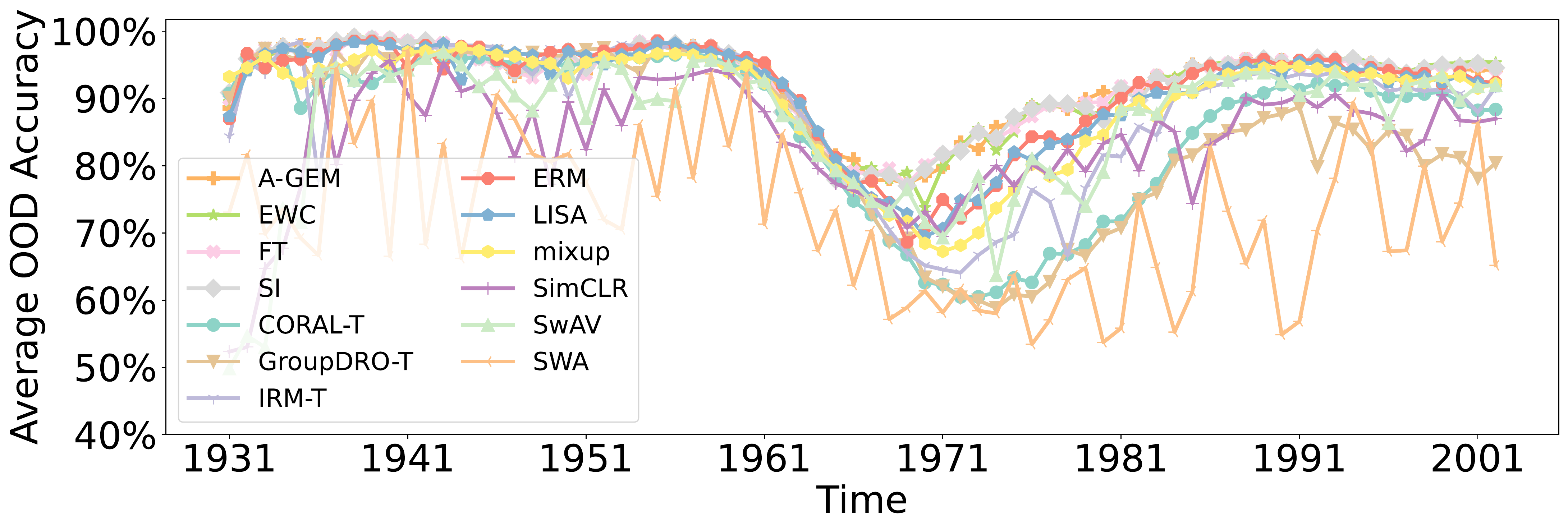}
    \caption{\revision{Yearbook Avg}}
\end{subfigure}
\begin{subfigure}[c]{0.48\textwidth}
		\centering
\includegraphics[width=\textwidth]{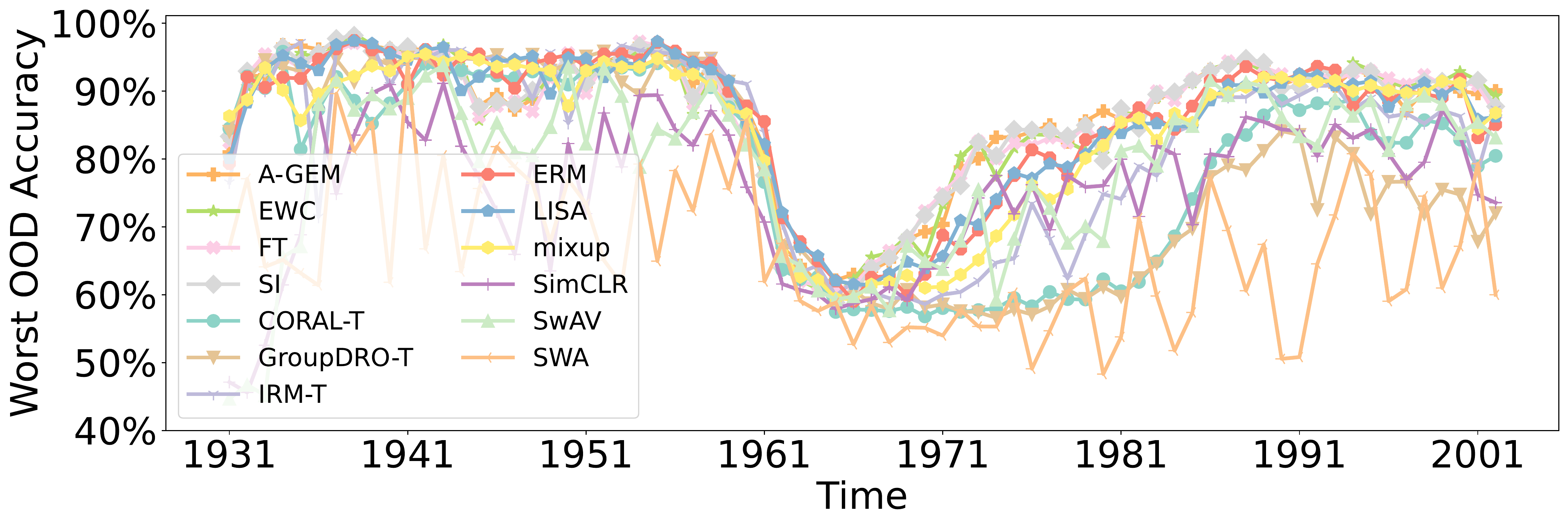}
    \caption{\revision{Yearbook Worst}}
\end{subfigure}
\begin{subfigure}[c]{0.24\textwidth}
		\centering
\includegraphics[width=\textwidth]{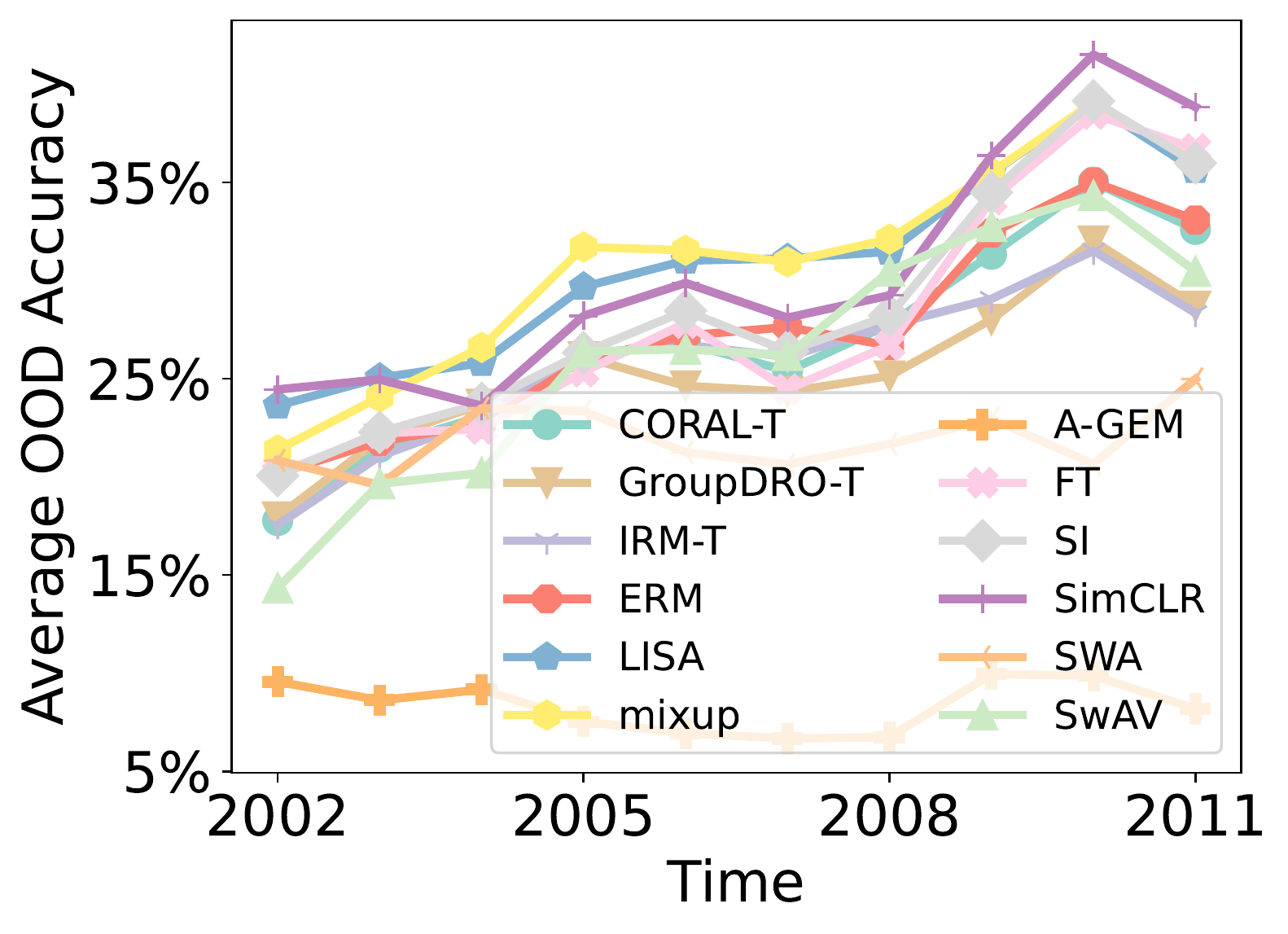}
    \caption{\revision{FMoW-Time Avg}}
\end{subfigure}
\begin{subfigure}[c]{0.24\textwidth}
		\centering
\includegraphics[width=\textwidth]{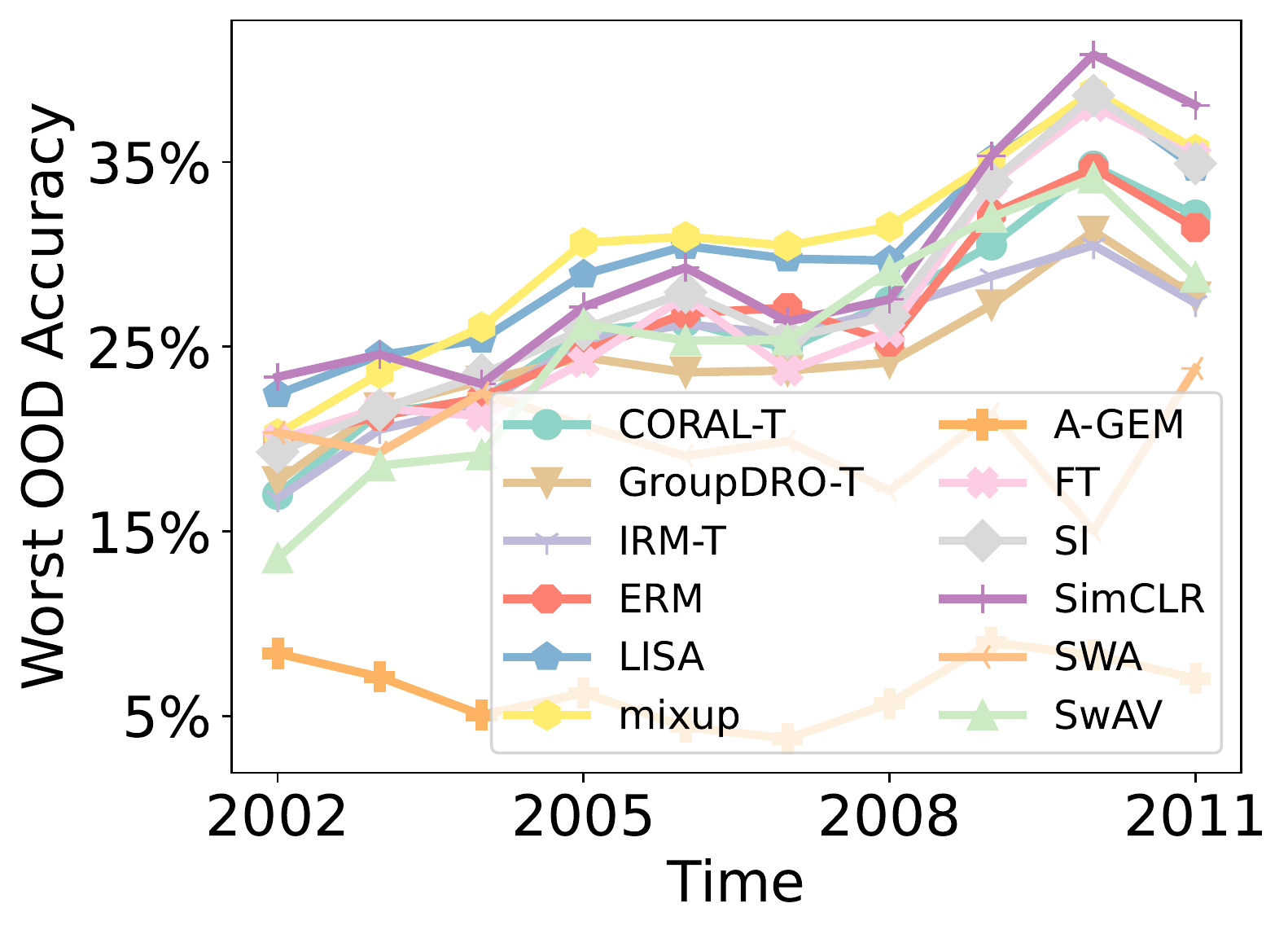}
    \caption{\revision{FMoW-Time Worst}}
\end{subfigure}
\begin{subfigure}[c]{0.24\textwidth}
		\centering
\includegraphics[width=\textwidth]{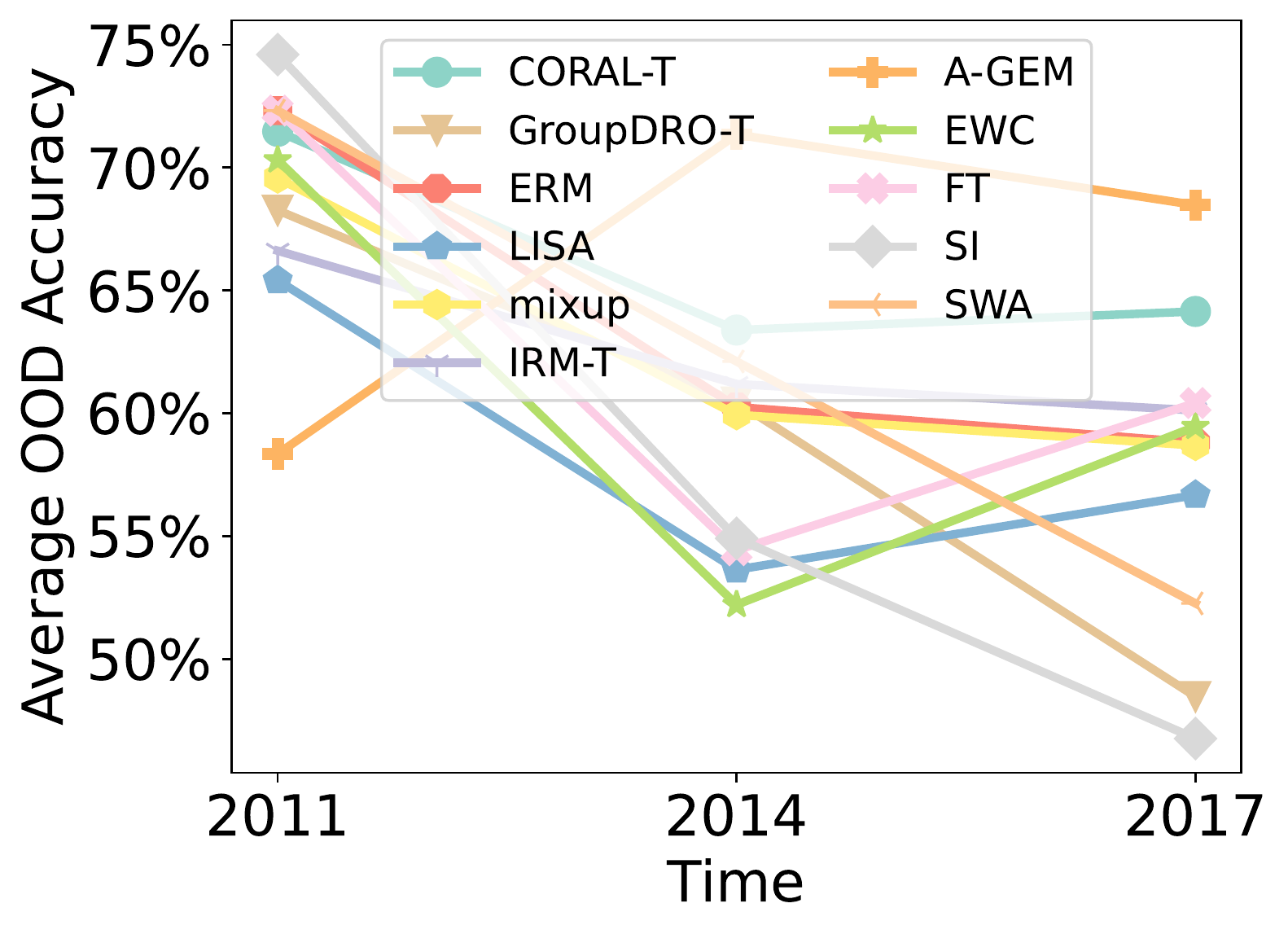}
    \caption{Readmission Avg}
\end{subfigure}
\begin{subfigure}[c]{0.24\textwidth}
		\centering
\includegraphics[width=\textwidth]{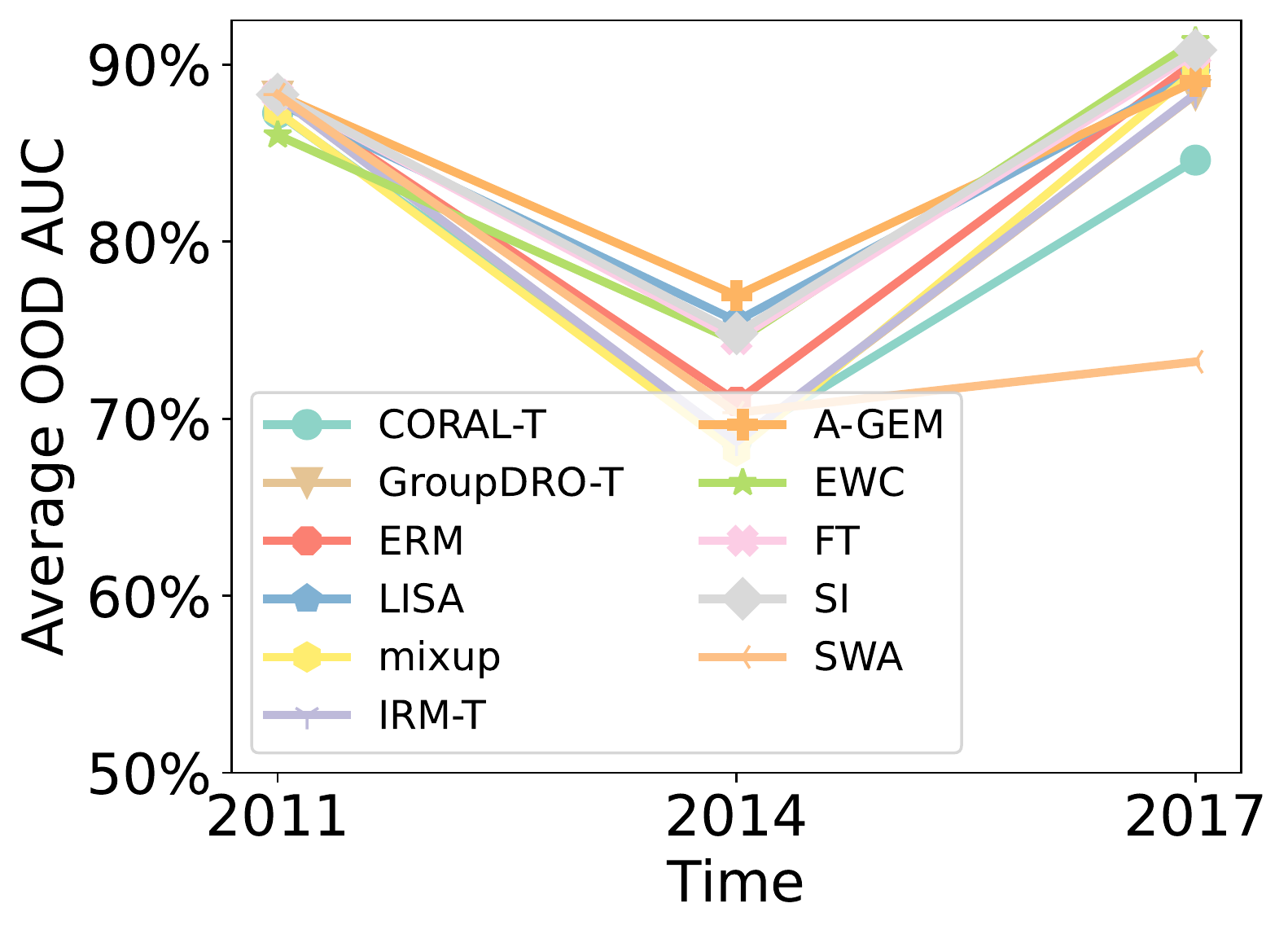}
    \caption{Mortality Avg}
\end{subfigure}
\begin{subfigure}[c]{0.24\textwidth}
		\centering
\includegraphics[width=\textwidth]{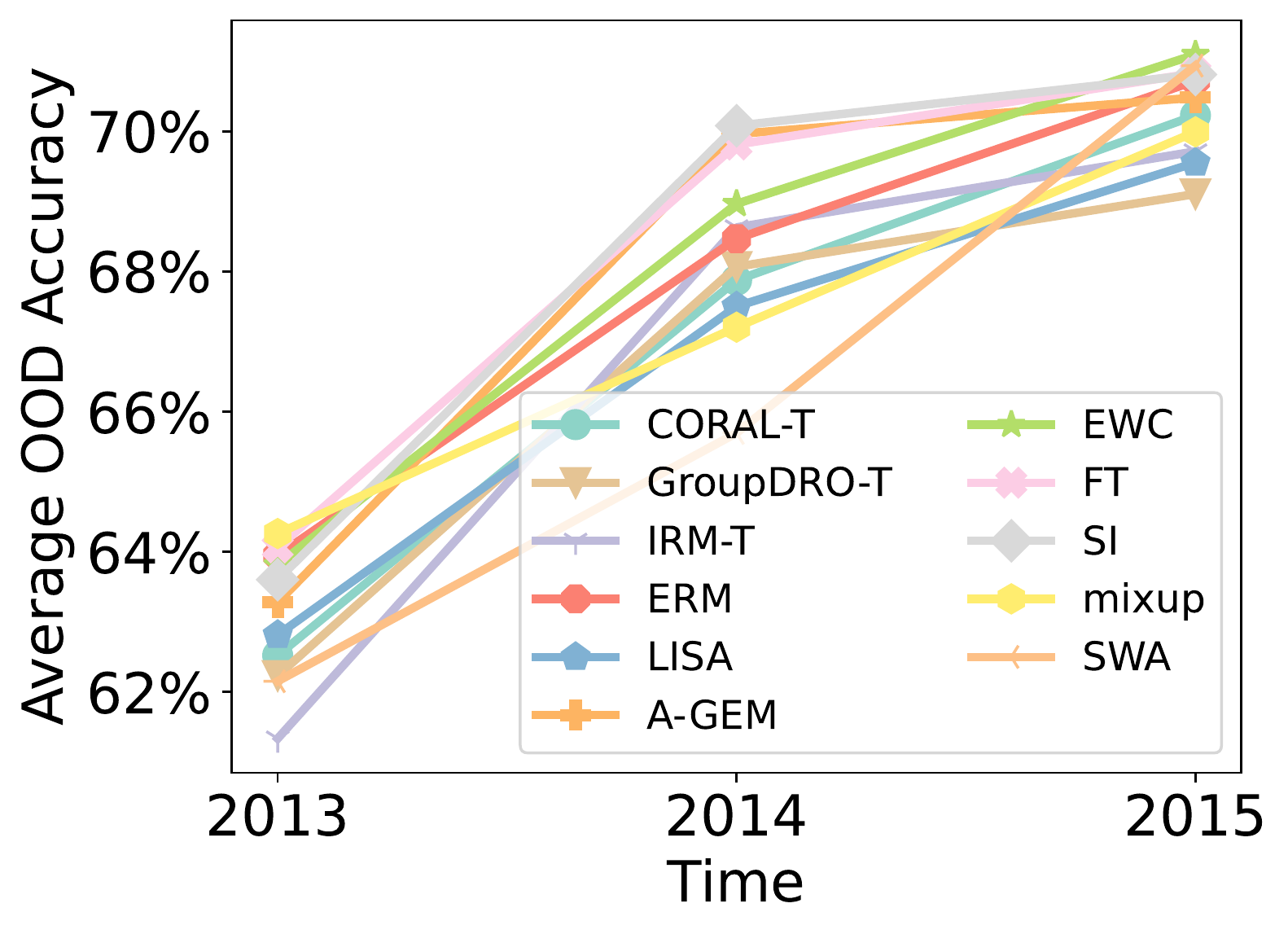}
    \caption{\label{fig:yearbook}: Huffpost Avg}
\end{subfigure}
\begin{subfigure}[c]{0.24\textwidth}
		\centering
\includegraphics[width=\textwidth]{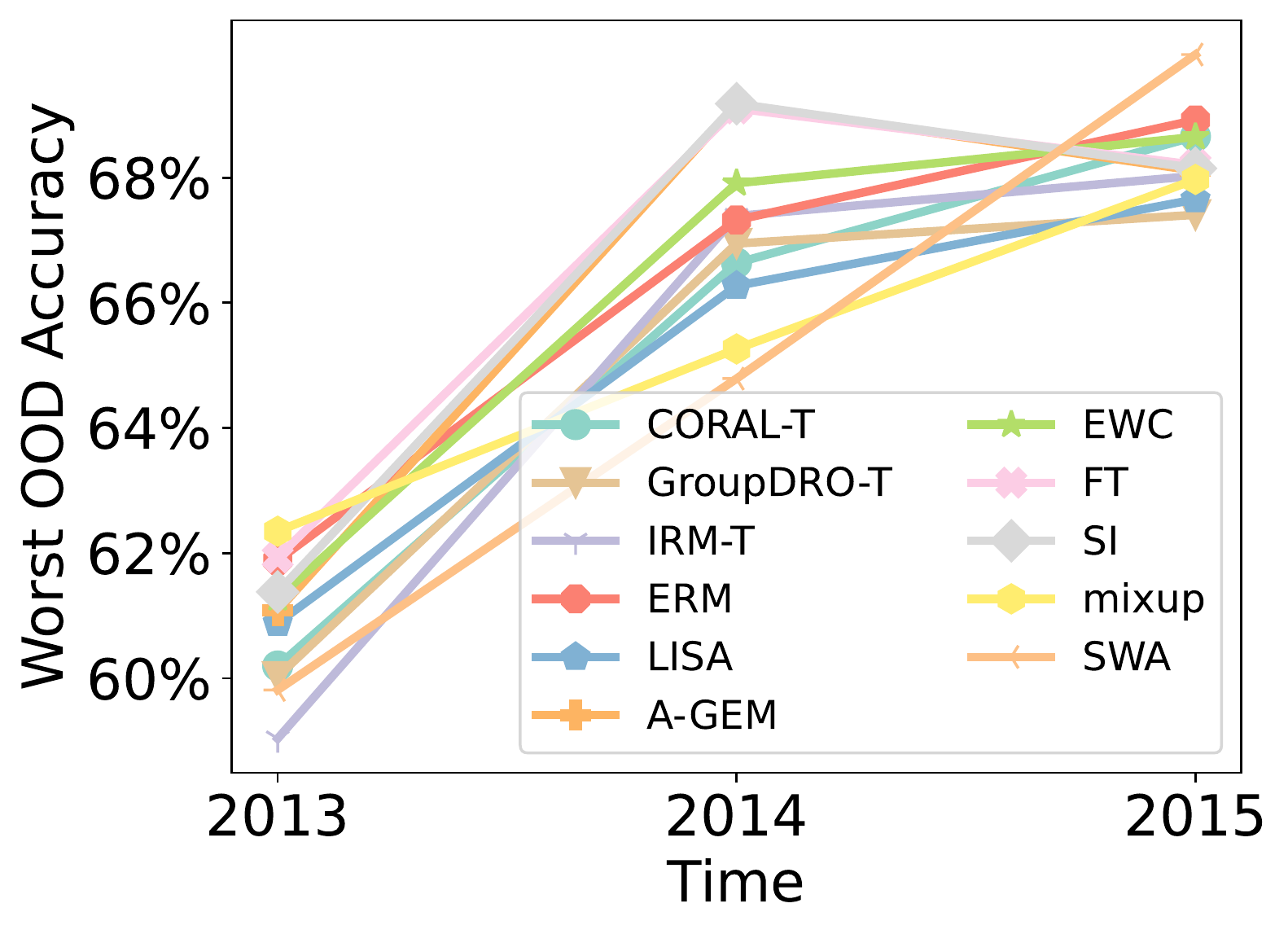}
    \caption{\label{fig:FMoW}: Huffpost Worst}
\end{subfigure}
\begin{subfigure}[c]{0.24\textwidth}
		\centering
\includegraphics[width=\textwidth]{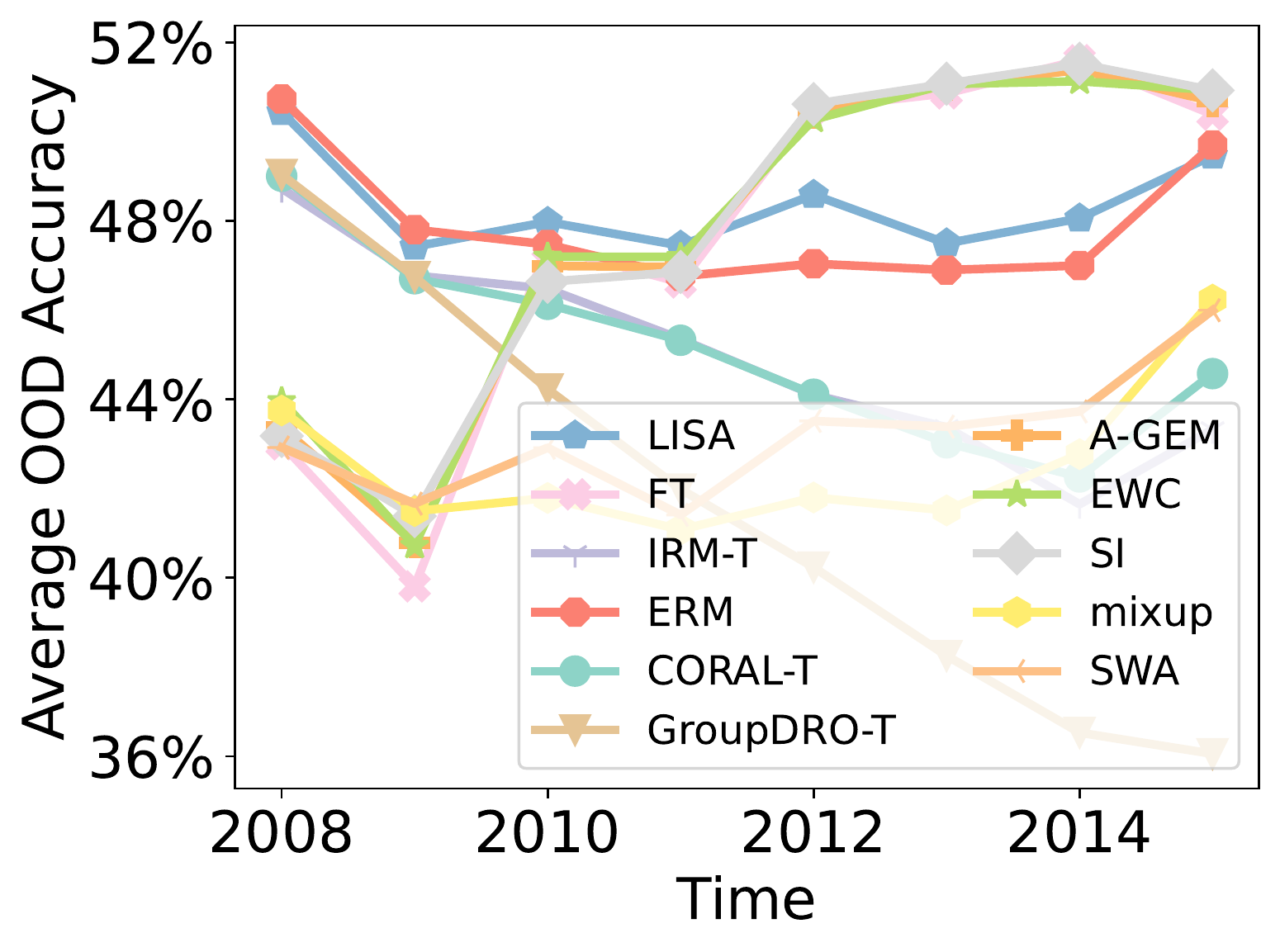}
    \caption{arXiv Avg}
\end{subfigure}
\begin{subfigure}[c]{0.24\textwidth}
		\centering
\includegraphics[width=\textwidth]{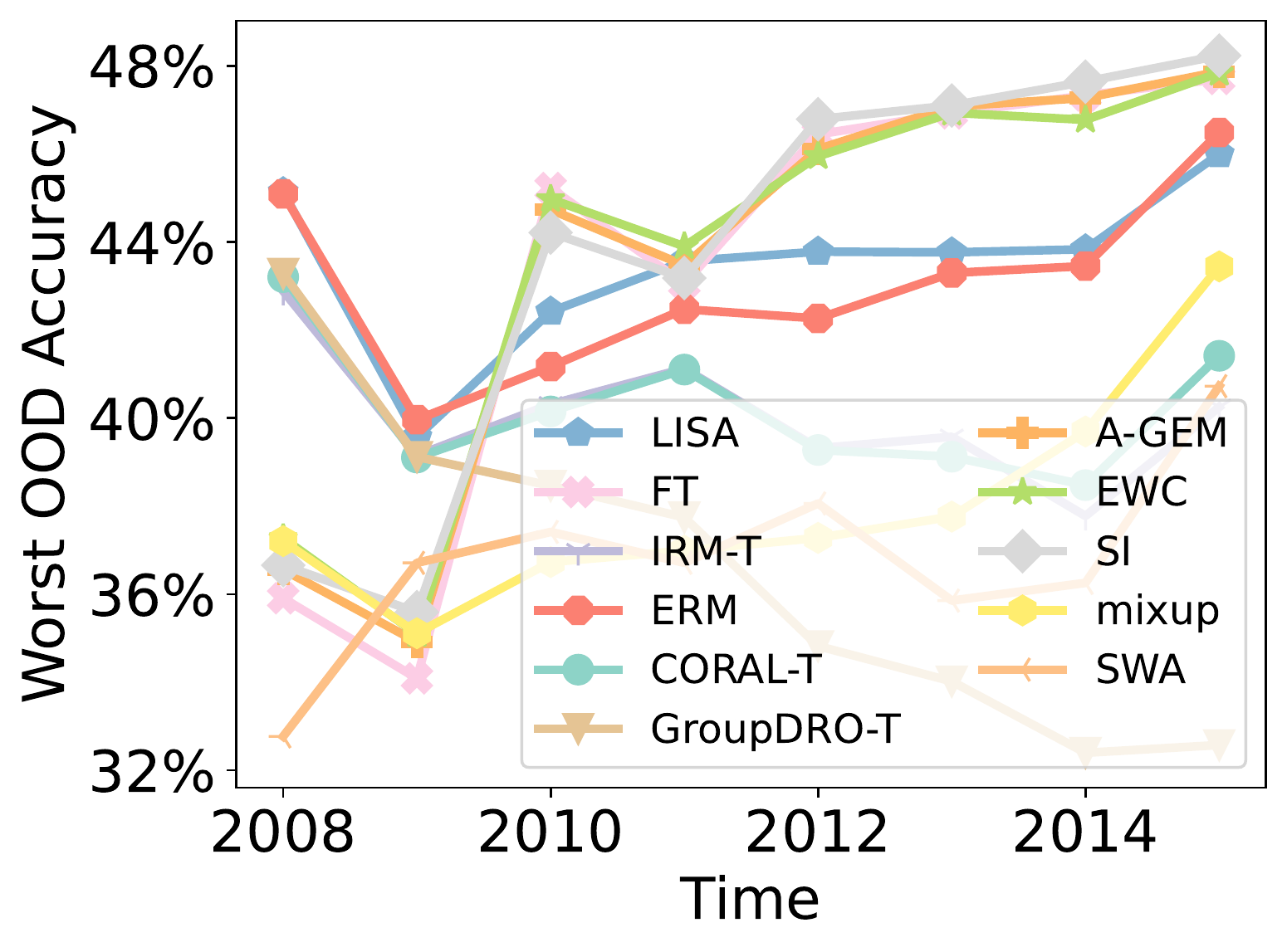}
    \caption{arXiv Worst}
\end{subfigure}
\caption{Results under Eval-Stream Setting. Note that for MIMIC-Readmission (e) and MIMIC-Mortality (f), we only report the average timestamp performance as we only evaluate on the next timestamp, which is a three-year block.}
\vspace{-1.0em}
\label{fig:ood_stream}
\end{figure}

Under Eval-Stream setting, we visualize the average performance and worst-time performance for every timestamp. For each timestamp, we calculate the average/worst performance over the evaluated time window. The results of all tasks are shown in Figure~\ref{fig:ood_stream}. The key observations are very close to the findings under Eval-Fix setting. Additionally, invariant learning approaches performs slightly better than continual learning approaches in most tasks.

\revision{Under the Eval-Stream setting, we further explain why continual learning approaches fail to improve over other baselines in the Eval-Stream setting from the following two reasons: (1) Most existing continual learning approaches focus on backward transfer (i.e., catastrophic forgetting). In Wild-Time, we focus on forward transfer, and evaluate performance on future timestamps (i.e., temporal robustness); (2) For continual learning approaches that also focus on forward transfer (e.g., A-GEM), most of these approaches only show improvements on manually delineated sets of tasks with artificial temporal variations (e.g., Split CUB, Split CIFAR), but are not evaluated on benchmarks with natural temporal distribution shifts, such as Wild-Time. Analogously, we note that invariant learning approaches show improvements in artificial datasets (e.g., ColoredMNIST, Waterbirds~\cite{sagawa2019distributionally}), but fail to outperform ERM in benchmarks with natural distribution shifts, e.g., WILDS~\cite{koh2021wilds}.
}


\section{Additional Experiments under Eval-Fix Setting}

\subsection{Standard Split vs. Mixed Split}
\label{sec:split_compare}
\revision{We verify that the performance gap between ID and OOD timestamps are not caused by the difficulty of examples from OOD timestamps.} First, we analyze the effect of the difficulty of OOD examples. We use two kinds of data splits -- standard split and mixed split. In the standard split, the model is trained on timestamps before the split time and then evaluated on examples from future timestamps. In the mixed split, the training data is merged from all timestamps, and the model is evaluated on the original OOD examples. We report the results in Table~\ref{table:dataset-split} and observe large performance gaps between standard split and mixed split on all Wild-Time tasks. The observation verifies that the performance gaps between ID and OOD are not caused by the difficulty.

\begin{figure}[h]
\centering
\begin{subfigure}[c]{0.31\textwidth}
		\centering
\includegraphics[width=\textwidth]{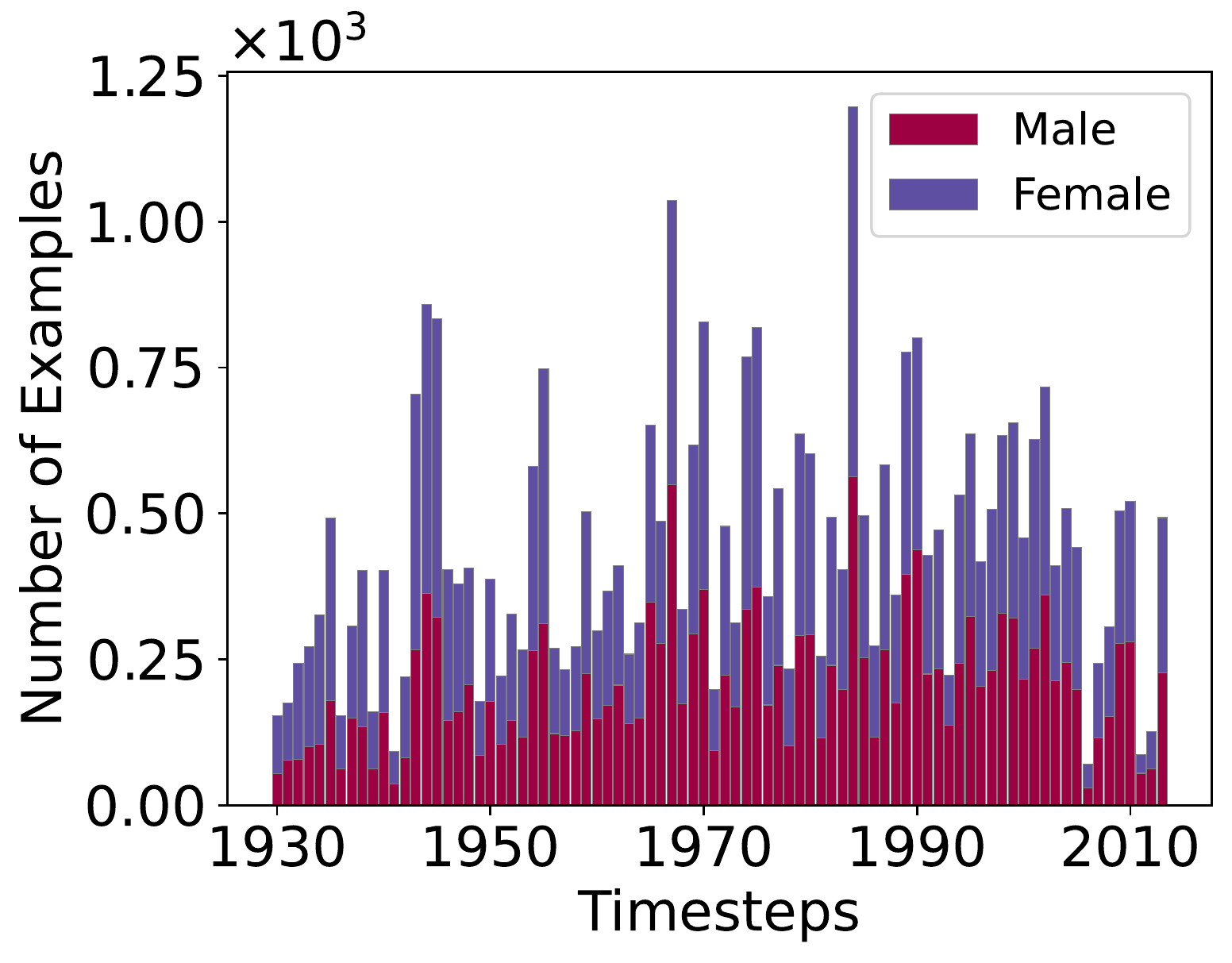}
    \caption{\revision{Yearbook}}
\end{subfigure} 
\begin{subfigure}[c]{0.31\textwidth}
		\centering
\includegraphics[width=\textwidth]{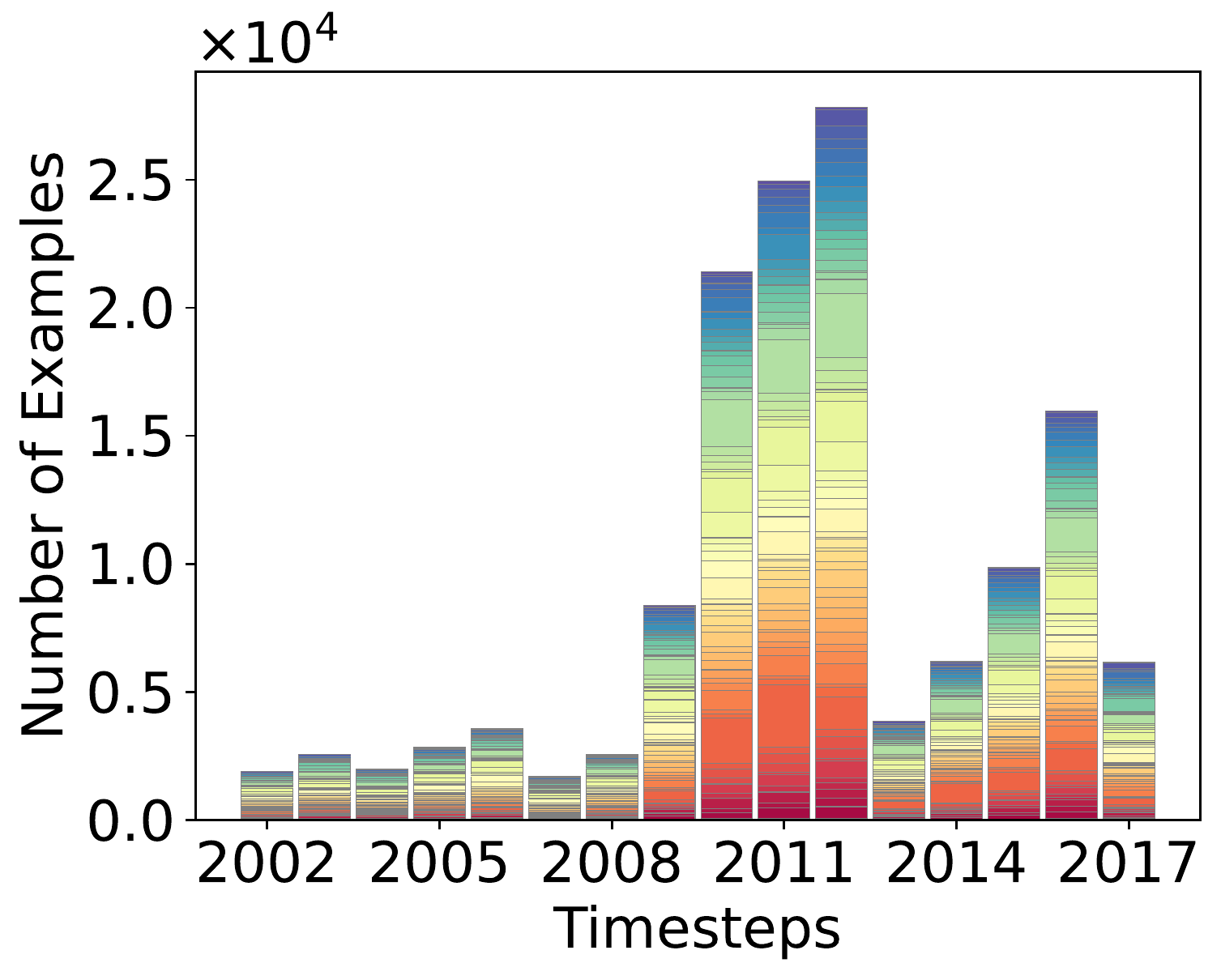}
    \caption{\revision{FMoW-Time}}
\end{subfigure}
\begin{subfigure}[c]{0.31\textwidth}
		\centering
\includegraphics[width=\textwidth]{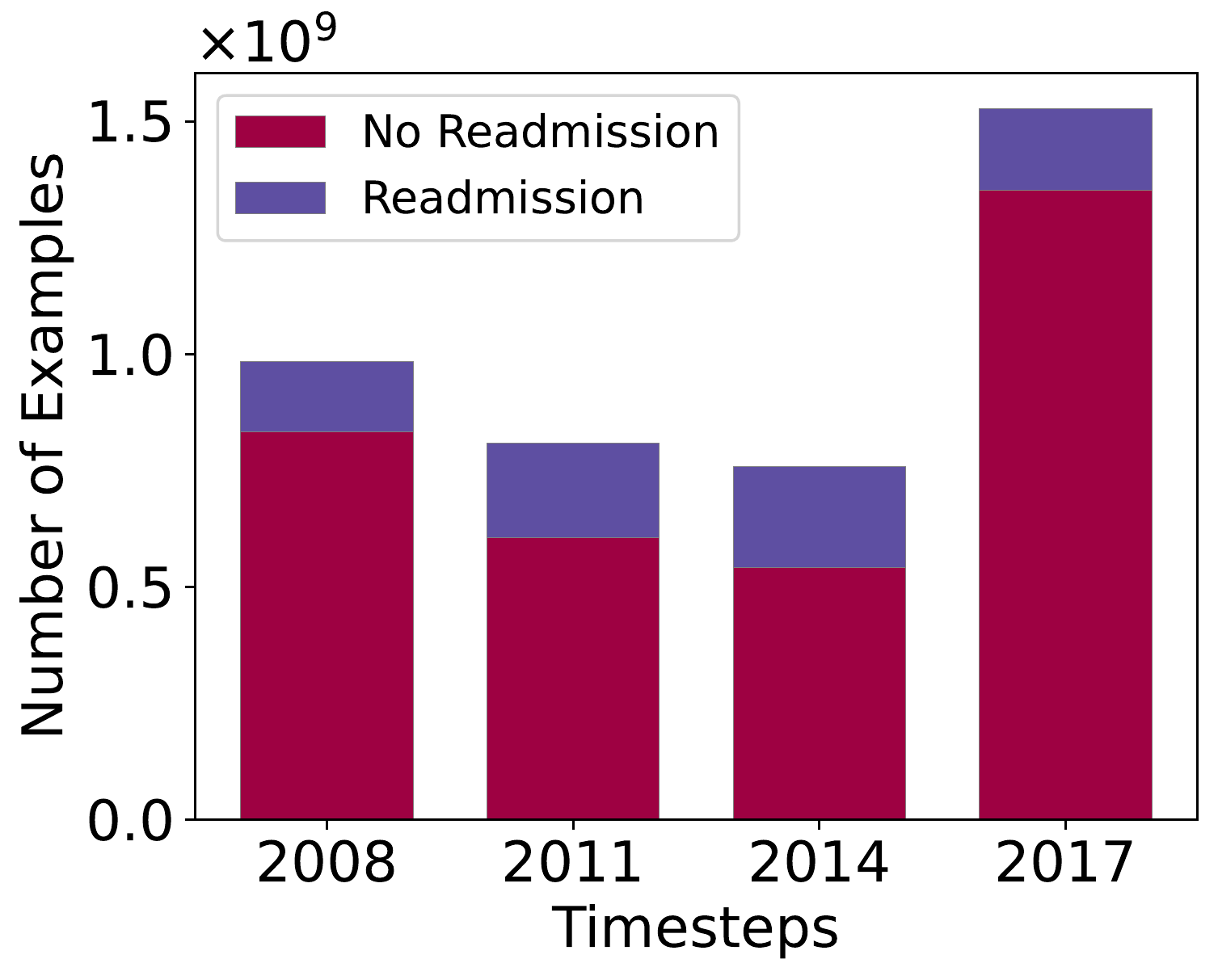}
    \caption{\revision{MIMIC-Readmission}}
\end{subfigure}
\begin{subfigure}[c]{0.31\textwidth}
		\centering
\includegraphics[width=\textwidth]{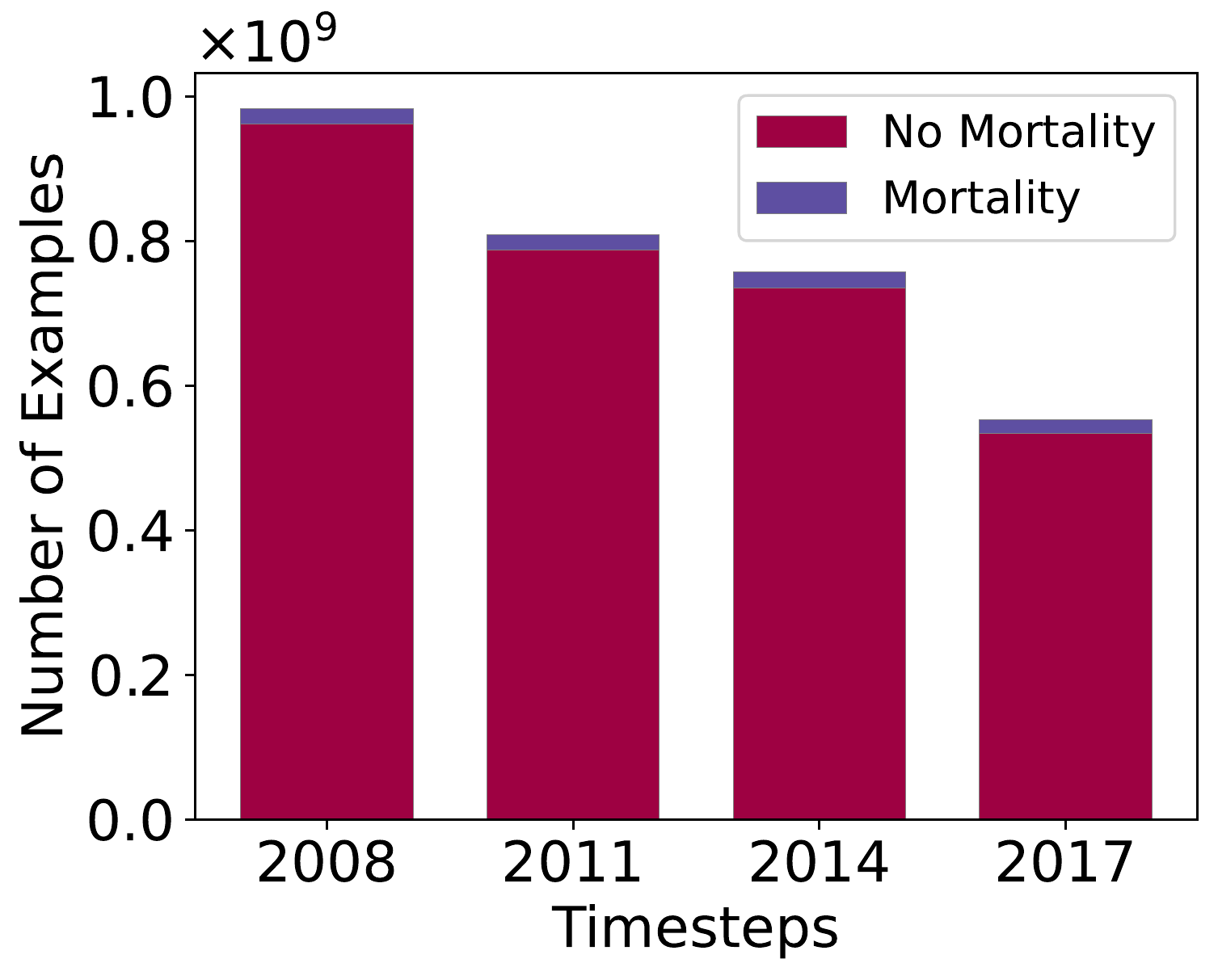}
    \caption{\revision{MIMIC-Mortality}}
\end{subfigure}
\begin{subfigure}[c]{0.31\textwidth}
		\centering
\includegraphics[width=\textwidth]{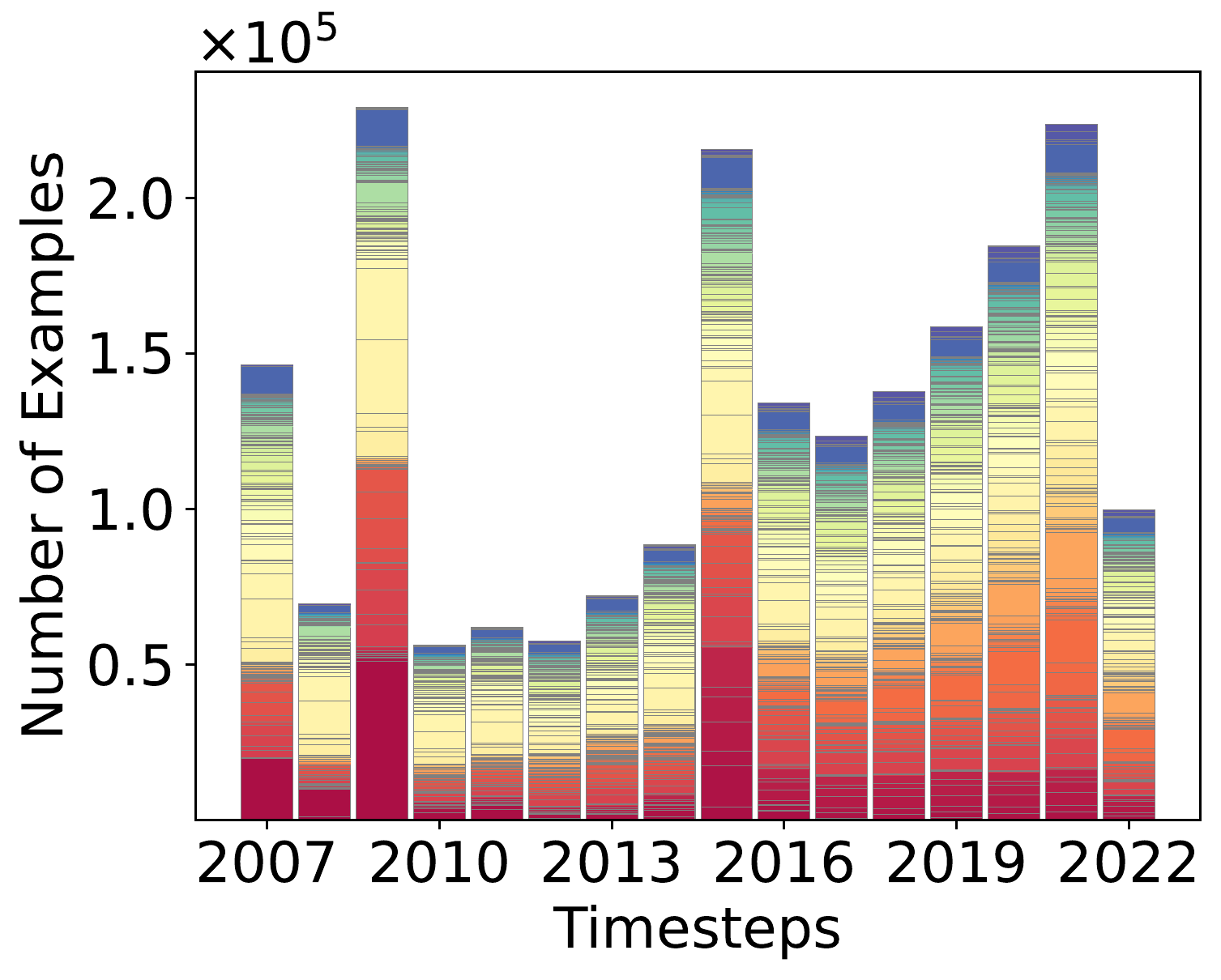}
    \caption{\revision{arXiv}}
\end{subfigure}
\begin{subfigure}[c]{0.31\textwidth}
		\centering
\includegraphics[width=\textwidth]{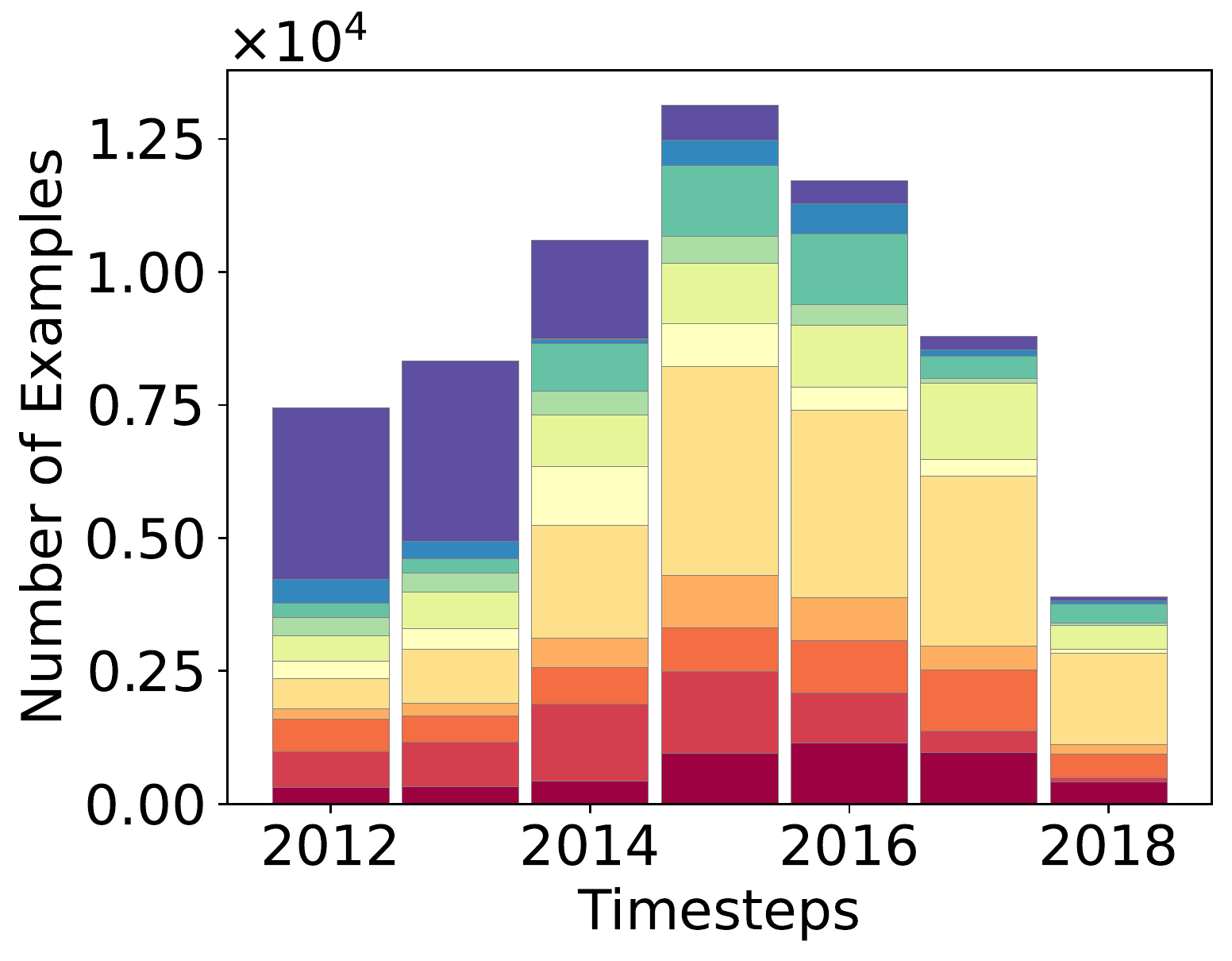}
    \caption{\revision{Huffpost}}
\end{subfigure}
\caption{\revision{Label distributions of all Wild-Time classification datasets over time. Note that we include legends for datasets with less than 8 classes due to space limitations.}}
\label{fig:label_dist}
\vspace{-1em}
\end{figure}

\revision{For each Wild-Time dataset, we plot the label distributions over time in Figure \ref{fig:label_dist}. We observe that the label distributions change over time for all Wild-Time datasets, as this is a naturally-occurring shift that we aim to tackle with the Wild-Time benchmark.}

\begin{table}[h]
\small
\centering
\caption{Performance drops of ERM with different splits under Eval-Fix setting. In the standard split, we train the model on timestamps before the split timestamp, and evaluate the model in the future timestamps. In the mixed split, we merge the training data from all timestamps, and evaluate the model on the original OOD set. The large gap between standard split and mixed split indicates that the performance drops between ID and OOD shown in Table~\ref{tab:gap} in the main paper are not caused by the difficulty of the examples from OOD timestamps.}
\begin{tabular}{lcc|cc}
\toprule
\multirow{2}{*}{Dataset} & \multicolumn{2}{c|}{Standard Split} & \multicolumn{2}{c}{Mixed Split}\\
& OOD Avg. & OOD Worst & OOD Avg. & OOD Worst \\
\midrule
Yearbook & 81.98\% & 69.62\% & 94.57\% & 78.57\% \\
FMoW-Time & 54.07\% & 46.00\% & 57.80\% & 52.00\% \\
MIMIC-Mortality & 72.89\% & 65.80\% & 91.00\% & 88.67\% \\
MIMIC-Readmission & 61.33\% & 59.46\% & 57.18\% & 54.84\% \\
Huffpost & 70.42\% & 68.71\% & 78.11\% & 76.87\% \\
arXiv & 45.94\% & 44.09\% & 52.12\% & 50.57\% \\
\bottomrule
\end{tabular}
\label{table:dataset-split}
\end{table}

\subsection{Temporal Adaptation of Invariant Learning Methods}
\label{sec:app_temporal_invariant_method}
In this section, we provide additional analysis for temporal adaptation, including the analysis of the effectiveness of temporal adaptation, the effect of time window size, and the comparison between overlapping and non-overlapping substreams.

\subsubsection{Temporal Adaptation Improves Performance}
We compare the temporal adapted invariant learning approaches with the original approaches. The results are listed in Table~\ref{tab:results_temporal_adaptation}. We observe that temporal adaptation indeed shows improved performance over vanilla invariant learning approaches, verifying the efficacy of the proposed strategy.

\begin{table*}[t]
\small
\caption{The vanilla versus temporal adapted invariant learning performance of each method evaluated on Wild-Time under the Eval-Fix setting.}
\vspace{-1em}
\label{tab:results_temporal_adaptation}
\begin{center}
\resizebox{\columnwidth}{!}{\setlength{\tabcolsep}{2.5mm}{
\begin{tabular}{l|cc|cc|cc}
\toprule
\multirow{3}{*}{} & \multicolumn{2}{c|}{Yearbook} & \multicolumn{2}{c|}{FMoW-Time} & \multicolumn{2}{c}{MIMIC-Readmission} \\
& \multicolumn{2}{c|}{(Accuracy (\%) $\uparrow$)} & \multicolumn{2}{c|}{(Accuracy (\%) $\uparrow$)} & \multicolumn{2}{c}{(Accuracy (\%) $\uparrow$)}\\
& OOD Avg. &  OOD Worst & OOD Avg. & OOD Worst & OOD Avg. &  OOD Worst \\\midrule
GroupDRO & 76.19 (1.58) & 59.61 (1.09) & \revision{42.54 (0.39)} & \revision{35.17 (1.08)}  & 55.69 (3.53) & 54.18 (2.79) \\
GroupDRO-T  & \textbf{77.06 (1.67)} & \textbf{60.96 (1.83)} & \revision{\textbf{43.87 (0.55)}} & \revision{\textbf{36.60 (1.25)}} & \textbf{56.12 (4.35)} & \textbf{53.12 (4.41)}  \\\midrule
CORAL & 76.29 (1.75) & 58.54 (2.91) & \revision{48.96 (0.23)} & \revision{40.17 (0.89)} & 56.62 (3.21) & 54.08 (3.50)  \\
CORAL-T  & \textbf{77.53 (2.15)} & \textbf{59.34 (1.46)} & \revision{\textbf{49.43 (0.38)}} & \revision{\textbf{41.23 (0.78)}}  & \textbf{57.31 (4.45)} & \textbf{54.69 (4.36)} \\\midrule
IRM & 77.08 (2.05) & 63.79 (1.27) & \revision{\textbf{45.25 (1.01)}} & \revision{\textbf{38.73 (0.69)}} & \textbf{57.89 (2.76)} & \textbf{53.02 (2.53)} \\
IRM-T  & \textbf{80.46 (3.53)} & \textbf{64.42 (4.38)} & \revision{45.00 (1.18)} & \revision{37.67 (1.17)} & 56.53 (3.36) & 52.67 (5.17) \\\midrule
\multirow{3}{*}{}  & \multicolumn{2}{c|}{MIMIC-Mortality} &\multicolumn{2}{c|}{HuffPost} & \multicolumn{2}{c}{arXiv} \\
& \multicolumn{2}{c|}{(AUC (\%) $\uparrow$)} & \multicolumn{2}{c|}{(Accuracy (\%) $\uparrow$)} & \multicolumn{2}{c}{(Accuracy (\%) $\uparrow$)} \\
& OOD Avg. & OOD Worst & OOD Avg. &  OOD Worst & OOD Avg. & OOD Worst\\\midrule
GroupDRO  & 74.93 (3.17) & 70.58 (3.46) & 68.33 (0.88) & 67.42 (1.27) & 37.37 (1.09) & 36.09 (0.93)\\
GroupDRO-T & \textbf{76.88 (4.74)} & \textbf{71.40 (6.84)} & \textbf{69.53 (0.54)} & \textbf{67.68 (0.78)} & \textbf{39.06 (0.54)} & \textbf{37.18 (0.52)}\\\midrule
CORAL  & 76.83 (2.70) & 64.62 (5.58) &  \textbf{70.64 (0.43)} & 67.82 (1.16) & 40.82 (1.16) & 38.16 (0.62)\\
CORAL-T  & \textbf{77.98 (2.57)} & \textbf{64.81 (10.8)} & 70.05 (0.63) & \textbf{68.39 (0.88)} & \textbf{42.32 (0.60)} & \textbf{40.31 (0.61)} \\\midrule
IRM & \textbf{76.25 (5.87)} & 69.91 (6.02) & \textbf{71.69 (1.33)} & \textbf{69.49 (1.46)} & 35.07 (0.55) & \textbf{34.22 (0.63)}  \\
IRM-T   & 76.16 (6.32) & \textbf{70.64 (8.99)} & 70.21 (1.05) & 68.71 (1.13) & \textbf{35.75 (0.90)} & 33.91 (1.09) \\
\bottomrule
\end{tabular}}}
\end{center}
\end{table*}

\revision{\subsubsection{Effect of Time Window Size}
We include an ablation in which we report the performance of CORAL-T, GroupDRO-T, and IRM-T when the time window size $L$ (defined in Section 4 of the main paper) is reduced. We report baseline results in Table~\ref{tab:reduce_kt}. We found that reducing $L$ marginally worsens the performance of invariant learning baselines.

\begin{table*}[t]
\small
\caption{\revision{Performance of the temporally adapted invariant learning baselines when decreasing the length of the time windows, $L$. We evaluate under the Eval-Fix setting and report the average and standard deviation (value in parentheses), computed over three random seeds. For Yearbook, performance worsens when $L$ is reduced, but improves for FMoW-Time.}}
\vspace{-1em}
\label{tab:reduce_kt}
\begin{center}
\small
\begin{tabular}{l|p{0.2cm}cc|p{0.2cm}cc}
\toprule
\multirow{3}{*}{} & \multicolumn{3}{c|}{Yearbook} & \multicolumn{3}{c}{FMoW-Time} \\
& \multicolumn{3}{c|}{(Accuracy (\%) $\uparrow$)} & \multicolumn{3}{c}{(Accuracy (\%) $\uparrow$)}\\    
& $L$ & OOD Avg. & OOD Worst & $L$ & OOD Avg. & OOD Worst \\\midrule
\multirow{3}{*}{GroupDRO-T}& 5 & \textbf{77.06 (1.67)} & \textbf{60.96 (1.83)} & 3 & 43.87 (0.55) & 36.60 (1.25)  \\ 
 & 4 & 72.84 (3.04) & 56.05 (0.75) & 2 & \textbf{45.17 (0.83)} & \textbf{36.97 (1.00)} \\ 
& 2 & 73.42 (2.29) & 56.99 (2.54) & n/a & n/a & n/a \\\midrule 
\multirow{3}{*}{CORAL-T}  & 5 & \textbf{77.53 (2.15)} & \textbf{59.34 (1.46)} & 3 & 49.43 (0.38) & 41.23 (0.78) \\ 
 & 4 & 77.09 (1.56) & 59.17 (1.89) & 2 & \textbf{49.67 (0.49)} & \textbf{41.63 (0.32)}\\ 
 & 2 & 76.92 (1.07) & 59.26 (1.38) & n/a & n/a & n/a \\\midrule 
\multirow{3}{*}{IRM-T} & 5 & \textbf{80.46 (3.53)} & \textbf{64.42 (4.38)} & 3 & 45.00 (1.18) & 37.67 (1.17) \\ 
& 4 & 79.56 (3.12) & 63.70 (3.85) & 2 & \textbf{48.50 (0.10)} & \textbf{40.63 (0.59)} \\ 
 & 2 & 79.47 (2.69) & 63.65 (3.91) & n/a & n/a & n/a \\\bottomrule 
\end{tabular}
\vspace{-1em}
\end{center}
\end{table*}
}

\subsubsection{Non-Overlapping Time Windows}
In the proposed temporal adaptations of the invariant learning methods (CORAL-T, GroupDRO-T, IRM-T), we use overlapping time windows to capture the gradual temporal distribution shift. Here, we run all invariant learning baselines using non-overlapping windows, and report the OOD performance in Table~\ref{tab:results_nonoverlapping_windows}. For the Yearbook, Huffpost, arXiv, MIMIC-Mortality, and MIMIC-Readmission, invariant learning baselines generally obtained better performance using overlapping time windows. Since, on the aggregate, using overlapping time windows resulted in better performance, we keep the results using non-overlapping windows in Table~\ref{tab:results_offline_main} of the main paper. 


\begin{table*}[t]
\small
\caption{\revision{Performance of CORAL-T, GroupDRO-T, and IRM-T baselines when trained on non-overlapping time substreams. OL: Overlapping; NOL: Non-overlapping}}

\vspace{-1em}
\label{tab:results_nonoverlapping_windows}
\begin{center}
\resizebox{\columnwidth}{!}{\setlength{\tabcolsep}{2.5mm}{
\begin{tabular}{ll|cc|cc|cc}
\toprule
\multicolumn{2}{c|}{\multirow{2}{*}{}} & \multicolumn{2}{c|}{Yearbook} & \multicolumn{2}{c|}{FMoW-Time} & \multicolumn{2}{c}{MIMIC-Readmission} \\
&  & \multicolumn{2}{c|}{(Accuracy (\%) $\uparrow$)} & \multicolumn{2}{c|}{(Accuracy (\%) $\uparrow$)} & \multicolumn{2}{c}{(Accuracy (\%) $\uparrow$)}\\
& & OOD Avg. & OOD Worst & OOD Avg. & OOD Worst & OOD Avg. & OOD Worst \\\midrule
\multirow{2}{*}{CORAL-T} & OL & \textbf{77.53 (2.15)} & \textbf{59.34 (1.46)} & 49.43 (0.38) & 41.23 (0.78) &  \textbf{57.31 (4.45)} & \textbf{54.69 (4.36)}\\
&  NOL  & 75.97 (0.63) & 57.47 (0.29) & \textbf{49.93 (0.64)} & \textbf{42.13 (0.96)} & 54.86 (2.93) & 51.44 (4.63) \\\midrule
\multirow{2}{*}{GroupDRO-T} & OL & \textbf{77.06 (1.67)} & \textbf{60.96 (1.83)} & 43.87 (0.55) & 36.60 (1.25) & \textbf{56.12 (4.35)} & \textbf{53.12 (4.41)} \\
& NOL & 76.94 (1.87) & 58.58 (1.82) & \textbf{48.67 (0.57)} & \textbf{45.50 (0.62)} & 53.96 (3.03) & 50.47 (4.43) \\\midrule
\multirow{2}{*}{IRM-T} & OL & \textbf{80.46 (3.53)} & \textbf{64.42 (4.38)} & 45.00 (1.18) & 37.67 (1.17) & \textbf{56.53 (3.36)} & \textbf{52.67 (5.17)} \\
& NOL & 77.21 (2.34) & 59.44 (1.72) & \textbf{49.67 (0.40)} & \textbf{42.50 (1.08)} & 54.31 (3.67) & 51.08 (5.23)  \\\midrule
\multicolumn{2}{c|}{\multirow{2}{*}{}}  & \multicolumn{2}{c|}{MIMIC-Mortality} &\multicolumn{2}{c|}{HuffPost} & \multicolumn{2}{c}{arXiv}\\
&  & \multicolumn{2}{c|}{(AUC (\%) $\uparrow$)}  & \multicolumn{2}{c|}{(Accuracy (\%) $\uparrow$)} & \multicolumn{2}{c}{(Accuracy (\%) $\uparrow$)}\\
& & OOD Avg. &  OOD Worst & OOD Avg. & OOD Worst  &  OOD Avg. & OOD Worst \\\midrule
\multirow{2}{*}{CORAL-T} & OL & \textbf{77.98 (2.57)} & \textbf{64.81 (10.8)} & \textbf{70.05 (0.63)} & \textbf{68.39 (0.88)} & \textbf{42.32 (0.60)} & \textbf{40.31 (0.61)}\\
& NOL & 71.57 (11.7) & 65.77 (15.3) & 68.11 (1.40) & 66.94 (1.50) & 42.07 (0.72) & 40.10 (0.72) \\\midrule
\multirow{2}{*}{GroupDRO-T} & OL & \textbf{76.88 (4.74)} & \textbf{71.40 (6.84)} & \textbf{69.53 (0.54)} & \textbf{67.68 (0.78)} & \textbf{39.06 (0.54)} & \textbf{37.18 (0.52)} \\
& NOL & 72.78 (10.7) & 67.40 (14.1) & 68.41 (0.41) & 67.26 (0.49) & 36.07 (1.35) & 33.98 (1.46)\\\midrule
\multirow{2}{*}{IRM-T} & OL & \textbf{76.17 (6.32)} & \textbf{70.64 (8.99)} & \textbf{70.21 (1.05)} & \textbf{68.71 (1.13)} & 35.75 (0.90) & 33.91 (1.09)\\
& NOL & 73.08 (9.99) & 67.69 (13.1) & 69.58 (0.79) & 68.16 (0.64) & \textbf{38.85 (0.44)} & \textbf{36.86 (0.42)} \\
\bottomrule
\end{tabular}}}
\vspace{-1em}
\end{center}
\end{table*}

\revision{\subsection{Effect of Model Backbones}
We investigate the effect of different models backbones on a Wild-Time image dataset (FMoW-Time) and text dataset (arXiv). Specifically, we use ResNet18 and ResNet50 backbones for FMoW-Time, and BERT and ALBERT backbones for arXiv. We report the performance of ERM and two representative invariant learning and continual learning approaches -- LISA and Fine-tuning -- under the Eval-Fix setting in Table~\ref{tab:backbone}.

\begin{table}[h]
\small
\caption{\revision{Performance comparison w.r.t. Different backbones.}}
\label{tab:backbone}
    \centering
    \begin{tabular}{l|c|c|c|c}
    \toprule
         & Backbone & ERM & Fine-tuning & LISA \\ \midrule
       \multirow{4}{*}{FMoW-Time}  & ResNet18 & 47.95 (0.39) & 40.95 (0.50) & 47.59 (0.36) \\ \cmidrule{2-5}
         & ResNet50 & 52.91 (0.46) & 45.68 (0.79) & 53.17 (0.85) \\ \cmidrule{2-5}
         & DenseNet101 & 54.07 (0.25) & 44.22 (0.56) & 52.33 (0.42) \\ \midrule\midrule
        \multirow{4}{*}{arXiv} & DistilBERT & 45.94 (0.97) & 50.31 (0.39) & 47.82 (0.47) \\ \cmidrule{2-5}
         & BERT & 47.51 (1.20) & 50.99 (0.52) & 49.05 (1.01) \\ \cmidrule{2-5}
         & ALBERT & 45.25 (0.65) & 49.76 (0.69) & 46.01 (0.52) \\ \bottomrule
    \end{tabular}
\end{table}

These results are consistent with our findings that neither invariant learning nor continual learning approaches make models more robust to temporal distribution shift, even with different backbones.

}

\revision{\subsection{Reducing the Number of Training Examples}
\label{sec:app_reduce_training_examples}
We analyze the performance of all baselines when reducing the number of training examples. Specifically, under the Eval-Fix setting, we randomly allocate 30\% of the data at each training timestamp as training, rather than 90\% in our original results (c.f., Table~\ref{tab:results_offline_main} in the main paper). We report all results in Table~\ref{tab:results_reduce_data}. We observe that ERM still outperforms invariant learning and continual learning approaches, corroborating our findings in the main paper.}

\begin{table*}[t]
\small
\caption{\revision{Performance of all baselines when reducing the amount of training data. We randomly allocate 30\% of the data at each timestamp to training, rather than 90\% in our original benchmark.}}
\vspace{-1em}
\label{tab:results_reduce_data}
\begin{center}
\resizebox{\columnwidth}{!}{\setlength{\tabcolsep}{2.5mm}{
\begin{tabular}{l|ccc|ccc}
\toprule
\multirow{3}{*}{} & \multicolumn{3}{c|}{Yearbook} & \multicolumn{3}{c}{FMoW-Time}  \\
& \multicolumn{3}{c|}{(Accuracy (\%) $\uparrow$)} & \multicolumn{3}{c}{(Accuracy (\%) $\uparrow$)}\\    
& ID Avg. & OOD Avg. &  OOD Worst & ID Avg. &  OOD Avg. & OOD Worst \\\midrule
Fine-tuning & 46.29 (1.17) & 52.00 (5.00) & 44.10 (2.15) & 40.30 (0.14) & 39.34 (0.57) & 30.91 (0.29) \\
EWC  & 45.50 (0.00) & 48.84 (0.01) & 42.86 (0.01) & 40.89 (0.66) & 39.99 (0.66) & 31.24 (0.43) \\
SI & 49.42 (6.73) & 47.03 (17.2) & 45.52 (4.61) & 41.03 (0.19) & 39.65 (0.53) & 30.91 (0.29) \\
A-GEM & 45.50 (0.00) & 46.98 (3.57) & 44.95 (3.45) & 40.52 (0.56) & 39.60 (0.41) & 31.05 (0.38) \\\midrule
ERM  & 93.96 (1.72) & \textbf{77.05 (5.13)} & \textbf{60.72 (2.92)} & 51.74 (0.77) & \textbf{50.63 (0.56)} & 40.29 (0.94) \\
GroupDRO-T  & 77.56 (11.5) & 60.45 (7.10) & 47.03 (7.99) & 40.16 (0.68) & 39.52 (0.86) & 31.62 (0.29) \\
mixup & 92.88 (2.35) & 77.31 (2.60) & 61.56 (3.00) & 54.23 (0.26) & 53.77 (0.23) & \textbf{43.04 (0.77)}  \\
LISA & 92.51 (4.03) & 74.17 (5.22) & 57.39 (1.77) & 51.34 (0.14) & 50.76 (0.83) & 40.12 (1.42) \\
CORAL-T & 75.35 (19.7) & 59.66 (10.1) & 44.00 (10.58) & 44.15 (0.93) & 43.85 (0.46) & 34.59 (0.25) \\
IRM-T & 77.04 (17.7) & 60.45 (7.10) & 47.03 (7.99) & 45.19 (0.94) & 44.11 (1.23) & 36.26 (1.36) \\\midrule
\multirow{3}{*}{} & \multicolumn{3}{c|}{MIMIC-Readmission} & \multicolumn{3}{c}{MIMIC-Mortality} \\
& \multicolumn{3}{c|}{(Accuracy (\%) $\uparrow$)} & \multicolumn{3}{c}{(AUC (\%) $\uparrow$)} \\
& ID Avg. & OOD Avg. &  OOD Worst & ID Avg. &  OOD Avg. & OOD Worst \\\midrule
Fine-tuning & 74.22 (2.96) & 64.20 (3.73) & 62.33 (5.25) & 87.92 (0.92) & 59.20 (0.71) & 50.00 (1.35) \\
EWC  & 74.49 (1.41) & 66.75 (0.99) & 65.93 (1.26) & 87.94 (0.08) & 60.07 (2.45) & 51.21 (3.11) \\
SI & 74.22 (2.96) & 64.20 (3.73) & 62.33 (5.25) & 87.92 (0.92) & 59.20 (0.71) & 50.00 (1.35) \\
A-GEM & 80.58 (0.14) & \textbf{69.90 (0.01)} & \textbf{68.48 (0.01)} & 70.27 (17.5) & 53.68 (3.76) & 48.00 (1.83) \\\midrule
ERM & 70.49 (2.47) & 55.28 (2.54) & 51.69 (5.47) & 89.53 (0.82) & 71.06 (7.63) & 65.76 (10.2) \\
GroupDRO-T  & 74.36 (2.65) & 59.90 (15.3) & 54.92 (21.3) & 89.48 (0.85) & \textbf{73.28 (7.58)} & 68.25 (9.97) \\
mixup  & 71.82 (3.61) & 41.57 (1.12) & 30.29 (0.00) & 89.48 (1.14) & 71.32 (8.55) & 65.65 (11.4)  \\
LISA  & 67.50 (2.22) & 40.48 (0.68) & 30.29 (0.00) & 90.01 (0.32) & 73.37 (10.5) & \textbf{68.97 (14.5)} \\
CORAL-T  & 74.48 (1.72) & 45.00 (4.50) & 34.58 (7.05) & 89.12 (1.43) & 71.55 (10.4) & 66.01 (13.6) \\
IRM-T  & 74.36 (1.90) & 52.00 (14.4) & 44.24 (20.0) & 88.24 (1.59) & 73.13 (10.1) & 68.66 (13.5) \\\midrule
\multirow{3}{*}{} &\multicolumn{3}{c|}{HuffPost} & \multicolumn{3}{c}{arXiv} \\
& \multicolumn{3}{c|}{(Accuracy (\%) $\uparrow$)} & \multicolumn{3}{c}{(Accuracy (\%) $\uparrow$)} \\
& ID Avg. & OOD Avg. &  OOD Worst & ID Avg. &  OOD Avg. & OOD Worst \\\midrule
Fine-tuning & 13.11 (1.03) & 14.12 (3.27) & 12.89 (2.46) & 50.34 (0.13) & \textbf{48.88 (0.26)} & \textbf{46.72 (0.25)} \\
EWC & 13.26 (1.27) & 13.65 (1.51) & 12.37 (1.03) & 50.31 (0.17) & 48.56 (0.05) & 46.38 (0.11) \\
SI & 13.06 (1.05) & 14.22 (3.26) & 12.95 (2.44) & 50.35 (0.13) & \textbf{48.88 (0.25)} & \textbf{46.72 (0.26)} \\
A-GEM & 13.07 (0.69) & \textbf{15.53 (2.18)} & \textbf{13.43 (2.03)} & 50.36 (0.18) & 48.79 (0.32) & 46.53 (0.39) \\\midrule
ERM & 15.86 (1.45) & 12.32 (2.64) & 11.32 (2.08) & 53.55 (0.21) & 46.07 (0.53) & 44.16 (0.50) \\
GroupDRO-T & 14.23 (1.05) & 11.82 (1.07) & 11.03 (0.76) & 50.01 (0.03) & 39.71 (0.63) & 37.79 (0.65) \\
mixup & 15.49 (0.92) & 13.35 (1.45) & 11.92 (1.08) & 52.66 (0.13) & 45.98 (0.47) & 44.00 (0.45) \\
LISA & 14.95 (0.68) & 13.26 (3.62) & 11.93 (2.54) & 49.17 (0.43) & 47.66 (0.27) & 45.71 (0.30) \\
CORAL-T & 16.56 (0.56) & 13.15 (5.17) & 11.82 (4.29) & 52.60 (0.06) & 42.72 (0.27) & 40.72 (0.24) \\
IRM-T & 14.06 (0.65) & 11.39 (0.41) & 11.05 (0.47) & 46.20 (0.12) & 35.85 (0.70) & 34.14 (0.75) \\
\bottomrule
\end{tabular}}}
\vspace{-2em}
\end{center}
\end{table*}

\section{Datasets without Gradual Temporal Distribution Shifts}
 \label{sec:app_additional_data}
In this section, we discuss two additional datasets that were not included in Wild-Time. These datasets do not satisfy the criteria discussed in Section~\ref{sec:criteria}. 
\subsection{Drug-BA}
\label{app:data:tdc}
\subsubsection{Dataset Setup}
\textbf{Problem Setting.} The task is predicting the binding affinity of candidate drugs to their target molecules. The input $x$ contains molecular information of both the drug and target molecules, and the label $y$ is the binding affinity value.

\textbf{Data.} The Therapeutics Data Commons (TDC) benchmark (MIT license). TDC offers the BindingDB dataset, which was curated from BindingDB, a public database that features drug-target binding affinities collected from a variety of sources, including patents, journals, and assays. Each entry in BindingDB consists of a small molecule and the corresponding target protein. We exclude data from the year $2021$ in the original TDC benchmark as $2021$ includes only one month's worth of data.

For Eval-Fix, we use the first 4 years ($2013 - 2016$) for training and allocate 4 years ($2017 - 2020$) for testing. For streaming evaluation (Eval-Stream), we treat each year as a single timestamp. We provide the number of examples allocated to ID Train, ID Test, and OOD Test for each timestamp in Table \ref{tab:drug-ba_n}.

\begin{table}
\centering
\small
\caption{
\label{tab:drug-ba_n}
Data subset sizes for the Drug-BA task.}
\begin{tabular}{l|rrr}
\toprule
Year & ID Train & ID Test & OOD Test \\
\midrule
2013 & 9,121 & 2,281 & 11,402 \\
2014 & 16,148 & 4,038 & 20,186 \\
2015 & 24,251 & 6,063 & 30,314 \\
2016 & 23,095 & 5,774 & 28,869 \\
\midrule
2017 & 41,203 & 10,301 & 51,504 \\
2018 & 32,924 & 8,231 & 41,155 \\
2019 & 33,607 & 8,402 & 42,009 \\
2020 & 5,557 & 1,390 & 6,947 \\
\midrule
Eval-Fix split & 72,615 & 18,156 & 141,615 \\
\bottomrule
\end{tabular}
\end{table}

\textbf{Evaluation Metrics.} We use Pearson Correlation Coefficient (PCC), which measures the amount of linear correlations between the true values and the predicted values, to evaluate model performance in predicting drug-target binding affinity. Eval-Stream evaluates performance across the next 3 years, which represents 37.5\% of all timestamps in the entire dataset.

\subsubsection{Baseline Results and Analysis}

\textbf{Experimental Setup.} We use the DeepDTA model from \cite{ozturk2018deepdta}, which achieves state-of-the-art performance on drug target binding affinity prediction by using CNNs to construct high-level representations of a drug and a target. We use the Adam optimizer with a learning rate of $2\times 10^{-5}$ or $5\times 10^{-5}$ (for different baselines) and batch size of 256. Baselines were trained for 5000 iterations under the Eval-Fix setting and for 500 iterations under the Eval-Stream setting. In terms of hyperparameters for CORAL, GroupDRO and IRM, the CORAL penalty, IRM penalty, learning rate, the number of substreams and the size of substreams are set as 0.9, $10^{-3}$, $5\times 10^{-5}$, 3, 2, respectively. Notice that LISA is only applicable to classification problem, thus we do not evaluate LISA on Drug-BA.

\begin{figure}[t]
\centering
\begin{subtable}{\linewidth}
        \centering
        \small
        \begin{tabular}{l|cc|cc}
\toprule
\multirow{2}{*}{Dataset} & \multicolumn{2}{c|}{Standard Split} & \multicolumn{2}{c}{Mixed Split}\\
& OOD Avg. & OOD Worst & OOD Avg. & OOD Worst \\
\midrule
Drug-BA & 0.357 & 0.244 & 0.724 & 0.710\\
\bottomrule
    \end{tabular}
    \caption{Performance drops of ERM with different splits on Drug-BA dataset.}
\end{subtable}  
\begin{subfigure}[c]{0.3\textwidth}
		\centering
\includegraphics[width=\textwidth]{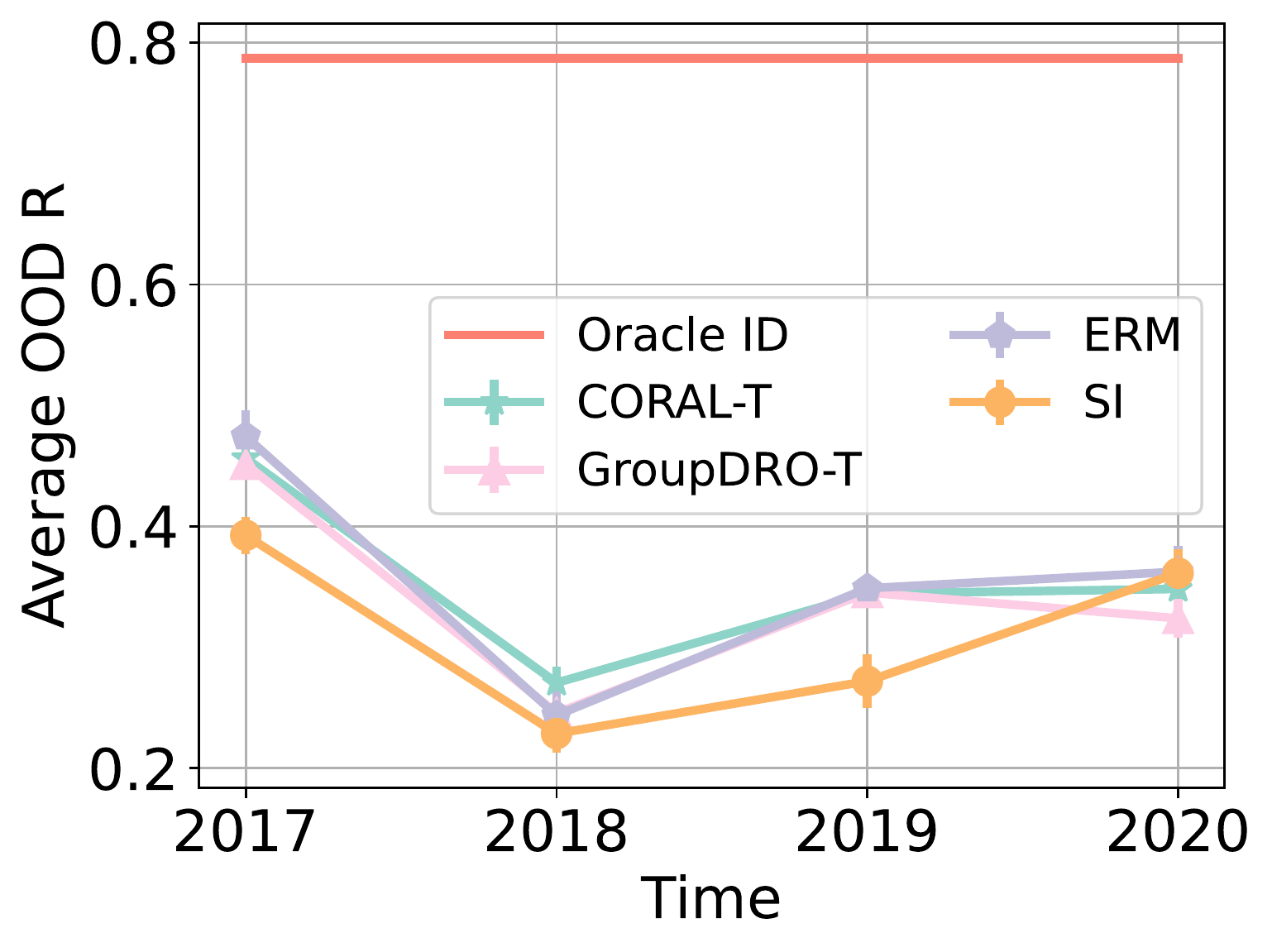}
    \caption{Eval-Fix}
\end{subfigure}
\begin{subfigure}[c]{0.3\textwidth}
		\centering
\includegraphics[width=\textwidth]{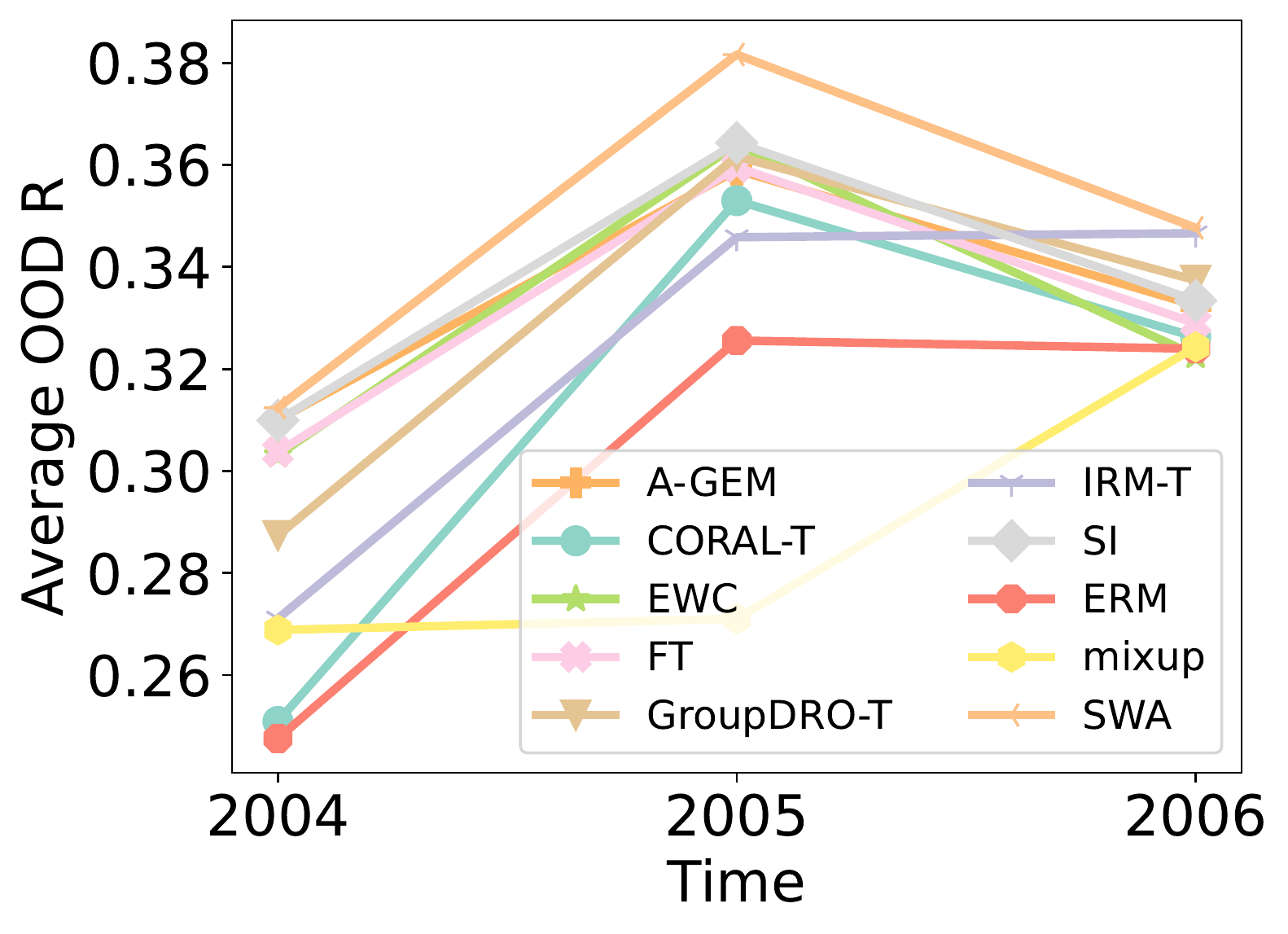}
    \caption{Drug-BA Avg (Eval-Stream)}
\end{subfigure}
\begin{subfigure}[c]{0.3\textwidth}
		\centering
\includegraphics[width=\textwidth]{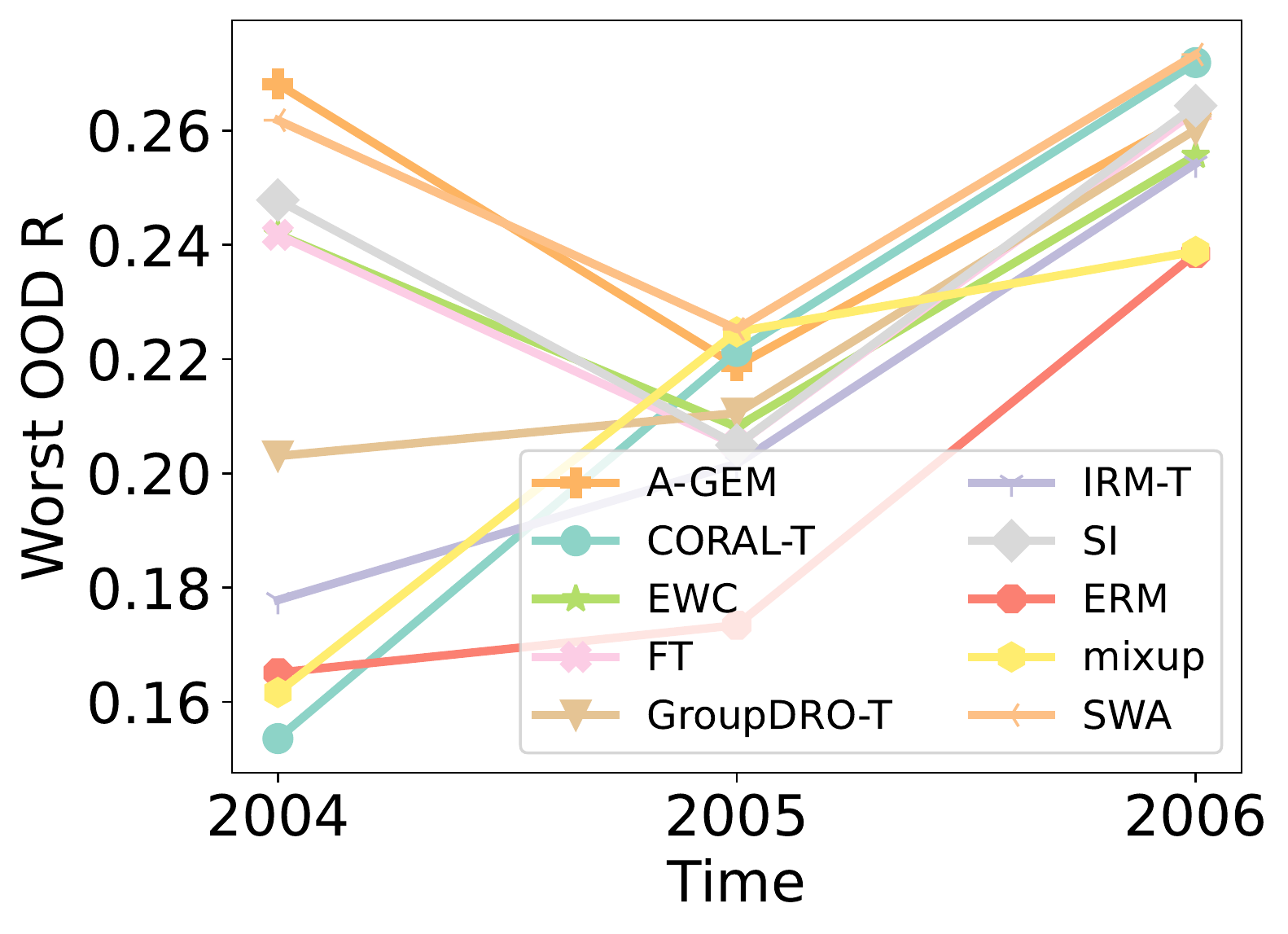}
    \caption{Drug-BA Worst (Eval-Stream)}
\end{subfigure}

\caption{Results on Drug-BA. (b) out-of-distribution performance per test timestamp under Eval-Fix setting; (b) (c): results under Eval-Stream setting.}
\vspace{-1em}
\label{fig:ood_results_drug_all}
\end{figure}

\textbf{Results.} In Drug-BA, similar to Table~\ref{table:dataset-split}, we reported the results of standard splits and mixed splits and the results of Eval-Stream and Eval-Fix in Figure~\ref{fig:ood_results_drug_all}. First, though the performance comparison between standard split and mixed split in the top table of Figure~\ref{fig:ood_results_drug_all}, we observe a significant drop between them, where OOD average performance drops from 0.724 (mixed split) to 0.357 (standard split). Second, the performance per test time in Figure~\ref{fig:ood_results_drug_all}(b) further indicates such a sudden performance drop between Oracle ID and ERM in 2017. These results suggest that the Drug-BA dataset violates our criterion about gradual temporal distribution shifts. Thus, we exclude it in the official Wild-Time benchmark.

\subsubsection{Broader Context} 
Drug discovery brings new candidate medications to potentially billions of people, allowing people to live longer and healthier lives. Traditional methods of drug discovery are via high-throughput, wet-lab experiments \cite{hughes2011principles}, which are expensive, time-consuming, and limited in their ability to search over large sets of drug candidates. Virtual screening is a computational pre-screening process in which the binding activity of a drug candidate with the target protein of a disease is predicted \cite{carpenter2018deep,svetnik2003random}. Recently, there has been a surge of interest in applying machine learning to virtual screening, which can reduce costs and increase the search space to avoid missing potential drug candidates. Recent binding activity prediction models investigate binding pairs between existing compounds and target proteins \cite{martin2019all,ozturk2018deepdta,yao2021improving,yao2021functionally}. In practice, new target proteins or new classes of compounds appear over time, requiring machine learning models that are robust to subtle domain shifts across time.

\subsection{Precipitation} \label{app:data:Precipitation}

\subsubsection{Dataset Setup}
\textbf{Problem Setting.} The task is classifying the precipitation level of a region. The input $x$ is tabular data consisting of 123 meteorological features (1 categorical feature and 122 continuous features). The label $y$ is one of 9 precipitation classes. 

\textbf{Data.} Precipitation is based on the Shifts Precipitation Prediction dataset \cite{malinin2021shifts} (Apache-2.0 license), which collected and processed tabular Precipitation data from the Yandex Precipitation Service to provide a domain shift benchmark for two tasks: temperature prediction (scalar regression) and precipitation classification (multi-class classification). The Shifts Precipitation Prediction dataset contains 10 million 129-column entries, consisting of 123 heterogeneous meteorological features, 4 meta-data attributes (e.g., time, latitude, longitude, and climate type), and 2 targets (temperature and precipitation class). The data is distributed uniformly between September 1, 2018 to September 1, 2019 and is partitioned by both time and climate type. 

We use a subset of the original Shifts Precipitation Prediction dataset, using measurements taken from October 2018 - August 2019. The Precipitation dataset consists of 123 heterogeneous meteorological features, 1 target (precipitation class), and 1 metadata attribute (time). We partition the dataset by month. Our fixed time split (Eval-Fix) uses data from October 2018 - April 2019 (7 months) for ID, and data from May 2019 - August 2019 (4 months) for OOD. For streaming evaluation (Eval-Stream), we treat each month as a single timestamp. We allocate 10\% of the data at each timestamp for test, and the rest for training. For OOD testing, all samples are used. Table \ref{tab:Precipitation} lists the number of examples allocated to ID Train, ID Test, and OOD Test for each timestamp.

\begin{table}
\centering
\small
\caption{Data subset sizes for the Precipitation task.}
\begin{tabular}{l|rrr}
\toprule
Month & ID Train & ID Test & OOD Test \\
\midrule
Sep 2018 & 698,134 & 77,570 & 775,705 \\
Oct 2018 & 714,265 & 79,362 & 793,628 \\
Nov 2018 & 613,885 & 68,209 & 682,095 \\
Dec 2019 & 707,274 & 78,586 & 785,861 \\
Jan 2019 & 739,325 & 82,147 & 821,473 \\
Feb 2019 & 665,745 & 73,971 & 739,717 \\
Mar 2019 & 729,527 & 81,058 & 810,586 \\
Apr 2019 & 691,366 & 76,818 & 768,185 \\
May 2019 & 673,058 & 74,784 & 747,843 \\
Jun 2019 & 548,793 & 60,976 & 609,770 \\
Jul 2019 & 680,152 & 75,572 & 755,725 \\
Aug 2019 & 681,035 & 75,670 & 756,706 \\
\midrule
Eval-Fix split & 4,868,155 & 540,903 & 3,638,229 \\\midrule
\end{tabular}
\label{tab:Precipitation}
\vspace{-2em}
\end{table}

The Shifts Precipitation Prediction dataset is provided as a CSV file.  We ignore the latitude, longitude, and climate type metadata and filter out samples where at least one of the meteorological features is NaN. We shuffle Precipitation measurements in each month, and randomly select 10\% of the measurements in each month as test-ID and allocate the remaining 90\% for training. For OOD testing, all samples in each month are used.

\textbf{Evaluation Metrics.} We evaluate models by their average and worst-time OOD accuracies. The former measures the model's ability to generalize across time, while the latter additionally measures model robustness to trends in seasonal Precipitation patterns.

Eval-Stream evaluates performance across the next 4 months, which represents 33.3\% of all timestamps in the entire dataset and tests a model's robustness to shifting meteorological measurements from seasonal Precipitation changes.

\subsection{Results}
\textbf{Experimental Setup.} We follow \cite{malinin2021shifts} and use a FTTransformer \cite{joseph2021pytorch}, which is well-suited for deep learning with tabular data. We use all default architecture settings for the FTTransformer, except that the deep MLP in our FTTransformer has 2 layers, each of size 32 units, and uses LeakyReLU activation.

We use the Adam optimizer with a fixed learning rate of $10^{-3}$ and train with a batch size of 128. Baselines were trained for 5000 iterations under the Eval-Fix setting and for 500 iterations under the Eval-Stream setting. In terms of hyperparameters for CORAL, GroupDRO and IRM, the CORAL penalty, IRM penalty, learning rate, the number of substreams and the size of substreams are set as 0.9, $1.0$, $10^{-3}$, 3, 4, respectively. 

\begin{figure}[t]
\centering
\begin{subtable}{\linewidth}
        \small
        \centering
        \begin{tabular}{l|cc|cc}
        \toprule
        \multirow{2}{*}{Dataset} & \multicolumn{2}{c|}{Standard Split} & \multicolumn{2}{c}{Mixed Split}\\
        & OOD Avg. & OOD Worst & OOD Avg. & OOD Worst \\
        \midrule
        Precipitation & 46.08\% & 44.15\% & 47.56\% & 45.51\% \\
        \bottomrule
    \end{tabular}
    \caption{Performance drops of ERM with different splits on Precipitation dataset.}
    \label{tab:app_pcpn_splits}
\end{subtable}  
\begin{subfigure}[c]{0.28\textwidth}
		\centering
\includegraphics[width=\textwidth]{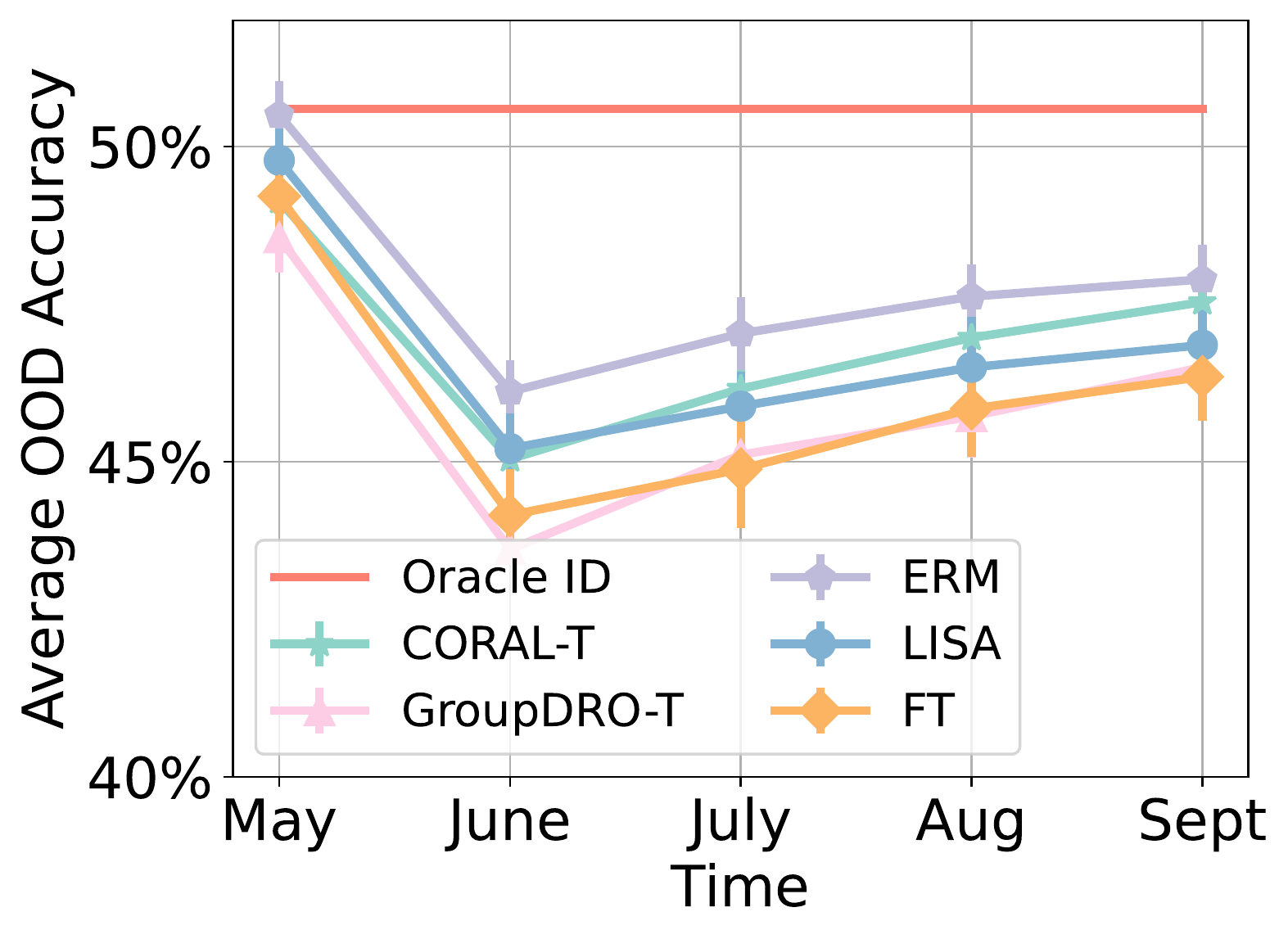}
    \caption{Eval-Fix}
\end{subfigure}
\begin{subfigure}[c]{0.28\textwidth}
		\centering
\includegraphics[width=\textwidth]{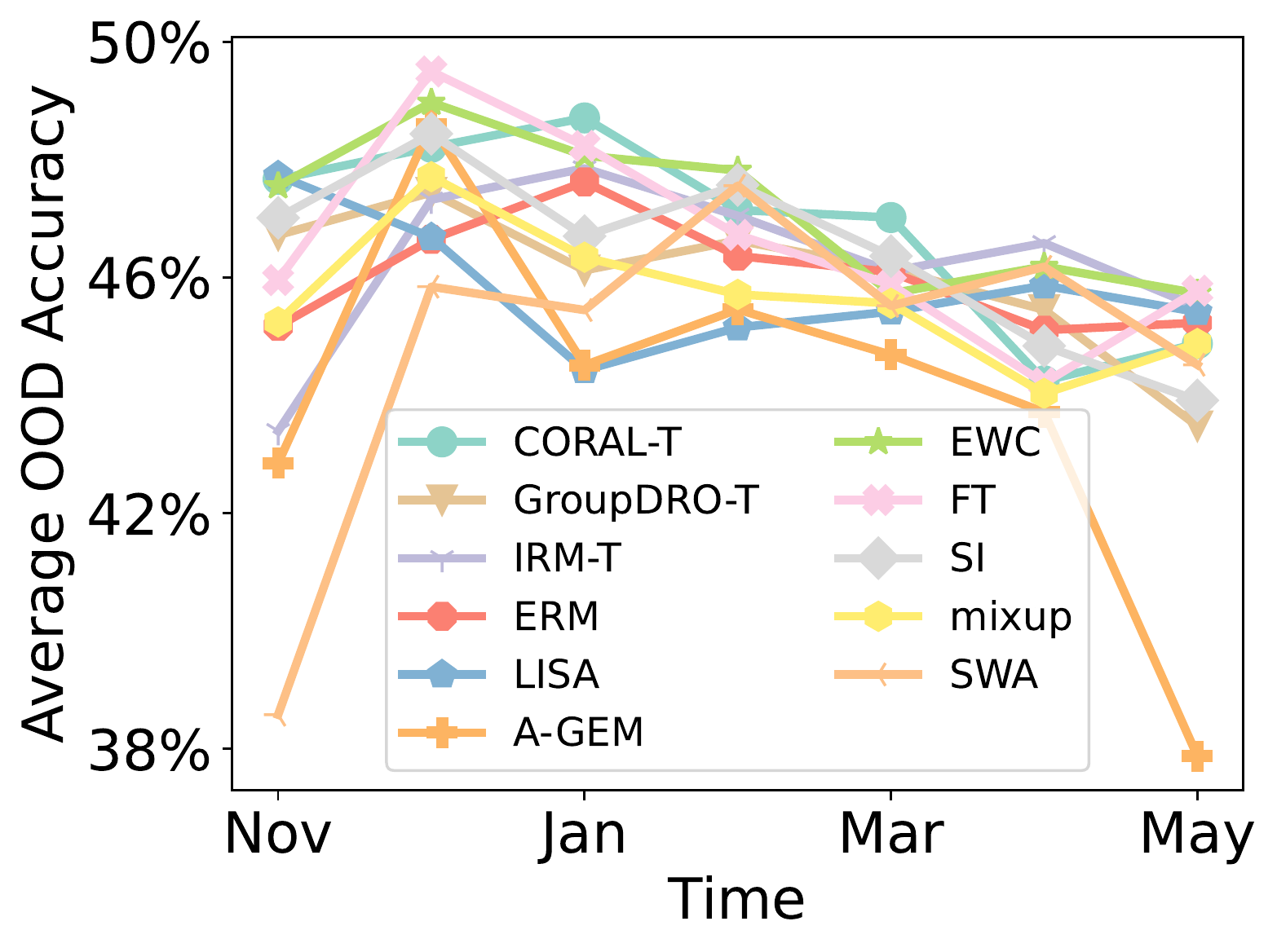}
    \caption{PCPN Avg (Eval-Stream)}
\end{subfigure}
\begin{subfigure}[c]{0.28\textwidth}
		\centering
\includegraphics[width=\textwidth]{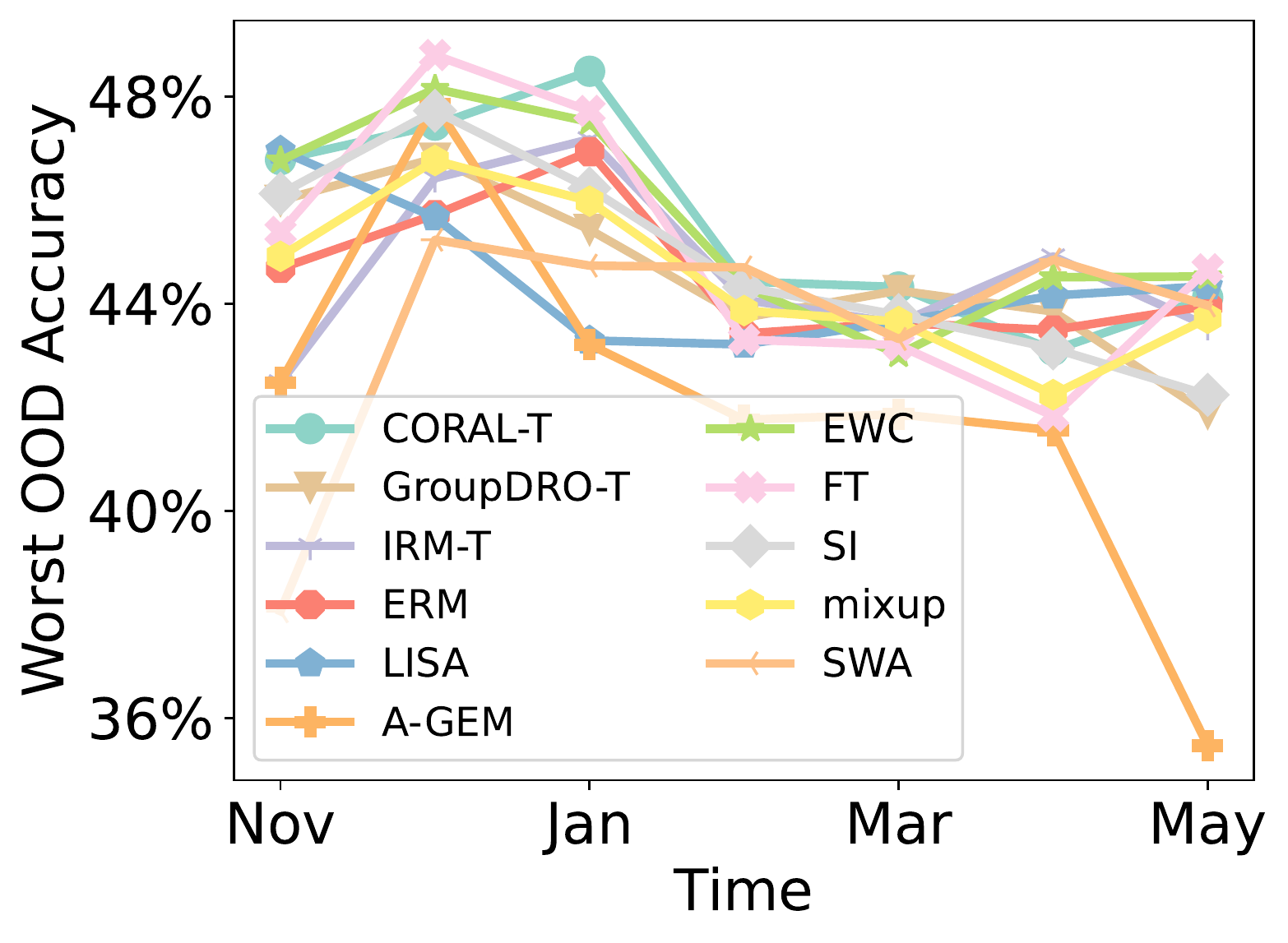}
    \caption{PCPN Worst (Eval-Stream)}
\end{subfigure}

\caption{Results on Precipitation. (b) out-of-distribution performance per test timestamp under Eval-Fix setting; (b) (c): results under Eval-Stream setting.}
\vspace{-1.5em}
\label{fig:ood_results_pcpn_all}
\end{figure}

\textbf{Results.} Similar to Table~\ref{table:dataset-split}, we reported the results of standard splits and mixed splits and the results of Eval-Stream and Eval-Fix in Figure~\ref{fig:ood_results_pcpn_all}. According to the performance between different splits, we can not observe clear performance gaps between standard split and mixed split. Thus, precipitation dataset violates our first dataset selection criterion, and we decide not to include this dataset in the Wild-Time benchmark.

\subsubsection{Broader Context}
Precipitation forecasting enhances public health, safety, and economic prosperity. Extreme Precipitation warnings can save lives and reduce property damage. Forecasts on temperature and precipitation are crucial to agriculture, and hence to traders on commodity markets. On a daily basis, many people use Precipitation forecasts on a daily basis. Precipitation forecasting comprises a large part of the economy: the United States alone spent 5.1 billion on Precipitation forecasting in 2009, resulting in benefits estimated to be 6 times as much \cite{powers2014fostering}.

The Precipitation dataset, which contains heterogeneous tabular data, exhibits data that changes over time due to seasonal changes in Precipitation patterns. In addition, the distribution of the measurement locations are distributed non-uniformly across the planet. Certain climate regions, such as the polar caps, are under-represented, presenting further challenges \cite{malinin2021shifts}.

\section{User Guide and Maintenance Plan}

\subsection{User Guide of Wild-Time}

\textbf{Licenses.} The Wild-Time datasets and baselines are freely available for research purposes. Though Drug-BA and Precipitation are not included in the formal Wild-Time benchmark, we still include these datasets in the Wild-Time package. All code for Wild-Time is available under the MIT license. We list the licenses for each Wild-Time dataset below:
\begin{itemize}[leftmargin=*]
    \item Yearbook: MIT License
    \item FMoW-Time: \href{https://raw.githubusercontent.com/fMoW/dataset/master/LICENSE}{The Functional Map of the World Challenge Public License}
    \item MIMIC-IV (Readmission and Mortality): \href{https://physionet.org/content/mimiciv/view-license/0.4/}{PhysioNet Credentialed Health Data License 1.5.0}
    \item Drug-BA: MIT License
    \item Precipitation: CC BY-NC 4.0
    \item Huffpost: CC0: Public Domain
    \item arXiv: CC0: Public Domain
\end{itemize}

\textbf{Hosting Platform.}
We will use \href{https://github.com/huaxiuyao/Wild-Time}{GitHub} as the hosting platform of code. We provide (1) detailed data preprocessing scripts to help users process the data from scratch, and (2) preprocessed data from each curated dataset except MIMIC-IV.

\textbf{Dependencies.} Wild-Time is built upon Python 3.8+, and depends on PyTorch, PyTorch Tabular, PyTorch Transformers, PyTDC, Huggingface-Hub. Additionally, it uses numpy, scipy, and scikit-learn for data manipulation.

\subsection{Using the Wild-Time Package}
\begin{figure}[h]
\centering\includegraphics[width=0.8\textwidth]{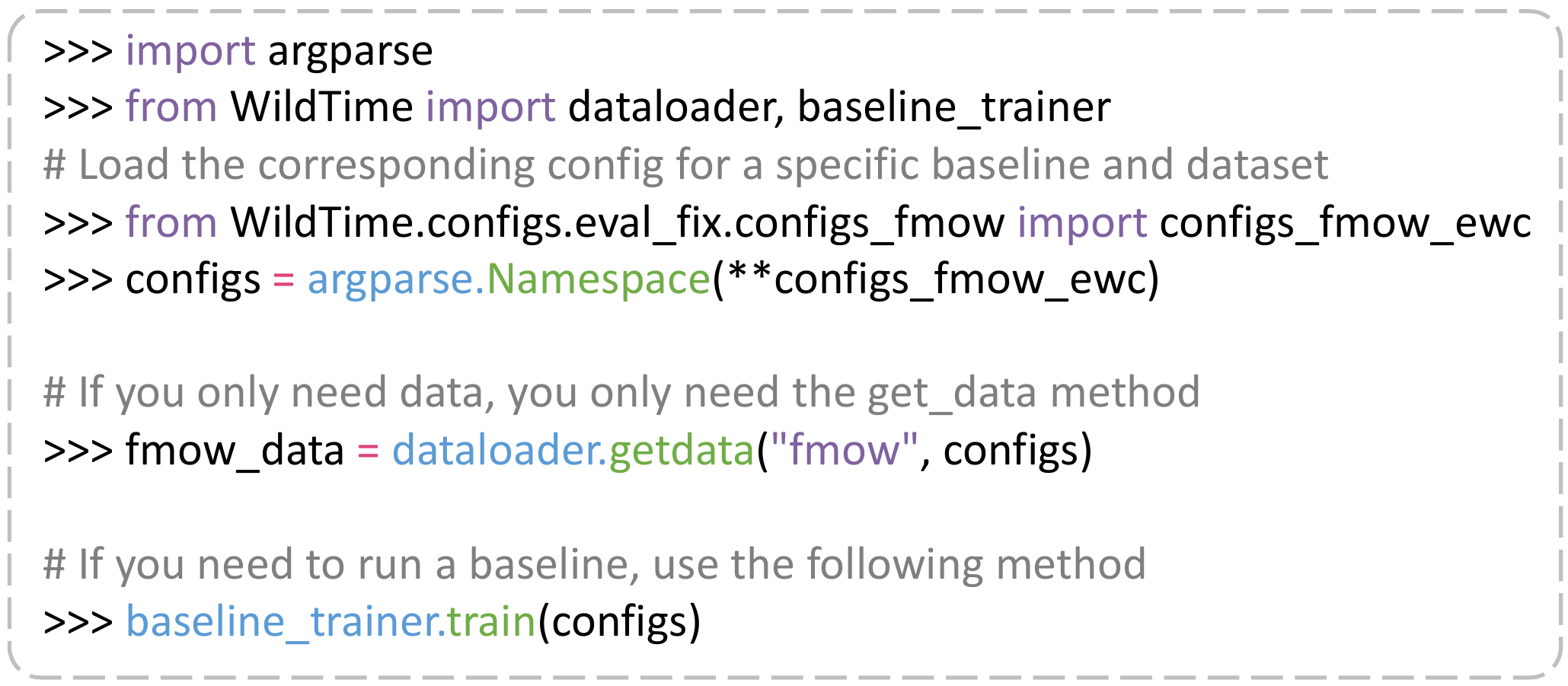}
\caption{Dataset initialization and baseline training.}
\vspace{-1em}
\label{fig:vis_code}
\end{figure}

In this section, we discuss our open-sourced Python package that provides a simple interface to use the Wild-Time benchmark. Our Python package allows the users to use our datasets with a few lines of code. In addition, users can easily construct their own datasets or baselines on top of our package. Specifically, Figure~\ref{fig:vis_code} shows how to use APIs to load the Wild-Time datasets and train a baseline. Beyond the current APIs, we plan to provide standardized evaluation of methods using our dataset in the future.

\subsection{Maintenance Plan}
Wild-Time will be maintained by the authors of this paper. The group can be contacted by raising an issue on the GitHub or by writing to the first authors. The dataset is currently hosted on Google Drive storage. The Wild-Time benchmark may be updated at the discretion of the authors. Updates may include adding more diverse baseline methods, datasets, and tasks, or updating infrastructure to improve efficiency. Updates which correct errors will replace previous versions of the datasets. 

We welcome contributions to the Wild-Time benchmark. Other parties may update the Wild-Time benchmark by submitting a pull request on GitHub. We are releasing the Wild-Time benchmark under the open-source MIT License. We permit other parties to create new datasets from the Wild-Time benchmark, given that the changes are documented and the Wild-Time benchmark is referenced.

\subsection{Author Statement} To the best of our knowledge, the released dataset and benchmark does
not violate any existing licenses. However, if such a violation were to exist, the authors claim responsibility for resolving these issues.

\section{Discussion}

\subsection{Limitations}
One limitation of this paper is that we do not categorize covariant shift and concept drift over time. Though, we've seen some sudden distribution shifts occur in our benchmark, we currently do not find a good way to precisely identify the reasons of sudden distribution shifts and further categorize them. We will focus on this in the next version.

\subsection{\revision{Ethics Discussion}}
The Wild-Time benchmark includes the Yearbook dataset, which is an adaptation of the Portraits dataset \cite{ginosar2015century}. The task is binary gender prediction from yearbook photos of American high schoolers. We recognize the harmful ramifications of binary gender prediction. A binary gender prediction task excludes nonbinary individuals, may misgender transgender individuals, and may reinforce problematic gender norms.

\revision{The FMoW-Time dataset, adapted from the WILDS benchmark \cite{koh2021wilds}, involves geographic region prediction from satellite imagery and has applications to remote sensing. We recognize the privacy and surveillance issues surrounding remote sensing. We remark that FMoW-Time uses a lower image resolution than other publicly available satellite data, such as Google Maps. We also recognize that the FMoW-Time dataset raises issues of systematic bias and fairness. Specifically, the WILDS benchmark \cite{koh2021wilds} found that models performed poorly on satellite images from Africa. As remote sensing is used for development and humanitarian purposes, poor model performance in certain geographic regions can harm certain populations. These issues are discussed in more detail in the UNICEF discussion paper by Berman et al. \cite{berman2018ethical}.

The MIMIC dataset, adapted from the MIMIC-IV database \cite{johnson2020mimic}, involves predicting patient mortality and readmission to the ICU. Ethical challenges associated with using artificial intelligence (AI) in healthcare include (1) informed consent to use, (2) safety and transparency, (3) fairness and algorithmic biases, and (4) data privacy \cite{gerke2020ethical}. The MIMIC-IV database adopted a permissive access scheme, allowing for broad reuse of data. With regards to patient privacy, we note that the MIMIC-IV database includes de-identified patient data \cite{johnson2020mimic}. We also note that the authors of Wild-Time followed proper credentialing protocol to access the MIMIC-IV dataset. To protect patient confidentiality, we do not release the MIMIC dataset. Instead, we provide instructions for how users can get credentialed on PhysioNet to download the MIMIC-IV dataset and provide a script to generate the MIMIC dataset. We recognize considerations of fairness and algorithmic bias for the MIMIC task. Several studies have found that AI algorithms exhibit biases with respect to ethnicity and gender \cite{mehrabi2021survey, noseworthy2020assessing}. Phenotype-related data in healthcare can similarly lead to biased models. This can result in incorrect diagnoses for certain subpopulations, endangering their safety. Finally, we emphasize the importance of robust and interpretable AI, especially in healthcare, where human safety is at stake. We hope that the MIMIC task can help lay the groundwork for further research in this direction. We refer readers to \cite{jobin2019global} for an in-depth discussion of the ethical issues surrounding AI in healthcare.

\subsection{Comments on Designing Temporally Robust Models}
In our experiments, we found that most existing approaches can not effectively mitigate natural temporal distribution shifts. We believe that there are two important aspects to consider in resolving natural distribution shift:
\begin{itemize}[leftmargin=*]
    \item \textbf{Learning changeable temporal invariance}. To build a robust model, it would be useful to learn invariance, which captures features in the data that remain invariant across different distributions. However, this is difficult to do when temporal distribution shift happens, as such invariance can also change over time, where one kind of invariance is only suitable for a specific time window. Capturing the correlations between different time windows and determining when and how to update the invariant model are crucial.
    \item \textbf{Leveraging supervised and unsupervised adaptation}. In addition to maintaining a temporally invariant model, adapting to new timestamps is also necessary in tackling temporal distribution shifts. Here, we can leverage labeled data from timestamps in the near past and unlabeled observations from the current timestamp to fine-tune the model. How to combine temporal invariance with supervised and unsupervised adaptation to achieve effective adaptation remains an open problem.

\end{itemize}
}

\begin{table*}[t]
\small
\caption{The in-distribution versus out-of-distribution test performance of each method evaluated on Wild-Time under the Eval-Fix setting. The average and standard deviation (value in parentheses) are computed over three random seeds. We bold the best OOD performance for each dataset.}
\vspace{-1em}
\label{tab:results_offline}
\begin{center}
\resizebox{\columnwidth}{!}{\setlength{\tabcolsep}{2.5mm}{
\begin{tabular}{l|ccc|ccc}
\toprule
\multirow{3}{*}{} & \multicolumn{3}{c|}{Yearbook} & \multicolumn{3}{c}{FMoW-Time}  \\
& \multicolumn{3}{c|}{(Accuracy (\%) $\uparrow$)} & \multicolumn{3}{c}{(Accuracy (\%) $\uparrow$)}\\
& ID Avg. & OOD Avg. &  OOD Worst & ID Avg. &  OOD Avg. & OOD Worst \\\midrule
Fine-tuning & 95.43 (1.65) & 81.98 (1.52) & \textbf{69.62 (3.38)} & 47.67 (0.81) & 44.22 (0.56) & 36.59 (0.98) \\
EWC  & 96.36 (0.47) & 80.07 (0.22) & 66.61 (1.95) & 47.42 (0.64) & 44.02 (0.65) & 36.42 (1.56) \\
SI & 96.40 (0.83) & 78.70 (3.78) & 65.18 (2.44) & 47.41 (0.36) & 44.25 (0.97) & 37.14 (1.34) \\
A-GEM & 97.18 (0.43) & 81.04 (1.40) & 67.07 (2.23) & 47.09 (0.36) & 44.10 (0.65) & 36.02 (0.61) \\\midrule
ERM  & 97.99 (1.40) & 79.50 (6.23) & 63.09 (5.15) & 58.07 (0.15) & \textbf{54.07 (0.25)} & 46.00 (0.79) \\
GroupDRO-T  & 96.04 (0.45) & 77.06 (1.67) & 60.96 (1.83) & 46.57 (0.57) & 43.87 (0.55) & 36.60 (1.25) \\
mixup  & 96.42 (0.26) & 76.72 (1.35) & 58.70 (1.36) & 56.93 (0.50) & 53.67 (0.49) & 44.57 (1.02) \\
LISA  & 96.56 (0.97) & 83.65 (4.61) & 68.53 (5.79) & 55.10 (0.26) & 52.33 (0.42) & 43.30 (0.75) \\
CORAL-T  & 98.19 (0.58) & 77.53 (2.15) & 59.34 (1.46) & 52.60 (0.10) & 49.43 (0.38) & 41.23 (0.78) \\
IRM-T  & 97.02 (1.52) & 80.46 (3.53) & 64.42 (4.38) & 46.60 (1.40) & 45.00 (1.18) & 37.67 (1.17) \\\midrule
\revision{SimCLR} & \revision{96.11 (0.92)} & \revision{78.59 (2.72)} & \revision{60.15 (3.48)} & 46.91 (0.65) & 44.76 (0.17) & 37.00 (0.44) \\
\revision{SwaV} & \revision{96.24 (0.58)} & \revision{78.38 (1.86)} & \revision{60.73 (1.08)} & 47.31 (0.46) & 44.92 (0.81) & 37.17 (0.52) \\\midrule
\revision{SWA} & \revision{98.46 (0.15)} & \revision{\textbf{84.25 (3.06)}} & \revision{67.90 (4.34)} & 58.05 (0.14) & 54.06 (0.23) & \textbf{46.01} (0.78) \\\midrule
\multirow{3}{*}{} & \multicolumn{3}{c|}{MIMIC-Readmission} & \multicolumn{3}{c}{MIMIC-Mortality} \\
& \multicolumn{3}{c|}{(Accuracy (\%) $\uparrow$)} & \multicolumn{3}{c}{(AUC (\%) $\uparrow$)} \\
& ID Avg. & OOD Avg. &  OOD Worst & ID Avg. &  OOD Avg. & OOD Worst \\\midrule
Fine-tuning & 69.71 (10.8) & 62.19 (3.71) & 59.57 (4.43) & 89.99 (0.98) & 63.37 (1.91) & 52.45 (2.64) \\
EWC  & 77.78 (0.38) & \textbf{66.40 (0.09)} & \textbf{64.69 (0.01)} & 89.53 (0.65) & 62.07 (1.52) & 50.41 (2.03) \\
SI & 71.28 (6.22) & 62.60 (3.27) & 61.13 (3.39) & 89.25 (0.84) & 61.76 (0.58) & 50.19 (1.25) \\
A-GEM & 73.56 (3.25) & 63.95 (0.14) & 62.66 (1.23) & 88.74 (0.17) & 61.78 (0.27) & 50.40 (0.51) \\\midrule
ERM & 73.00 (2.94) & 61.33 (3.45) & 59.46 (3.66) & 90.89 (0.59) & 72.89 (8.96) & 65.80 (12.3) \\
GroupDRO-T  & 69.70 (4.71) & 56.12 (4.35) & 54.69 (4.36) & 89.22 (0.46) & \textbf{76.88 (4.74)} & \textbf{71.40 (6.84)} \\
mixup  & 70.08 (2.14) & 58.82 (4.03) & 57.30 (4.77) & 89.75 (1.04) & 73.69 (7.83) & 66.83 (11.1) \\
LISA  & 70.52 (1.10) & 56.90 (0.95) & 54.01 (0.92) & 89.29 (0.47) & 76.34 (8.94) & 71.14 (12.4) \\
CORAL-T  & 70.18 (4.72) & 57.31 (4.45) & 54.69 (4.36) & 88.77 (0.97) & 77.98 (2.57) & 64.81 (10.8) \\
IRM-T  & 72.33 (1.50) & 56.53 (3.36) & 52.67 (5.17) & 89.49 (0.17) & 76.17 (6.32) & 70.64 (8.99) \\\midrule
\revision{SWA} & 72.62 (3.60)  & 59.88 (5.48) & 57.68 (6.36) & 89.53 (1.96) & 69.53 (1.60) & 60.83 (2.73) \\\midrule
\multirow{3}{*}{} &\multicolumn{3}{c|}{HuffPost} & \multicolumn{3}{c}{arXiv} \\
& \multicolumn{3}{c|}{(Accuracy (\%) $\uparrow$)} & \multicolumn{3}{c}{(Accuracy (\%) $\uparrow$)} \\
& ID Avg. & OOD Avg. &  OOD Worst & ID Avg. &  OOD Avg. & OOD Worst \\\midrule
Fine-tuning & 76.79 (0.51) & 69.59 (0.10) & 68.91 (0.49) & 51.42 (0.15) & 50.31 (0.39) & 48.19 (0.41) \\
EWC & 76.26 (0.32) & 69.42 (1.00) & 68.61 (0.98) & 51.34 (0.13) & \textbf{50.40 (0.11)} & \textbf{48.18 (0.18)} \\
SI & 76.97 (0.30) & 70.46 (0.27) & 69.05 (0.52) & 51.52 (0.19) & 50.21 (0.40) & 48.07 (0.48) \\
A-GEM & 77.15 (0.07) & 70.22 (0.50) & 69.15 (0.88) & 51.57 (0.18) & 50.30 (0.37) & 48.14 (0.40) \\\midrule
ERM & 79.40 (0.05) & 70.42 (1.15) & 68.71 (1.36) & 53.78 (0.16) & 45.94 (0.97) & 44.09 (1.05) \\
GroupDRO-T & 78.04 (0.26) & 69.53 (0.54) & 67.68 (0.78) & 49.78 (0.22) & 39.06 (0.54) & 37.18 (0.52) \\
mixup & 80.15 (0.17) & \textbf{71.18 (1.17)} & 68.89 (0.38) & 51.40 (0.20) & 45.12 (0.71) & 43.23 (0.75) \\
LISA & 78.20 (0.53) & 69.99 (0.60) & 68.04 (0.75) & 50.72 (0.31) & 47.82 (0.47) & 45.91 (0.42) \\
CORAL-T & 78.19 (0.31) & 70.05 (0.63) & 68.39 (0.88) & 53.25 (0.12) & 42.32 (0.60) & 40.31 (0.61) \\
IRM-T & 78.38 (0.51) & 70.21 (1.05) & 68.71 (1.13) & 46.30 (0.53) & 35.75 (0.90) & 33.91 (1.09) \\\midrule
SWA & 80.40 (0.22) & 70.98 (0.05) & \textbf{69.52 (0.10)} & 51.42 (0.30) & 44.36 (0.77) & 42.54 (0.68) \\
\bottomrule
\end{tabular}}}
\vspace{-2em}
\end{center}
\end{table*}

\end{document}